\documentclass[journal]{IEEEtran}
\usepackage{stfloats}
\usepackage{amsmath}
\usepackage{amssymb}
\usepackage{textcomp,booktabs}
\usepackage[dvipsnames, svgnames, x11names]{xcolor}
\usepackage{wrapfig}
\usepackage{colortbl,booktabs}
\usepackage{verbatim}
\usepackage{epstopdf}
\usepackage[colorlinks,linkcolor=blue]{hyperref}

\usepackage{floatrow}

\usepackage{booktabs,multirow,rotating}
\usepackage{cite}

\usepackage{subfigure}
\usepackage{bm}
\usepackage{caption}

\usepackage{float}

\makeatletter
\renewcommand{\@thesubfigure}{\hskip\subfiglabelskip}
\makeatother

\usepackage{colortbl}
\definecolor{mygray}{gray}{.9}
\usepackage{diagbox}
\usepackage{makecell}
\usepackage{hhline}

\usepackage{amsthm}

\begin{document}
%

\title{Salient Object Detection via Dynamic Scale Routing}


\author{Zhenyu Wu$^{1}$  ~~~~~~Shuai Li$^{1,4}$
 ~~~~~~Chenglizhao Chen$^{1,2*}$\thanks{Corresponding author: Chenglizhao Chen, cclz123@163.com.}  ~~~~~~Hong Qin$^3$ ~~~~~~Aimin Hao$^{1,4}$\\

$^1$State Key Laboratory of Virtual Reality Technology and Systems, Beihang University\\
$^2$China University of Petroleum (East China)~~~$^3$Stony Brook University ~~~$^4$ Peng Cheng Laboratory \\

%
}

\markboth{IEEE Transactions on Image Processing, VOL.XX, NO.XX, XXX.XXXX}%
{Shell \MakeLowercase{\textit{et al.}}: Bare Demo of IEEEtran.cls for Journals}

\maketitle

\IEEEtitleabstractindextext{
\begin{abstract}
Recent research advances in \textbf{s}alient \textbf{o}bject \textbf{d}etection (SOD) could largely be attributed to ever-stronger multi-scale feature representation empowered by the deep learning technologies. The existing SOD deep models extract multi-scale features via the off-the-shelf encoders and combine them smartly via various delicate decoders. However, the kernel sizes in this commonly-used thread are usually "fixed". In our new experiments, we have observed that kernels of small size are preferable in scenarios containing tiny salient objects. In contrast, large kernel sizes could perform better for images with large salient objects. Inspired by this observation, we advocate the "dynamic" scale routing (as a brand-new idea) in this paper. It will result in a generic plug-in that could directly fit the existing feature backbone. This paper's key technical innovations are two-fold. First, instead of using the vanilla convolution with fixed kernel sizes for the encoder design, we propose the \textbf{d}ynamic \textbf{p}yramid \textbf{con}volution (DPConv), which dynamically selects the best-suited kernel sizes w.r.t. the given input. Second, we provide a self-adaptive bidirectional decoder design to accommodate the DPConv-based encoder best. The most significant highlight is its capability of routing between feature scales and their dynamic collection, making the inference process scale-aware. As a result, this paper continues to enhance the current SOTA performance.
Both the code and dataset are publicly available at \emph{\url{https://github.com/wuzhenyubuaa/DPNet}}.
\end{abstract}


}
\maketitle
\IEEEdisplaynontitleabstractindextext
\IEEEpeerreviewmaketitle


\section{Introduction}
\label{intro}

\IEEEPARstart{S}alient \textbf{o}bject \textbf{d}etection (SOD) aims to identify the most visually attractive objects from a given
image, which can be applied to video edit \cite{wang2019inferring}, photo cropping\cite{wang2020paying} and semantic segmentation \cite{wang2022looking}. 
Early SOD
literatures~\cite{OurTIP15,cheng2015global,ma2003contrast,achanta2009frequency}
have primarily followed the bottom-up methodology, which typically
relies on various hand-crafted saliency cues. However, the common
nature of these hand-crafted features is not generic in
essence, so they tend to yield unsatisfactory performance in complex
scenarios. In the deep learning era, the advent of
\textbf{c}onvolutional \textbf{n}eural \textbf{n}etwork\textbf{s}
(CNNs), which is essentially top-down, significantly improves the
conventional hand-crafted approaches with ever stronger feature
representation, meanwhile the overall computational speed also gets
boosted after the widespread deployment of the seminal \textbf{f}ully
\textbf{c}onvolutional \textbf{n}etwork\textbf{s}
(FCNs)~\cite{long2015fully}, where the time-consuming fully connected
layers are replaced by convolutional
layers. To
date, various FCN-based end-to-end SOD models have been proposed,
whose key technical innovations could be coarsely categorized as
follows: modifying loss
functions~\cite{wei2020f3net,pang2020multi,zhou2020interactive},
utilizing edge
labels~\cite{AFNet,su2019selectivity,zhao2019egnet,wei2020label,zhou2020interactive},
combining with depth
information~\cite{li2020rgb,zhang2021uncertainty,fu2021siamese},
devising attention-based
fusion~\cite{chen2018reverse,PiCANet,wang2019salient,PAGRN,CPD},
adopting progressive
training~\cite{wang2016saliency,wei2020f3net,wang2020progressive}, and
exploring the appropriate multi-scale feature
aggregation~\cite{Amulet,RADF,pang2020multi,BASNet19,wei2020f3net}.

\begin{figure}[!t]
\centering
\includegraphics[width=1\linewidth]{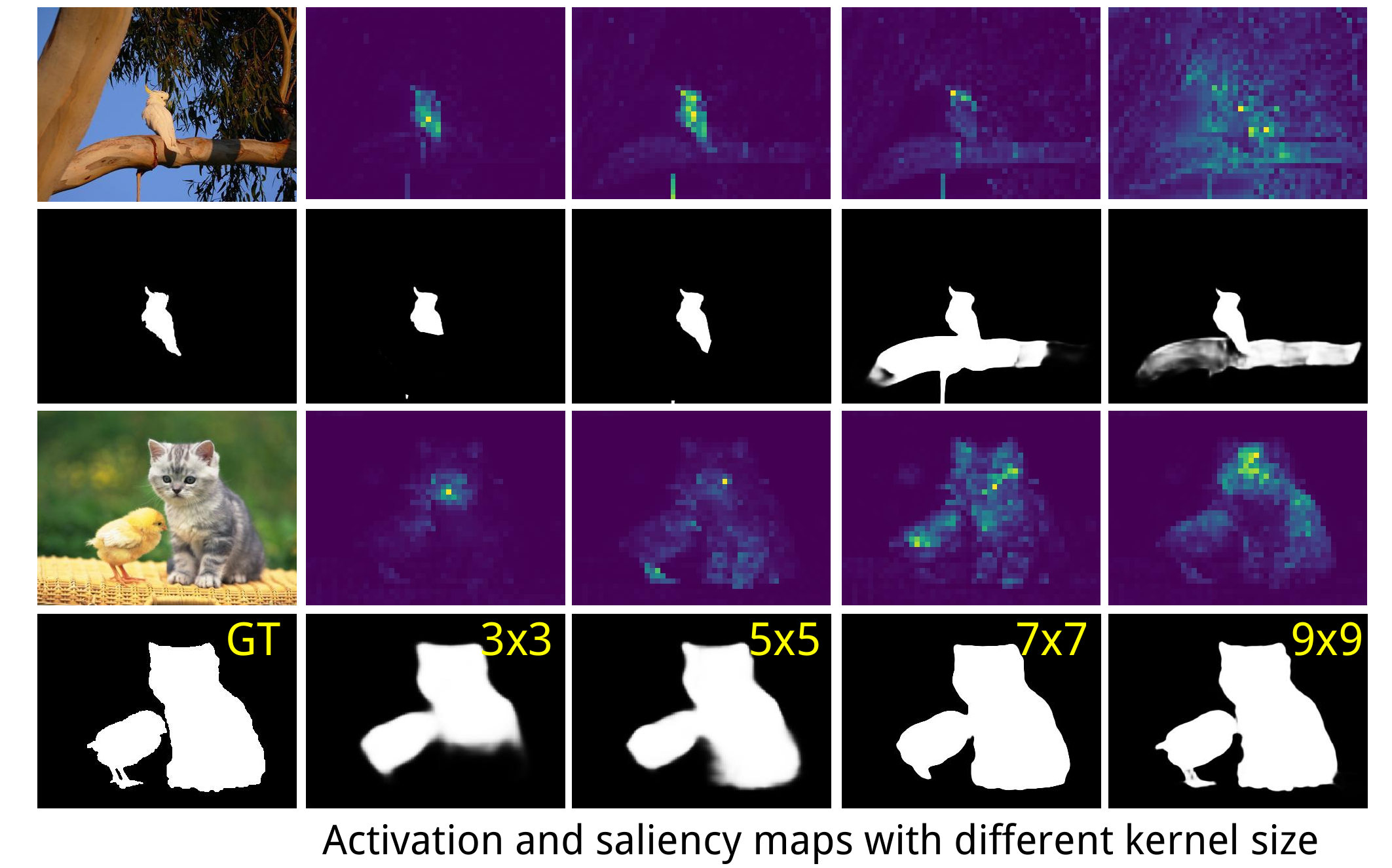}
\vspace{-0.7cm}
\caption{Two most representative illustrations of activation maps at
  Conv5\_3 of ResNet50. As can be seen in the first two rows, the salient
  bird, which is a relatively small-sized object, could get
  massive false-alarms in its surroundings if we use a large kernel
  size (e.g., $9\times 9$), yet a small kernel size (e.g., $3\times
  3$) can easily achieve better detection.  Large kernel sizes are preferable to the big objects, as expected.\vspace{-0.3cm}}
\label{multi_scale}
\end{figure}

Although remarkable progress has been achieved, there remain big
challenges for the existing SOD models to meet the requirements in
real-world applications, where failure cases keep occurring. Through
our extensive and careful studies of existing literature (and also
augmented by our most recent experiments), we have observed one
critical factor, which has long been ignored in previous literature,
could have the potential to continue to enhance the SOD performance. 
This factor hinges upon the fact that the existing SOD models usually
perform their convolutional operations using the pre-defined kernels
of fixed sizes for all images, and inappropriate kernel sizes most likely cause failure cases. For example, successful
detections of large salient objects tend to appear in those
layers using large kernel sizes, while kernels with small sizes are
preferable for other images. We believe this
is a chicken-and-egg problem since the widely-used training process
and network are completely scale-unaware. We attribute this
phenomenon as \textit{``scale confusion''}, which causes 
learning ambiguity and encoding difficulty towards a perfect
inference, as shown in Fig.~\ref{multi_scale}.

Previous
models~\cite{Amulet,RADF,pang2020multi,BASNet19,wei2020f3net}
have attempted to alleviate this dilemma by collecting multi-scale
features obtained from each layer of an encoder in a ``static.''
fashion, i.e., kernels' sizes are all fixed during the whole
process. The features provided by a static encoder usually have a very
the unique characteristic, i.e., each of which respectively represents its
own individual scale independently. Thus, given the existing
\textbf{s}tate-\textbf{o}f-\textbf{t}he-\textbf{a}rt (SOTA) models,
these scale-specific features shall be considered equally important in
advance, with the hope that their decoders could automatically formulate
a series of feature combination rules in pursuit of a near-optimal
full feature complement towards the desired learning objective. In
fact, the nature of this naive ``feature combining'' is a variant of
the well-known weighted summation, which selectively integrates all
saliency cues revealed in different reception fields, where the
weights are formulated during the online learning process. But
the problem is that this widely-used learning process is completely
static with its own shortcoming since the SOD's problem domain is
extremely large, so those implicitly learned feature combination rules
are essentially too sparse to have a compact fit. Instead of
continuing to seek better feature combination weights in the learning
process pertinent to the existing SOTA encoders, our current thought
is to replace the widely-used static kernels with possible
dynamically-adjusted ones in an encoder to combat this
dilemma.

\begin{figure*}[!t]
\centering
\includegraphics[width=0.9\linewidth]{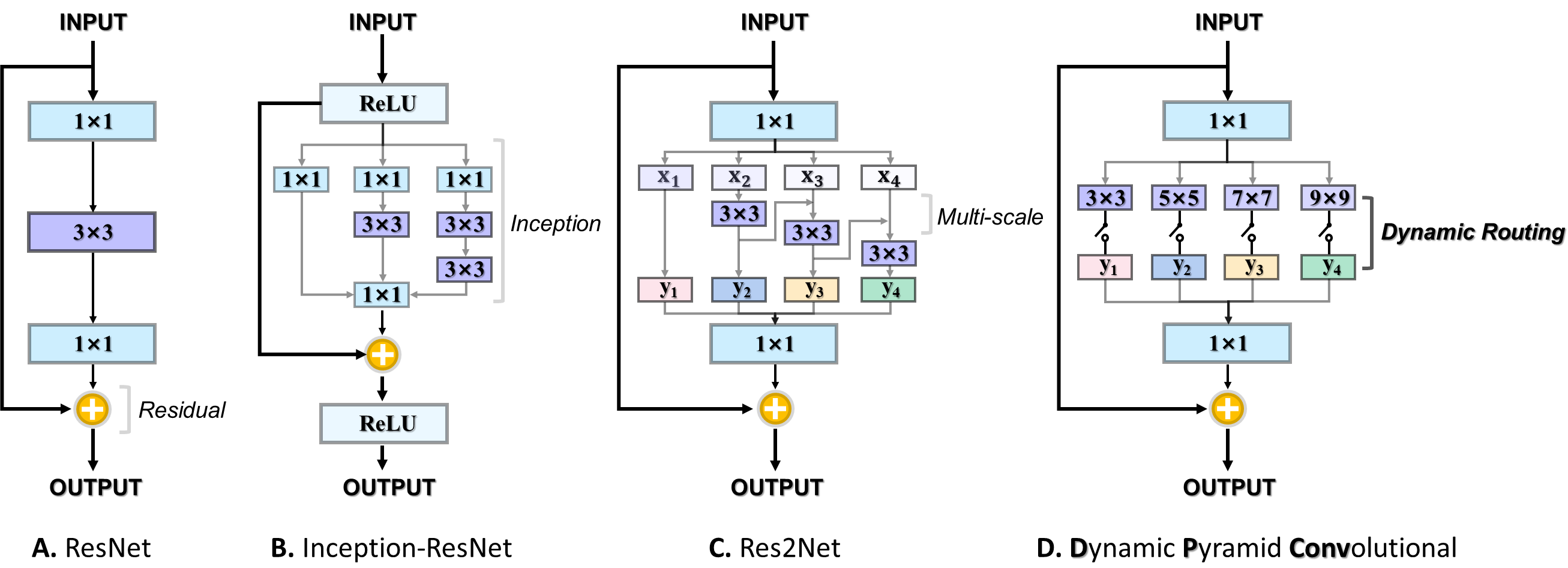}
\vspace{-0.0cm}
\caption{Comparing (A) ResNet~\cite{he2016deep}, (B)
  Inception-v4~\cite{szegedy2017inception}, and (C)
  Res2Net~\cite{2021res2net} with (D) our DPConv, where ``+''
  represents the addition operation. The major advantage of our DPConv
  is its dynamic routing capability of between scales.}
\label{bottleneck}
\end{figure*}

In this paper, we propose a novel yet effective \textbf{d}ynamic
\textbf{p}yramid \textbf{net}work (DPNet), which comprises two
technical innovations relevant to both the encoder and decoder
respectively. As for the encoder's design, we have newly devised the
\textbf{d}ynamic \textbf{p}yramid \textbf{conv}olution (DPConv) to
address the challenge in the current widely-used paradigm, where the
``static'' convolutional kernels are replaced by ``dynamic'' ones.
We refer readers to Fig.~\ref{bottleneck}, which illustrates the
differences between our DPConv and the widely-used common threads for a better understanding. The primary differences of the proposed DPConv can be highlighted as follows. First, it provides the flexibility of
choosing eligible feature scales at the encoding stage, while the
existing methods can only resort to their decoders for an appropriate
scale combination. Second, the proposed DPConv could directly serve as
a generic plug-in for the existing feature backbone with a consistent
and enhanced performance (refer to Table~\ref{ablation_tab1} for the
quantitative evidence).

\begin{figure}[t]
\centering
\includegraphics[width=0.95\linewidth]{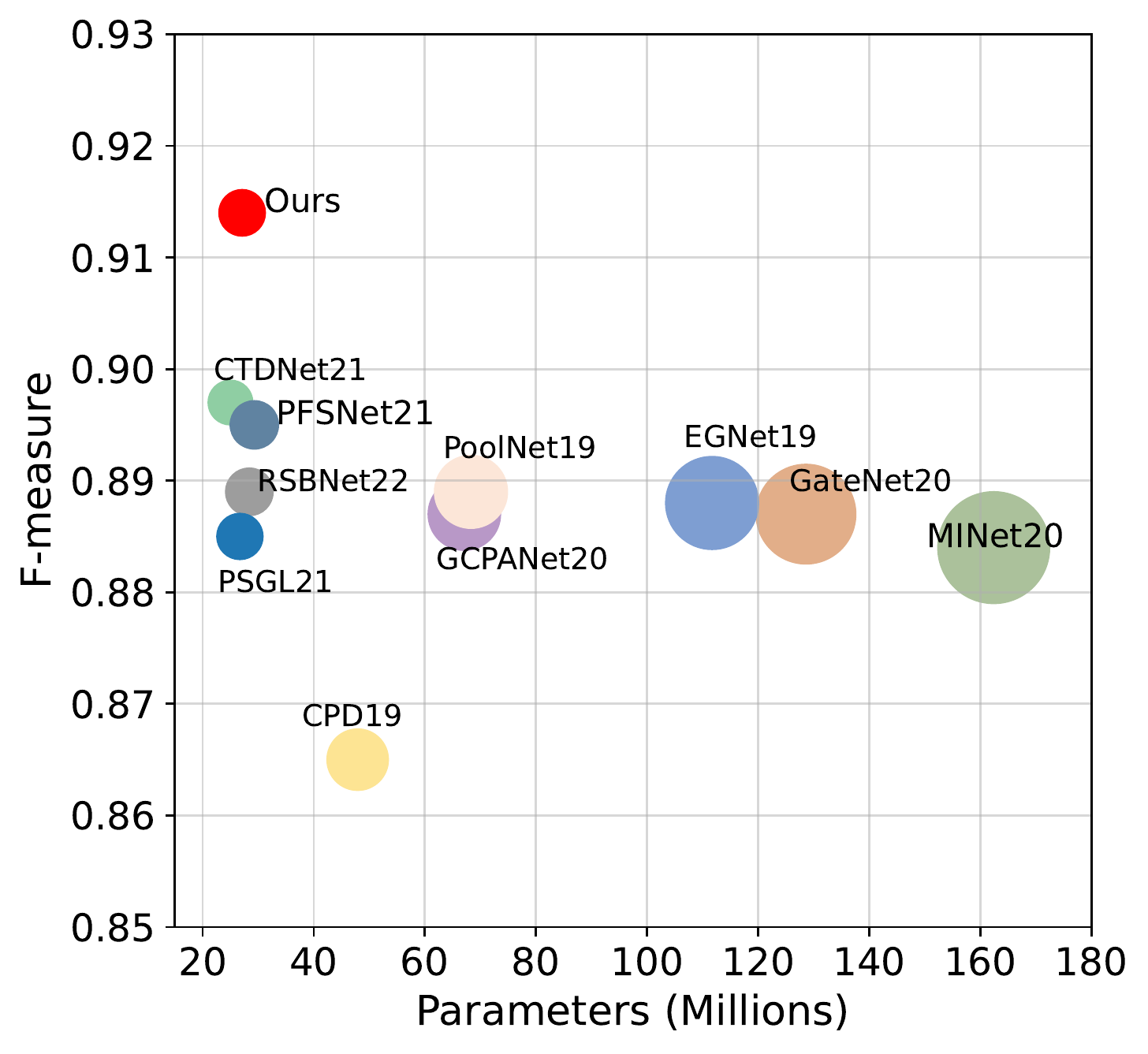}
\vspace{-0.0cm}
\caption{\textbf{F-measure (max) and Parameters} comparisons of our
  DPNet with other existing SOTA models on the challenging DUTS-TE
  set, where each circle's size is positively proportional to the
  model's parameter number. More results can be found in
  Fig.~\ref{params_flops_model_size}.}
\vspace{-0.2cm}
\label{params}
\end{figure}

As for the decoder design, we present a novel self-adaptive
bidirectional decoder (see Fig.~\ref{decoder}), which also follows the
same ``dynamic'' rationale, whose goal is to achieve scale-aware
feature combination. Most of the existing decoders usually utilize the
static one-directional feature aggregation, which either complements
low-level tiny details with high-level coarse localization
information, or vice versa. Since they have adopted the conventional
static encoders (i.e., with fixed kernel sizes), their decoders' input
could individually represent different feature scales, making the
static one-directional feature aggregation reasonable and
feasible. To better match the new attribute of DPConv, it calls for a novel decoder design which could combine them densely and dynamically in both directions, and that's why we devised the self-adaptive bidirectional
decoder to perform feature collection in both bottom-up and top-down
fashions.

Extensive quantitative experiments have verified that the proposed
DPNet significantly outperforms the existing SOTA models on six
popular datasets under various evaluation metrics. As shown in
Fig.~\ref{params}, on the DUTS-TE\cite{wang2017learning} dataset,
our DPNet shows $1.6\%$ improvement in terms of max F-measure
over the second best CTDNet21 \cite{zhao2021complementary}, while the
computational costs are essentially the same. In addition, our DPNet
achieves higher than GateNet20~\cite{zhao2020suppress}
(88.7$\rightarrow$91.6$\%$), and the number of parameters of DPNet is
only 27.1 million, which is 4.8$\times$ fewer than that of GateNet20.
In summary, this paper makes the following salient contributions:
\begin{itemize}
\item
We have conducted an in-depth investigation into the relationship
between kernel sizes and the model's SOD performance, which rise to a possible technical solution of using dynamic kernel sizes with more
potential to break the current SOTA performance bottleneck.
\item
Motivated by this dynamic rationale, we have also devised the new
DPConv, which could serve as a generic plug-in. This solution
addresses the challenge of the widely-used paradigm built upon static
convolutional kernels, making the features obtained from the encoder
completely scale-aware.
\item
We have also afforded a delicate solution for the decoder design,
which turns the static feature aggregation into a dynamic one. Collectively, we present a feasible roadmap with the potential to
guide future research in the improved design of DPConv-compatible
decoders. We also conducted extensive quantitative evaluations to confirm the effectiveness of each participating component adopted in our new paradigm.

\end{itemize}

\section{Related Works}
\label{sec:related}

\subsection{Deep SOD Model}

Existing SOD approaches can be divided into two groups: conventional handcrafted and deep learning-based. Early non-deep learning SOD works usually follow the bottom-up methodology, which collects low-level handcrafted features (e.g., color contrast~\cite{cheng2015global}, background prior~\cite{wei2012geodesic}) to formulate the final detection. However, these works tend to show poor generalization ability, and their results in complex scenarios are extremely worse.
More content regarding this topic can be found in~\cite{borji2015salient}.

After entering the deep learning era, more and more CNNs based works achieved impressive results in various computer vision tasks, and the deep learning based SOD has received extensive research attentions \cite{xu2021locate,liu2021rethinking,li2021salient,wu2021decomposition,tang2021disentangled,zhuge2022salient,wu2022synthetic,wu2020deeper,wu2022recursive,CCTIP17,chen2020xuehao,ChenPR16}.
Although these methods achieved promising performances against the handcrafted ones, they still have various limitations, e.g., both computation and memory costs are relatively high.
To improve it, FCNs~\cite{long2015fully} based works formulate the SOD task as a pixel-wise end-to-end binary classification, and ever since then, this methodology has dominated the whole SOD research field in terms of both accuracy and efficiency.
Subsequent works mainly focused on appropriately aggregating multi-scale features toward the given learning objective.
Most of them~\cite{SRM,Amulet,RADF,DGRL,BMP,wang2019an} followed the encoder-decoder structure.
The encoder, also known as the feature backbone, extracts features in a ``static'' manner --- each of its convolutional layers strictly correlates to a fixed scale, and thus these features together could be able to span a multi-scale feature space.
Thus, the primary target of the decoder is to formulate a series of decision-making rules in this feature space for the given SOD task.
We shall review some most representative works.

Hou~\textit{et al}.~\cite{DSS} proposed to integrate both high-level and low-level features via one-directional cross-scale short connections. Different to~\cite{DSS}, which only uses features in specific levels, Zhang~\textit{et al.}~\cite{Amulet} performed the bi-directional feature integration, which could interact with low-level features and high-level features, achieving the mutual complementary status.
Similarly, Wang~\textit{et al.}~\cite{wang2020progressive} proposed a novel multi-scale feature refinement scheme, which could be able to polish its features progressively.
Pang~\textit{et al.}~\cite{pang2020multi} proposed an interactive integration network which fuses multi-scale features to deal with the prevalent scale variation issue.
Chen~\textit{et al.}~ \cite{chen2020global} proposed a global context-aware progressive aggregation network, which integrates low-level details, high-level semantics, and global contexts in an interweaved way. As pointed by the \cite{wang2021salient}, exploring dynamic network structures in SOD is promising for improving efficiency and effectiveness. Wang~\textit{et al.}~\cite{wang2019an} integrated both top-down and bottom-up saliency inferences in an iterative and cooperative manner. 

The works most relevant to us are \cite{wang2019an,tan2020efficientdet}.
Our DPNet has three main differences with the \cite{wang2019an}: \textbf{1)} From the perspective of motivation, our DPNet was designed to solve the "scale confusion" phenomenon while the \cite{wang2019an} was proposed to mimic the human visual information processing, i.e., bottom-up and top-down processing.
\textbf{2)} From the perspective of function, our DPNet focus on "dynamically" generating better feature representation according to the content of the given input image while the \cite{wang2019an} was proposed to learn the complementary features by aggregating low-level and high-level features. \textbf{3)} From the perspective of decoder, our DPNet adopt the Bidirectional Cross-Fusion Module (BiCFM) while the \cite{wang2019an} use employ RNN for top-down/bottom-up inference.
In addition, our BiCFM has two main differences with the BiFPN \cite{tan2020efficientdet} in network structure.  \textbf{1)} Each node of our BiCFM module has two sets of features while the top and bottom node of BiFPN has one input edge with no feature fusion. \textbf{2)} Moreover, our BiCFM and the BiFPN has different cross-scale feature fusion module, i.e., Cross Feature Module (CFM) v.s. Weighted Feature Fusion (WFF). Formally, give a list of cross-scale features $\{F_{l_1}^{in}, F_{l_2}^{in},... \}$, where $F_{l_i}^{in}$ represents the feature at level $l_i$. Compared with WFF, our CFM avoids redundant information introduced to $F_{l_{i}}^{in}$ and $F_{l_{i+1}}^{in}$, which may pollute the original features and bring adverse effect to the generation of saliency maps. By multiple feature crossings, $F_{l_{i}}^{in}$ and $F_{l_{i+1}}^{in}$ will gradually absorb useful information from each other to complement themselves, i.e., noises of $F_{l_{i}}^{in}$ will be suppressed and boundaries of $F_{l_{i+1}}^{in}$ will be sharpened.

\subsection{Dynamic Network}

In contrast to the standard paradigm which uses fixed kernel sizes, dynamic CNNs \cite{bolukbasi2017adaptive,wang2018skipnet,wu2018blockdrop,klein2015dynamic,jia2016dynamic,yang2019condconv,chen2020dynamic,li2019selective} use dynamic modules, kernels, width or depth conditioned on the given input.

One representative line is to learn an auxiliary controller to determine which network's components should be skipped.
As one of the pioneering works, Bolukbasi~\textit{et al.}~\cite{bolukbasi2017adaptive} devised an adaptive network evaluation scheme which adaptively chooses the most valuable network's components.
In an associative reinforcement learning setting, SkipNet~\cite{wang2018skipnet} and BlockDrop~\cite{wu2018blockdrop}, which adopted very similar ideas to that of~\cite{bolukbasi2017adaptive}, proposed to learn policy networks to dynamically determine their networks' architectures.

Another line is to determine the model's capacity dynamically.
DCNet~\cite{klein2015dynamic} proposed to generate convolution kernels dynamically via a linear layer.
DFNet~\cite{jia2016dynamic} introduced a new framework, the dynamic filter network, where spatial filters are generated dynamically to suppress spatial redundancy.
CondConv~\cite{yang2019condconv} and DCNet~\cite{chen2020dynamic} proposed to replace the static convolution kernels by a linear combination of $n$ experts with an identical kernel size. Theoretically, CondConv and DCNet belong to linear aggregation, which is insufficient to provide powerful adaptation ability to fit all its given learning objectives.
In contrast, we propose the DPConv, which aggregates information from multiple different kernel sizes to provide the flexibility of choosing eligible feature scales, ensuring a more effective multi-scale feature representation. 
SKNet \cite{li2019selective} proposed a novel Selective Kernel convolution to improve the efficiency and effectiveness of object recognition by adaptive kernel selection in a soft-attention manner. Our DPConv has two main differences from the SKNet. \textbf{1)} From the perspective of motivation, our DPConv was designed to solve the ``scale confusion'' phenomenon, while the SKNet was proposed to mimic the receptive fields of neurons in the primary visual cortex of the human visual system. \textbf{2)} 
From the implementation perspective, our DPConv consists of a set of pyramid convolution $\{3\times3, ..., N\times N\}$ while SK convolution only consists of kernel size $3\times3$ and $5\times5$. Thus, the SK convolution is only a special case of our DPCconv. In addition, our DPNet is more lightweight than SKNet (parameters: 27.1M v.s 27.5M).

\begin{figure*}[!t]
\centering
\includegraphics[width=0.8\linewidth]{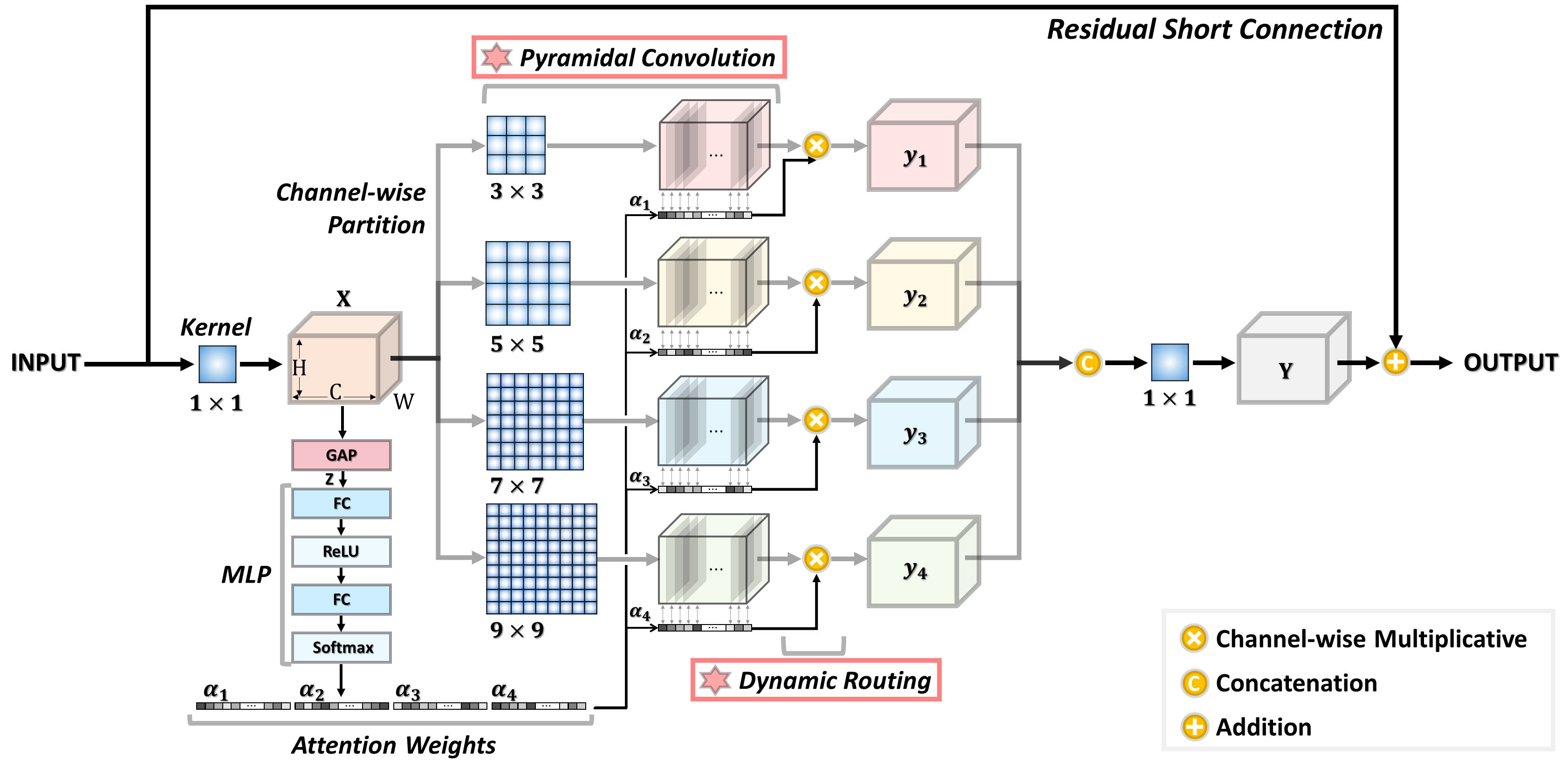}
\caption{Network detail of the proposed DPConv block, where the key technical innovations include 1) pyramid convolution (Sec.~\ref{sec:GDC}) and 2) dynamic routing (Sec.~\ref{sec:dr}).}
\label{detailed_block}
\end{figure*}

\section{Dynamic Pyramid Convolution}
\label{DPConv}

To address the ``scale confusion'' as shown in Fig. \ref{multi_scale}, we present a novel plug-and-play module, i.e., dynamic pyramid convolution (DPConv), which can deal with the scale confusion by ``dynamically'' generating better feature representation according to the content of the given input image. The proposed DPConv is flexible in choosing eligible feature scales while retaining a very small parameter size.

Our rationale is mainly based on the following insights: the receptive fields of neurons in the primary visual cortex are not ``static'' but ``dynamically'' modulated by various stimuli. Analogously, the receptive field sizes of one convolutional layer should also be dynamically adjusted accordingly.
The proposed DPConv consists of a group of ``pyramidal convolution (Sec.~\ref{sec:GDC})'' and ``dynamic routing (Sec.~\ref{sec:dr})''. A detailed illustration of DPConv's structure can be seen in Fig.~\ref{detailed_block}.

\subsection{Pyramid Convolution}
\label{sec:GDC}
In a classic convolutional layer, a series of pre-defined kernels with fixed sizes (e.g., $3 \times 3$) is used for all input examples. In our pyramidal convolution block, we replace the $3 \times 3$ filters with a group of pyramidal kernels (e.g., $3 \times 3,..., K_m \times K_m$), and the exact kernel number is defined by the hyperparameter $m$.

For any given input feature tensor $\textbf{X}$, we apply a series of convolutions $\{\textbf{W}_1, \textbf{W}_2, ..., \textbf{W}_m\}$ to each individual input feature, then the intermediate feature representation of each convolution group is $\textbf{y}_i = \textbf{W}_{i}*\textbf{X} $, and $i\in \{1,2,...,m\}$.
By using these ``pyramidal'' convolutions, the learning capacity can be increased.
Besides, the proposed pyramidal convolution using multiple kernels with different sizes has a unique advantage, i.e., \emph{\textbf{it enables each convolution block to be inner multi-scale}}.

However, a plain implementation of pyramid convolution could lead to additional parameters.
To make the pyramid convolution block more efficient, we resort to the group convolution~\cite{resnext2017}, where the total parameter number of a standard convolution ($\mathcal{S}_{para}$) is:
\begin{flalign}
\label{paraS}
&& \mathcal{S}_{para} & = C_{out} \times C_{in} \times {K}^2,&
\end{flalign}
While after being replaced by group convolution, the parameter number becomes:
\begin{flalign}
\label{paraG}
&& \mathcal{G}_{para} & = \sum_{i=1}^{m}\Big(C_{out} \times \frac{C_{in}}{g_i} \times K_i^2\Big),&
\end{flalign}
where $K$ represents kernel size, $g$ is the group size, $C_{in}$ and $C_{out}$ are the number of input and output channels, respectively.

Here we shall explore the potential ranges of $g$ and $K$ to ensure a lightweight implementation.
For simplicity, in our pyramid convolution block, we constraint the output channel numbers of each convolution level to be the same, thus the $\mathcal{G}_{para}$ can be expanded to:
\begin{equation}
\begin{aligned}
\label{P_params_r}
\ \ \ \ \ \mathcal{G}_{para} = \frac{C_{out}}{m} \times \frac{C_{in}}{g_1}& \times (K_1)^2  + \\[-1ex]
... &+ \frac{C_{out}}{m} \times \frac{C_{in}}{g_m} \times (K_m)^2.
\end{aligned}
\end{equation}
Based on Eq.~\ref{paraS} and Eq.~\ref{paraG}, we can establish the condition to ensure a lightweight design of the proposed pyramidal group convolution in the following Proposition.

\noindent\textbf{Proposition 1.} \textit{The total parameter number of the proposed pyramidal group convolution can be more lightweight than that of the standard convolution in most cases.}
\begin{proof}
The problem can be interpreted as:
\begin{equation}
\label{OverallObj}
\ \ \ \ \ \ \ \ \ \ \ \ \ \ \ \ \ \ \ \ \ \ \ \ \ \ \ \ \ \ \ \ \ \ \mathcal{G}_{para} \leq \mathcal{S}_{para}.
\end{equation}
We can expand both sides of Eq.~\ref{OverallObj} by using Eq.~\ref{paraS} and Eq.~\ref{paraG}, and it leads to
\begin{equation}
\begin{aligned}
\label{P_params_r1}
\ \ \Big\{\frac{C_{out}}{m} \times& \frac{C_{in}}{g_1} \times (K_1)^2+...\\[-0ex]
&+ \frac{C_{out}}{m}\times \frac{C_{in}}{g_m} \times (K_m)^2\Big\}\\[-0ex]
&\ \ \ \ \ \ \ \ \ \ \ \ \ \ \ \ \ \ \ \ \ \ \ \ \ \ \leq C_{out} \times C_{in} \times K^2,
\end{aligned}
\end{equation}
which can be reduced to
\begin{equation}
\begin{aligned}
\label{P_params_r2}
\ \ \ \ \ \ \ \ \ \ \ \ \Big(\frac{K_1}{K}\Big)^2 \times \frac{1}{g_1}   + ... + \Big(\frac{K_m}{K}\Big)^2 \times \frac{1}{g_m} \leq m,
\end{aligned}
\end{equation}
and, finally, it can be  reached to the following condition:
\begin{equation}
\label{eq:condi}
\ \ \ \ \ \ \ \ \ \ \ \ \ \ \ \ \ \ \ \ \ \ \ \ g_i  > \Big(\frac{K_i}{K}\Big)^2, \ \ \ i\in\{1,2,...,m\}.
\end{equation}
It is easy to see that Eq.~\ref{eq:condi} is a very generic condition which can be ensured easily, showing the proposed pyramidal group convolution can be more lightweight than that of the standard convolution yet with better learning capacity.
\end{proof}

\subsection{Dynamic Routing}
\label{sec:dr}
To enable the proposed DPConv to have similar capabilities as humans---we humans could automatically adjust the focused range of our \textbf{f}ield \textbf{o}f \textbf{v}iew (FOV), we devise the dynamic routing, which routings between different kernel sizes, aiming for a finer control towards the receptive fields.

As shown in Fig.~\ref{detailed_block}, the key idea of our dynamic routing is simple and straightforward, where we utilize the channel-wise attention of the grouped pyramidal convolution’s original input \textbf{X} to dynamically control the routing process, which can be detailed as follows.

We gather the global contextual information with channel-wise statistics by using \textbf{g}lobal \textbf{a}verage \textbf{p}ooling (GAP).
This process embeds the input $\textbf{X}\in\mathbb{R}^{1\times C_{in}\times H\times W}$ to a learnable latent vector $\textbf{Z}\in\mathbb{R}^{1\times C_{in}}$ by performing GAP on $\textbf{X}$ over the spatial dimension, where $C_{in}$ denotes the channel number of the input tensor $\textbf{X}$.
Thus, the $c$-th component of \textbf{Z} can be detailed as follows:
\begin{flalign}
&& \ \textbf{Z}(1,c) & =  \frac{1}{H \times W} \sum_{i=1}^H \sum_{j=1}^W \ \textbf{X}(1,c,i,j). &
\end{flalign}

One could apply \textbf{Z} as the channel-wise attention to the grouped pyramidal convolution directly since the value of each element in \textbf{Z} represents the importance of the corresponding feature slice in \textbf{X}.
However, we shall perform an additional embedding regarding \textbf{Z} via an MLP consisting of two fully-connected layers, a non-linearity ReLU, and a softmax operation.

There are two reasons for this implementation: 1) this MLP enables the latent vector $\textbf{Z}$ to be learnable, making our routing process to be dynamical and completely different to the widely-used static attention mechanism; 2) to ensure a generic nature, this MLP can convert $\textbf{Z}$ to an arbitrary dimension, since there exist cases that, if the grouping process allows overlap between channels, the channel numbers of the grouped pyramidal convolutions is different to the original input.

The dynamic weights $\bm{\alpha}$ for the grouped pyramidal convolution can be formulated as follows:
\begin{flalign}
\label{alpha}
&& \ \bm{\alpha} & = Softmax\bigg(FC\Big(ReLU\big(FC(\textbf{Z})\big)\Big)\bigg), &
\end{flalign}
where $\bm{\alpha}\in\mathbb{R}^{1\times C_{group}}$ and, in our implementation, we set $C_{group} = C_{in}$ for simplicity.

In sharp contrast to the classic residual bottlenecks, which are limited to fixed kernel sizes, we can dynamically route between kernels with different sizes by using $\bm{\alpha}$ as the sole indicator.
The output \textbf{Y} of this dynamic routing process can be represented as follows:
\begin{equation}
\label{finalY_1}
\begin{aligned}
\textbf{Y} = \textbf{X}+Concat\big(\bm{\alpha}_1\times(&\textbf{W}_{1}*\textbf{X}),\\
&\bm{\alpha}_2\times(\textbf{W}_{2}*\textbf{X}),\\
&\ \ \ \ \ \ \ \ ...,\bm{\alpha}_m \times(\textbf{W}_{m}*\textbf{X})\big),
\end{aligned}
\end{equation}
where $*$ represents the convolution operation, and $Concat$ denotes the concatenation operation.

In summary, the final version of our DPConv has three distinguished merits: 1) it provides the flexibility of adaptively choosing different feature scales; 2) it consumes less memory and computational resources; 3) it can replace the conventional convolutional layer directly without additional network modification. 

As a plug-and-play module, the DPConv can be easily combined with ResNet, coined DPResNet.
The most relevant one is Res2Net~\cite{2021res2net}, Fig. \ref{bottleneck} shows an overall comparison, where Res2Net explores multi-scale information of each convolution layer. Our DPResNet is different from Res2Net in both motivation and internal structure.
In contrast, our goal is to generate feature representation according to the input image, and thus this could be more consistent with the real human attention mechanism.

\section{DPConv-matched Decoder}
As previously mentioned, the proposed DPConv addresses the ``scale confusion'' problem (Fig.~\ref{multi_scale}) partly by generating scale-aware feature representation.
Another issue that deserves further investigation is how to use these scale-aware features for the SOD task appropriately.

\begin{figure*}[!t]
\centering
\includegraphics[width=0.95\linewidth]{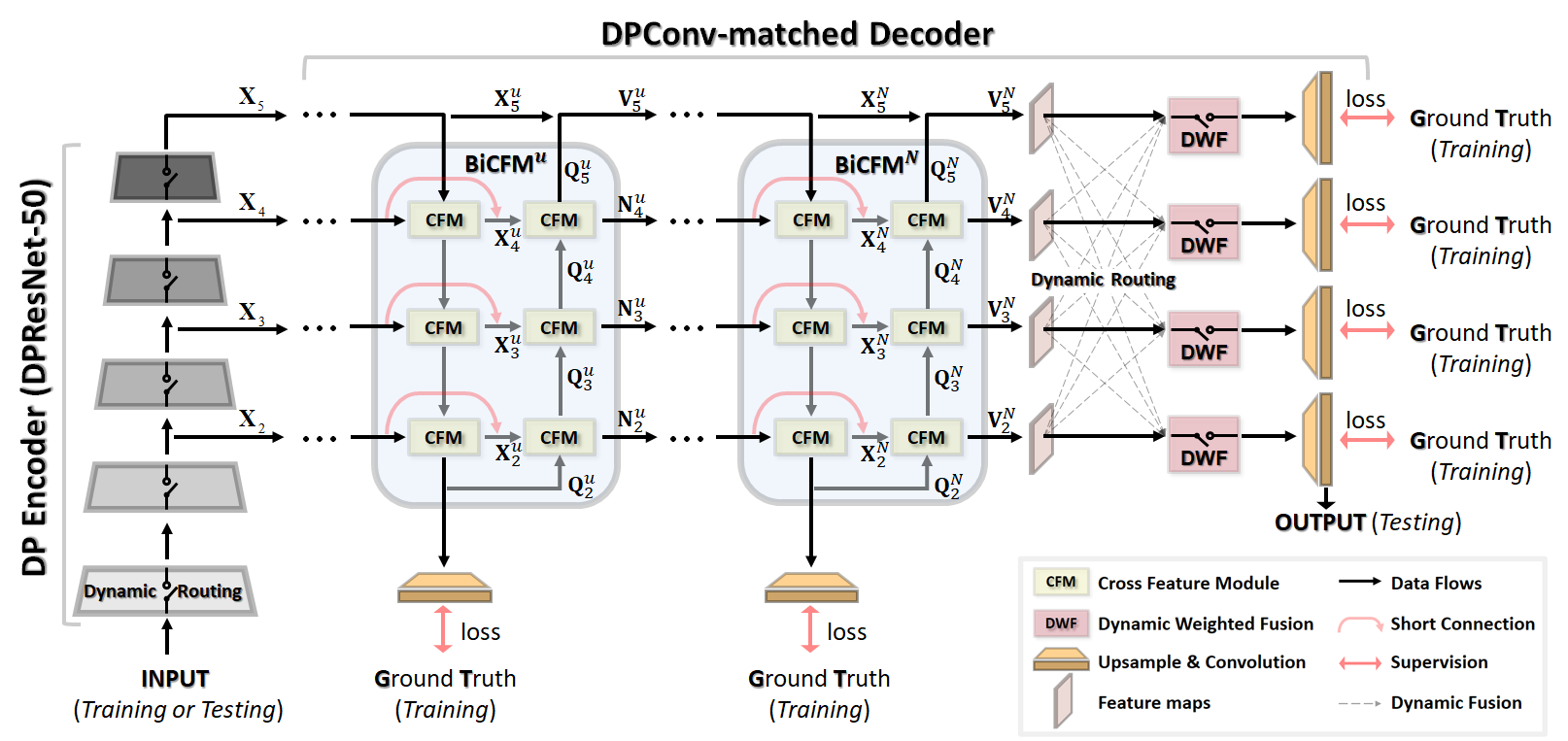}

\caption{The architecture of our DPNet consists of three key modules: 1) DPRestNet-50 generates customized scale-aware features, 2) $N$ BiCFMs are stacked to fuse these features bidirectionally, and 3) the DWF module is applied for dynamic feature collection.
 }
\label{decoder}
\end{figure*}

The main difference between features respectively obtained from DPResNet and other SOTA backbones are that, in SOTA backbones, each convolution layer has its distinct scale information. In contrast, in our DPResNet, multi-scale information of each convolution layer might get mixed.
Therefore, the widely-used network design --- delicately interacting between multi-scale information via static short-connections, is not an optimal choice for the DPConv-based DPResNet.
Instead, we shall consider the feature collection process densely and dynamically.

Here we present the DPConv-matched decoder, whose technical innovations include: 1) \textbf{bi}directional \textbf{c}ross-scale \textbf{f}usion \textbf{m}odule (BiCFM, Sec.~\ref{sec:bicfm}) and 2) \textbf{d}ynamic \textbf{w}eighted \textbf{f}eature (DWF, Sec.~\ref{sec:dwff}).
Fig.~\ref{decoder} shows the architecture overview, and we call this entire network as \textbf{d}ynamic \textbf{p}yramid \textbf{net}work (DPNet), and, clearly, DPNet is made of DPRestNet encoder and DPConv-matched decoder.

\subsection{\textbf{Bi}directional \textbf{C}ross-scale \textbf{F}usion \textbf{M}odule (BiCFM)}
\label{sec:bicfm}

\begin{figure*}[!t]
\centering
\includegraphics[width=0.95\linewidth]{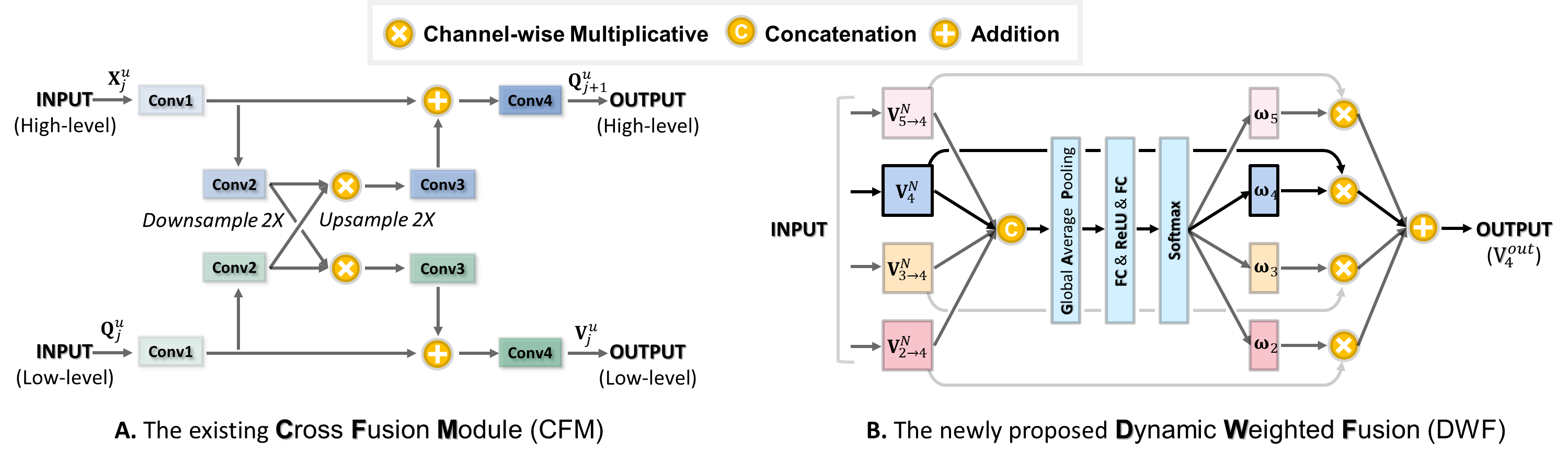}
\vspace{-0.5cm}
\caption{The architecture details of CFM and DWF, where the existing CFM was designed to mitigate the discrepancy between high-level and low-level features, and the newly devised DWF aims to make the final inference period suitable towards those scale-aware features provided by the DPConv-based DPResNet. }
\label{CFM_DWF}
\end{figure*}

Generally, w.r.t. our human visual system, high-level neurons are usually applied for the salient object localization, while those low-level ones are more likely to be activated by local textures and patterns~\cite{zeiler2014visualizing}.
Following this rationale, previous decoders~\cite{liu2018path,BMP} have adopted the top-down path, where the cross fusion module (CFM~\cite{wei2020f3net}, see Fig.~\ref{CFM_DWF}-A) is one of the most representatives.
However, the conventional one-directional feature collection process is not suitable for the DPConv-based DPResNet, since features generated by DPConv are no longer belonging to single scales, where cross-scale information exists in each encoder layer.
Besides, some previous studies (e.g., \cite{liu2018path}) have also indicated that low-level features could also be able to help the localization process.
Thus, we shall consider collecting multi-scale features in a bi-directional manner.

We propose the bidirectional cross-scale fusion module (BiCFM), and its structure can also be found in Fig.~\ref{decoder}.
Unlike the existing top-down fusion modules (e.g., CFM), the proposed BiCFM can collect multi-scale information in bottom-up aggregation paths.
To be more specific, it starts from the penultimate lowest level ($\textbf{X}_2^u$) and gradually approaches $\textbf{X}_5^u$, where $u$ denotes the $u$-th bidirectional structure.

We use $\textbf{V}_i^u,$ and $\textbf{Q}_{i+1}^u$ to denote the newly generated lateral and vertical features respectively corresponding to $ \textbf{X}_i^u$.
In the bottom-up path, each CFM block takes a lateral feature $\textbf{X}_i^u$ and a vertical feature $\textbf{Q}_i^u$ to generate the new feature $\textbf{V}_{i}^u$.
Moreover, we introduce shortcut connections from the original input (e.g., $\textbf{X}_5^u$) to BiCFM's output, which requires no additional computational cost yet leads to better feature representation.
We treat each bidirectional (top-down and bottom-up) path as one building block and repeat the same block multiple times to enable more efficient feature fusion.
Thus, there are $N$ BiCFMs stacked sequentially in our decoder, and the exact choice of $N$ will be explored in Table~\ref{paraN}.

\subsection{\textbf{D}ynamic \textbf{W}eighted \textbf{F}usion (DWF)}

\label{sec:dwff}
A common way for saliency inference could be directly upsampling the multi-scale features to the same size as the input image and then combining them via element-wise summation or concatenation.

Previous decoders~\cite{zhang2017amulet,wang2020progressive} have treated all multi-scale features equally without considering their characteristics.
In our DPResNet, those multi-scale features generated by our DPConv are already scale-aware, some of which are more helpful than others.
Thus, the final inference stage should follow the dynamic routing rationale also, i.e., to automatically select those really valuable features.
Therefore, we propose the \textbf{d}ynamic \textbf{w}eighted \textbf{f}usion (DWF) module, which learns the importance of each of its input.

As shown in Fig.~\ref{CFM_DWF}-B, we take the $4$-th level, for instance, all features in other levels (e.g., $\textbf{V}_j^N$, where $j\not=4$) are resized to an identical size as that of the $4$-th level.
Next, a dynamic weighted fusion operation is introduced to fuse features from different levels selectively, and this process can be detailed as follows:
\begin{equation}
\begin{aligned}
\label{weighted_fusion}
\textbf{V}_4^{out}={\bm{\omega}}_1 \times \textbf{V}_{2\rightarrow4}^N + {\bm{\omega}}_2 \times &\textbf{V}_{3\rightarrow4}^N + \\
&{\bm{\omega}}_3 \times \textbf{V}_{4}^N + {\bm{\omega}}_4 \times \textbf{V}_{5\rightarrow4}^N,
\end{aligned}
\end{equation}
where $\textbf{V}_{i \rightarrow j}^N$ denotes the features resized from level $i$ to level $j$, and $\bm{\omega}\in\mathbb{R}^{1\times C_{out}}$ is a learnable weight vector, which follows an identical rationale to $\bm{\alpha}$ mentioned in Eq.~\ref{alpha}, and $\bm{\omega}$ is defined as:
\begin{equation}
\begin{aligned}
\label{weighted_fusion1}
&Softmax\bigg(FC\Big(ReLU\big(FC({\textbf{P}})\big)\Big)\bigg)\rightarrow \bm{\omega},\\[-2.5ex]
&\ \ \ \ \ \ \ \ \ \ \ \ \ \ \ \overbrace{GAP\big(Concat(\textbf{V}_{2\rightarrow4}^N, \textbf{V}_{3\rightarrow4}^N, \textbf{V}_4^N, \textbf{V}_{5\rightarrow4}^N)\big)}^{\Uparrow}
\end{aligned}
\end{equation}
where $GAP$ is the \textbf{g}lobal \textbf{a}verage \textbf{p}ooling operation; ${\bm{\omega}}_i$ is normalized via softmax, which represents the importance of each input channel. Features in other levels proceed in a similar way. This method dynamically aggregates the features at all levels toward the given SOD objective.

\subsection{DPNet vs. SOTA SOD Models}
\label{differences_text}

To highlight our innovations, we shall provide an in-depth discussion regarding the relationship between our DPNet and the existing SOTA SOD models from the perspective of encoder and decoder designs.

W.r.t. the encoder, most of the existing SOTA models have directly adopted the off-the-shelf ResNet as their feature backbones, where the ResNet was designed for image recognition tasks rather than dense prediction tasks (e.g., our SOD task).
Instead of using the off-the-shelf ResNet directly, we propose the DPConv-based DPResNet, which could alleviate the scale confusion problem by providing scale-aware feature representation.

In view of the decoder's design, ITSD20~\cite{wang2019an} and BMP18~\cite{BMP} are the two most relevant works to our DPNet.
Beyond using a bidirectional architecture simply as BMP18, our DPNet has additionally considered the discrepancy between features of different scales via the proposed BiCFM, which simultaneously takes advantage of low-level features in detail retaining and high-level features in localization.

On the other hand, unlike ITSD20, which treats all its intermediate features equally in the final inference period, our model has designed a novel dynamic weighted fusion module to route between the scale-aware features provided by DPResNet selectively.
And this dynamic process makes our encoder suit the proposed DPResNet well while being more consistent with the real human visual system.

\section{Experiments}
\label{experiments}

\begin{table*}[!t]
\centering
\LARGE{
\linespread{2}
\renewcommand\arraystretch{1.2}
\setlength\tabcolsep{2pt}
\resizebox{1\textwidth}{!}{
\begin{tabular}{l||ccccc|ccccc|ccccc|ccccc|ccccccc}
\Xhline{1pt}
\multirow{2}{*}{Methods}
& \multicolumn{5}{c|}{DUT-OMRON}
& \multicolumn{5}{c|}{DUTS-TE}
& \multicolumn{5}{c|}{ECSSD}
& \multicolumn{5}{c|}{HKU-IS}
& \multicolumn{5}{c}{PASCAL-S}
\\
\cline{2-26}
 &$F_\beta$  & MAE$$ &W-$F_\beta$  & S-m$$ & E-m$$   
&$F_\beta$  & MAE$$  &W-$F_\beta$  & S-m$$ &E-m$$   
&$F_\beta$  & MAE$$  &W-$F_\beta$& S-m$$  &E-m$$  
&$F_\beta$  & MAE$$  &W-$F_\beta$  & S-m$$ &E-m$$  
&$F_\beta$  & MAE$$ &W-$F_\beta$ & S-m$$ &E-m$$ \\
\hline\hline

\textbf{DPNet(Ours)} 
 & \textcolor{red}{.837} &    \textcolor{blue}{.049} &   \textcolor{red}{.770}  & \textcolor{red}{.853}  & \textcolor{red}{.876} 
&  \textcolor{red}{.916}   &  \textcolor{red}{.029} & \textcolor{red}{.849} & \textcolor{red}{.912}  & \textcolor{red}{.921} 
&  \textcolor{red}{.958}  &   \textcolor{red}{.028} &  \textcolor{red}{.935} & \textcolor{red}{.937} &  \textcolor{blue}{.944 }
& \textcolor{red}{.954}  &  \textcolor{red}{.023} & \textcolor{red}{.920} & \textcolor{red}{.934}  & \textcolor{blue}{.942} 
&   \textcolor{red}{.912}   &  \textcolor{red}{.072} & \textcolor{red}{.817}  & \textcolor{red}{.856} &  \textcolor{red}{.867} \\

RSBNet22 \cite{ke2022recursive}
& .801   &  \textcolor{red}{.046}   & \textcolor{blue}{.754}  & .835  &{.861}
& .889  &  .035    & {.828}  & .879  & \textcolor{blue}{.917}
& .944   &  .033   & {.915}  & .922  & \textcolor{red}{.951}
& .938   &  .027   & \textcolor{blue}{.908}  & .919  & \textcolor{red}{.955}
& .897   &  \textcolor{blue}{.077}   & .795  & .836  &{.851} \\
PFSNet21  \cite{ma2021PFSNet} 
& {.823}   &  .055  & .742 & .843  &   .813
&   {.895}  &   {.036} & .819 & {.892} &   .873 
&  \textcolor{blue}{.952}  &   \textcolor{blue}{.031} &  .914 & \textcolor{blue}{.930} & .919 
&  \textcolor{blue}{.943}  &   \textcolor{blue}{.026} & .903  & \textcolor{blue}{.924}&   .919 
&  .899 &   {.079} & \textcolor{blue}{.798} & .843 & .848 \\

CTDNet21 \cite{zhao2021complementary}

& \textcolor{blue}{.824} & .052  &  .746 &\textcolor{blue}{.844}  &.825 
& \textcolor{blue}{.897} & {.034}  &  .824 &{.892}  &.888
& .950 & .032  &  .910 & 924  &.924
& .941 & .027  &   .903 & .921& .928
& \textcolor{blue}{.901} & .080  &  .796 &.841  &.846 \\

PSGL21 \cite{yang2021progressive} 
& .811 &  .052  &  .742 & .833 & .846
& .885 &  .036  &  .819 & .883  & {.902}
& .949 &  .032  &   {.915}  & .925  & .939
& .938 &  .027  &  .903  & .920  & .939
& .896 &   .081 &   .793 & .837  & {.851} \\

SAMNet21 \cite{liu2021samnet} 
& .803 & .065 & .648  & .830  &   .688 
&   .836 & .058 &   .682 & .848 &   .712  
&   .928 & .050 &  .834 &  .907 &  .801 
 & .915 & .045 &   .811 & .898 & .786 
&   .857 & .113 &  .724 & .803 &  .765 \\

 JCS21 \cite{li2021uncertainty}
& .822  & .051 & .752& .843  & \textcolor{blue}{.864} 
& .894  & \textcolor{blue}{.032} & \textcolor{blue}{.832} & \textcolor{blue}{.899}  & .915
& .945  & .030 & \textcolor{blue}{.919} & \textcolor{blue}{.930}  & .940 
& .936  & .026 & .904 & .923  & .913 
& .900  & .082 & .795 & \textcolor{blue}{.850}  & \textcolor{blue}{.854}  \\

GCPANet20 \cite{chen2020global} 
& .812 &   .056 &  .708  & {.839}  & .779 
&   .887 &   .038 &    .787  & {.890}&  .835
&  {.949} &   .035 &   .890  & {.926} & .887  
 & {.938}  &   .031 & .875 & {.921}  &   .876 
&   {.898}  & \textcolor{blue}{.077} &   {.790} & {.847} &   {.833} \\

GateNet20 \cite{zhao2020suppress} 
 & .812 &   .056 & .698 & {.838} &   .767
&   .887 &   .038 &     .771& .885 & .816 
&  {.949} &   .035 &   .882 & .920 &   .881 
& {.938}  &   .031&   .866 & .915 & .873 
&   {.898}  & \textcolor{blue}{.077}&     .760 & .831 &   .818 \\

MINet20 \cite{pang2020multi} 
& .810  & .056 & {.719} & .833 &  {.798}   
&   .884 & .038 & {.797}& .883 & {.852}   
&.943  & {.034} &   {.903} & {.925} &  {.906} 
&   .935 & .029  &   {.888} & {.919} &  {.901} 
 &.889  & .083 &    .779   & .833 & {.835}\\

F3Net20 \cite{wei2020f3net}
& .813 & {.053} & {.720} & {.839} &  {.803} 
& {.891} & {.036} &    {.801} & {.888} &  {.864} 
 &  .945 &   {.033}  &  {.902} & .924 &  {.915}  
 & .937&   {.028} &     {.890} & .917 &  {.912}  
 & {.895} &   .080 &   {.789} & {.840} & {.843} \\

{RANet20}~\cite{RANet20}
& .799 & .058 &   .671& .825& .742
&  {.874} & .044 & .743&.874 & .776
& {.941} & .042 & .866&.917 & .844 
&{.928}  & .036 & .846&.908 & .841
&  {.866}   &{.078} &.757& {.847} & .812\\

CPD19~\cite{CPD}  
 &    .797   &  .056 &  .705 & .825 & {.786} 
&   .865 &   .044  & .769 & .868 &.838 
& .939 & .037 & .889& .918& {.902}
&   .925 & .034  & .866 &.906 & .888 
&     .884 & .089 & .771 &.828 & .827 \\

PoolNet19~\cite{PoolNet} 
& .805 & {.054} & .696 & .831& .775
&   .889 & {.037} & .775 &{.886} & .819 
&   {.949} & .035 & .890 & {.926} & .877
&   .936 & .030 & .873 & {.919} & .870
& .902 & .081 & .781 &{.847} & .826 \\

AFNet19~\cite{AFNet}
  & .797  & .057 & .690 & .826 & .760
 &   .862 & .046  & .747 & .866 & .785
 & .935 & .042 & .867 &.914 & .849
 &   .923 & .036 & .848 & .905 & .839
 &   .885 & .086 & .772 &.833 & .810 \\

EGNet19~\cite{zhao2019egnet} 
 & .815 & {.053} & .701 & {.841}& .760 
 & .888 &{.040} & .769 &{.886} & .802 
& {.947} & {.037} & .887 & {.925}& .870 
& {.935} & {.031} & .870 &{.918} & .860 
& .891 & .090 & .777 & .835& .821\\

BASNet19 \cite{qin2019basnet}
& .805  & .057 & .752 & .836 & .857

& .859  & .048 & .793  & .865 & .886

& .942  & .037 &  .904& .916   &.938

& .929  & .032 &.889  & .909   & .936

& .876  & .092 & .776  & .819 & .834\\

PFAN19 \cite{zhao2019pyramid}
&  .808 & .051 & .753 & .841 & .864
& .870 & .041 & .787 & .874 & .825
& .922 & .047 & .856 & .904 & .857
&  .926 & .035 & .892 & .913 & .938
& .891 & .077 & .781 & .841 &  .838\\

\Xhline{1.4pt}
 \end{tabular}  }}
  \vspace{-0.5em}
\caption{ Quantitative comparisons between the proposed DPNet and SOTA models. Instead of copying and pasting the performance score from the original paper, we evaluate these SOTA models by using the same [\href{https://github.com/ArcherFMY/sal_eval_toolbox}{Code}] with the authors provided saliency maps (except for the MRNet20). {These saliency maps are available at our [\href{https://github.com/wuzhenyubuaa/DPNet}{Github}].} }
\label{results_tab2}
\end{table*}

\begin{figure*}[t]
\centering

\resizebox{1\linewidth}{!}{
\subfigure{
\begin{minipage}[b]{0.302\linewidth}
\includegraphics[width=1\linewidth]{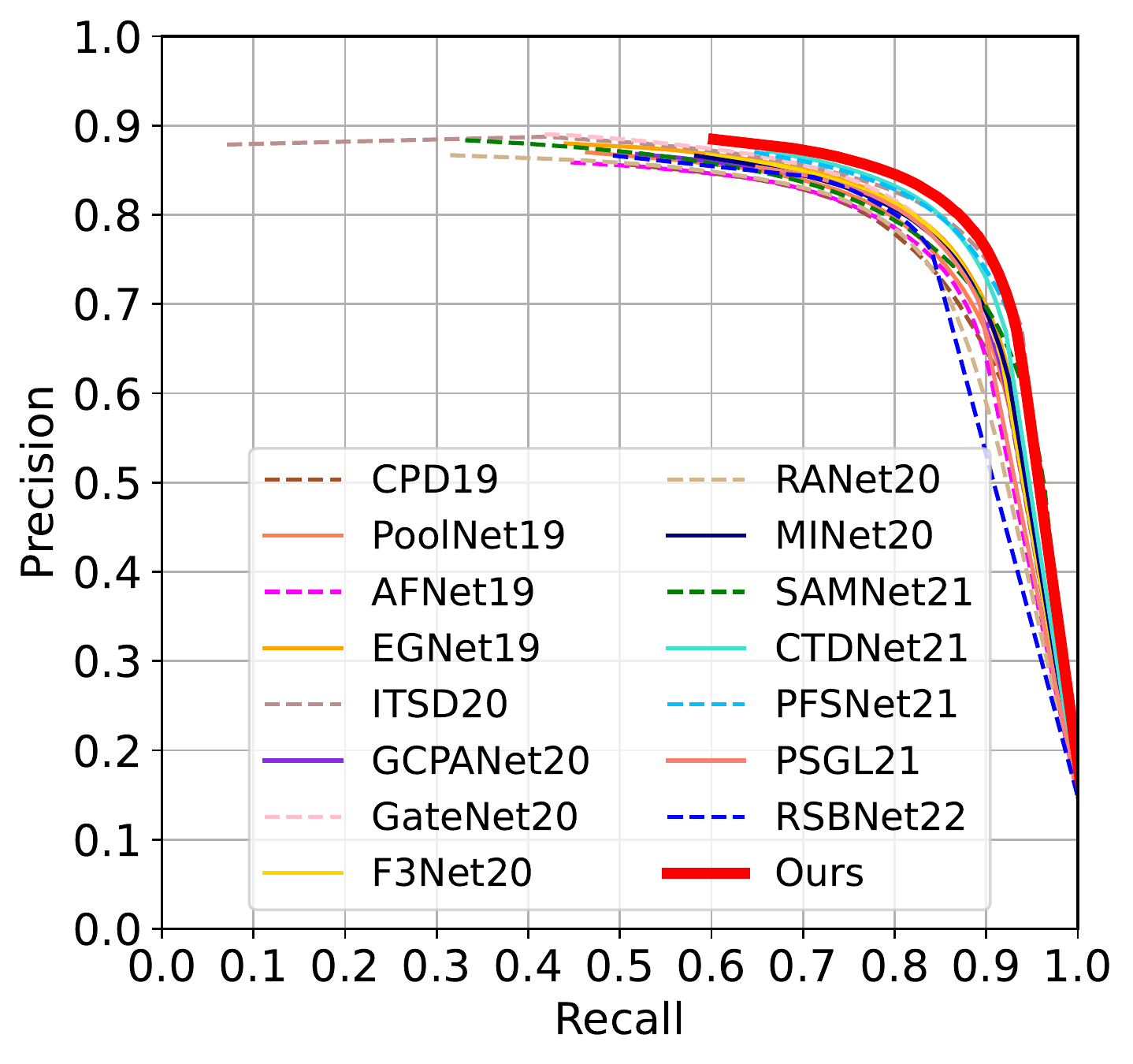}
\end{minipage}}\hspace{0.1pt}\hspace*{-0.8em}
\subfigure{
\begin{minipage}[b]{0.302\linewidth}
\includegraphics[width=1\linewidth]{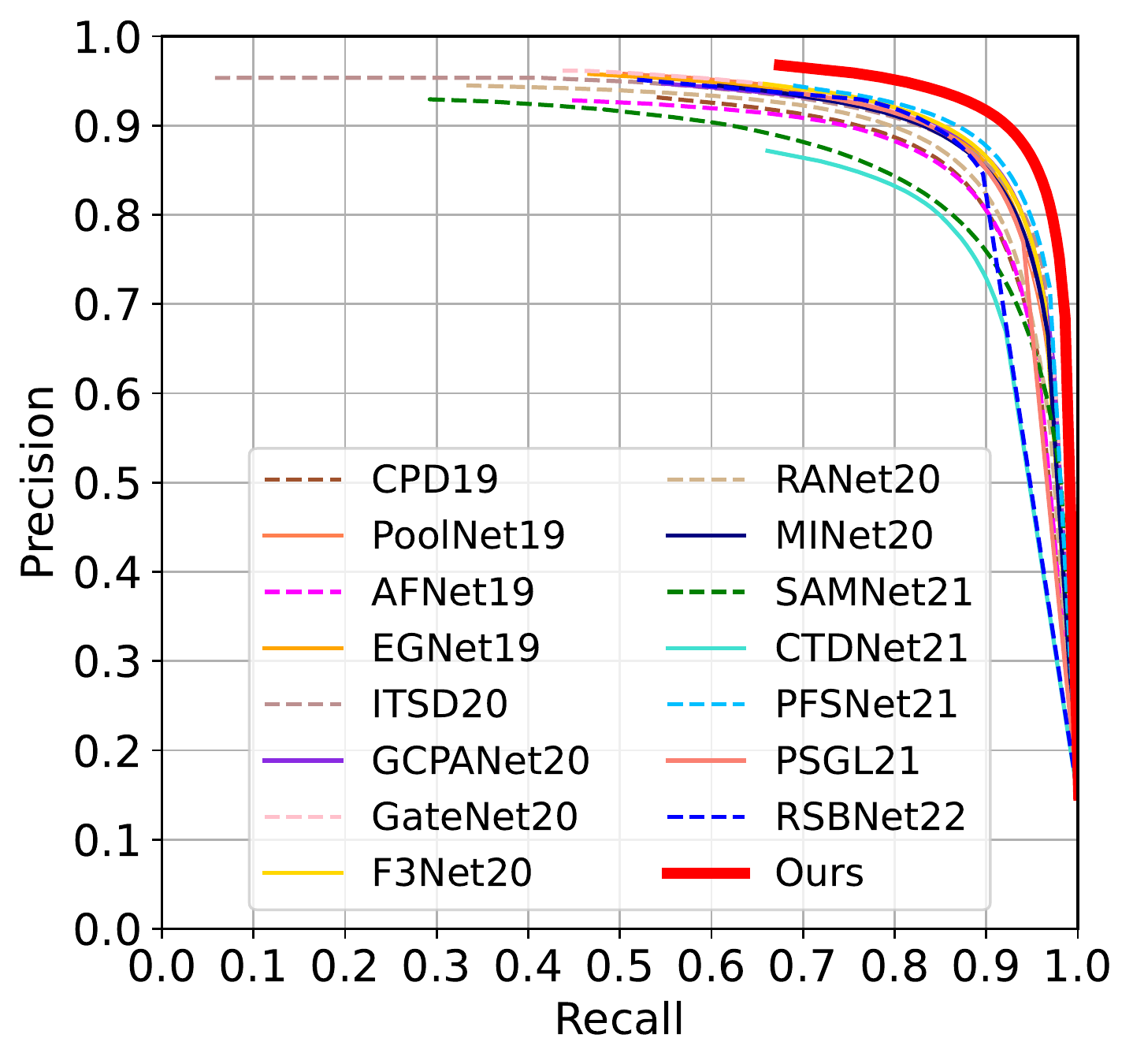}
\end{minipage}}\hspace{0.1pt}\hspace*{-0.8em}
\subfigure{
\begin{minipage}[b]{0.302\linewidth}
\includegraphics[width=1\linewidth]{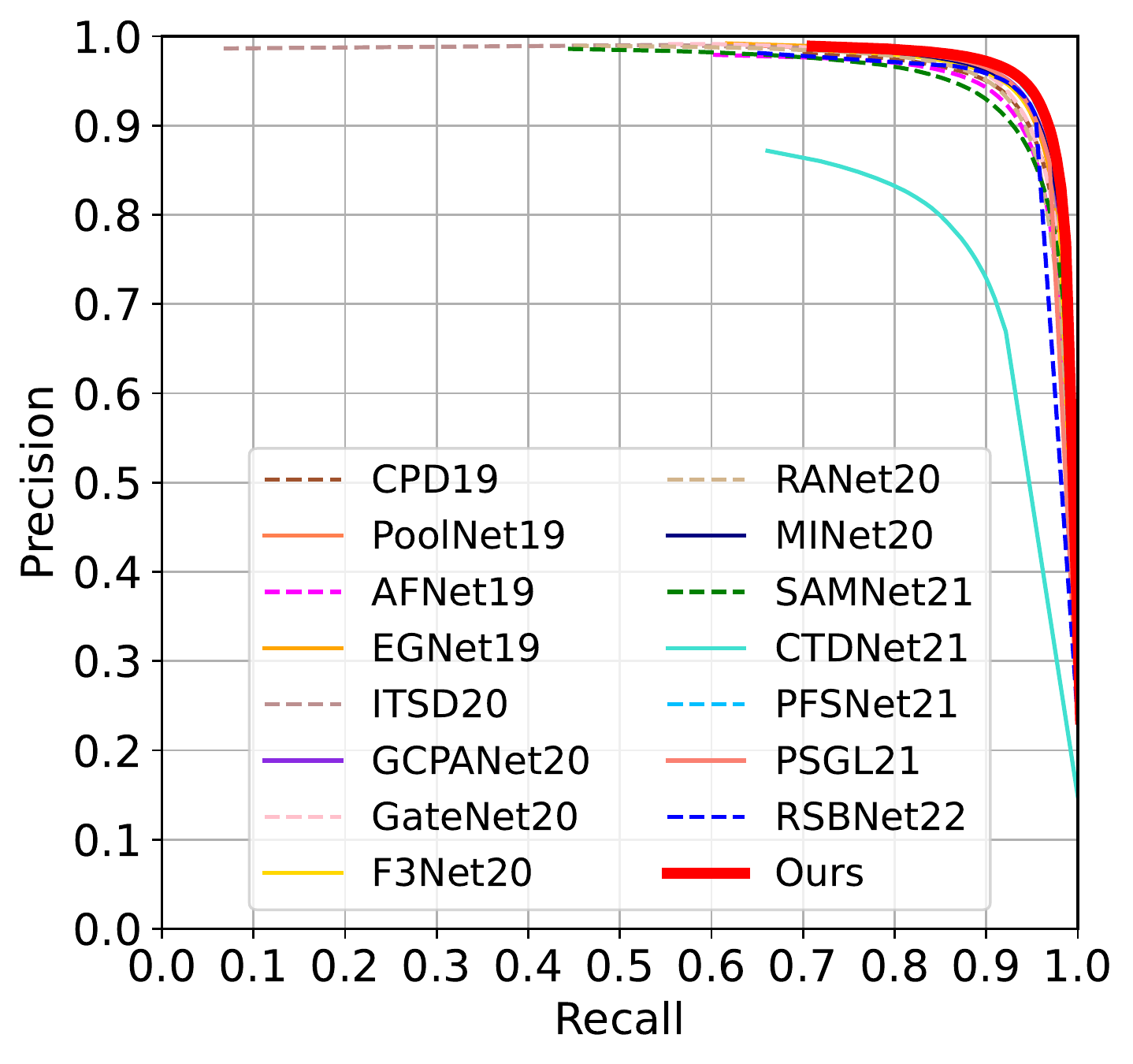}
\end{minipage}}\hspace{0.1pt}\hspace*{-0.8em}
\subfigure{
\begin{minipage}[b]{0.302\linewidth}
\includegraphics[width=1\linewidth]{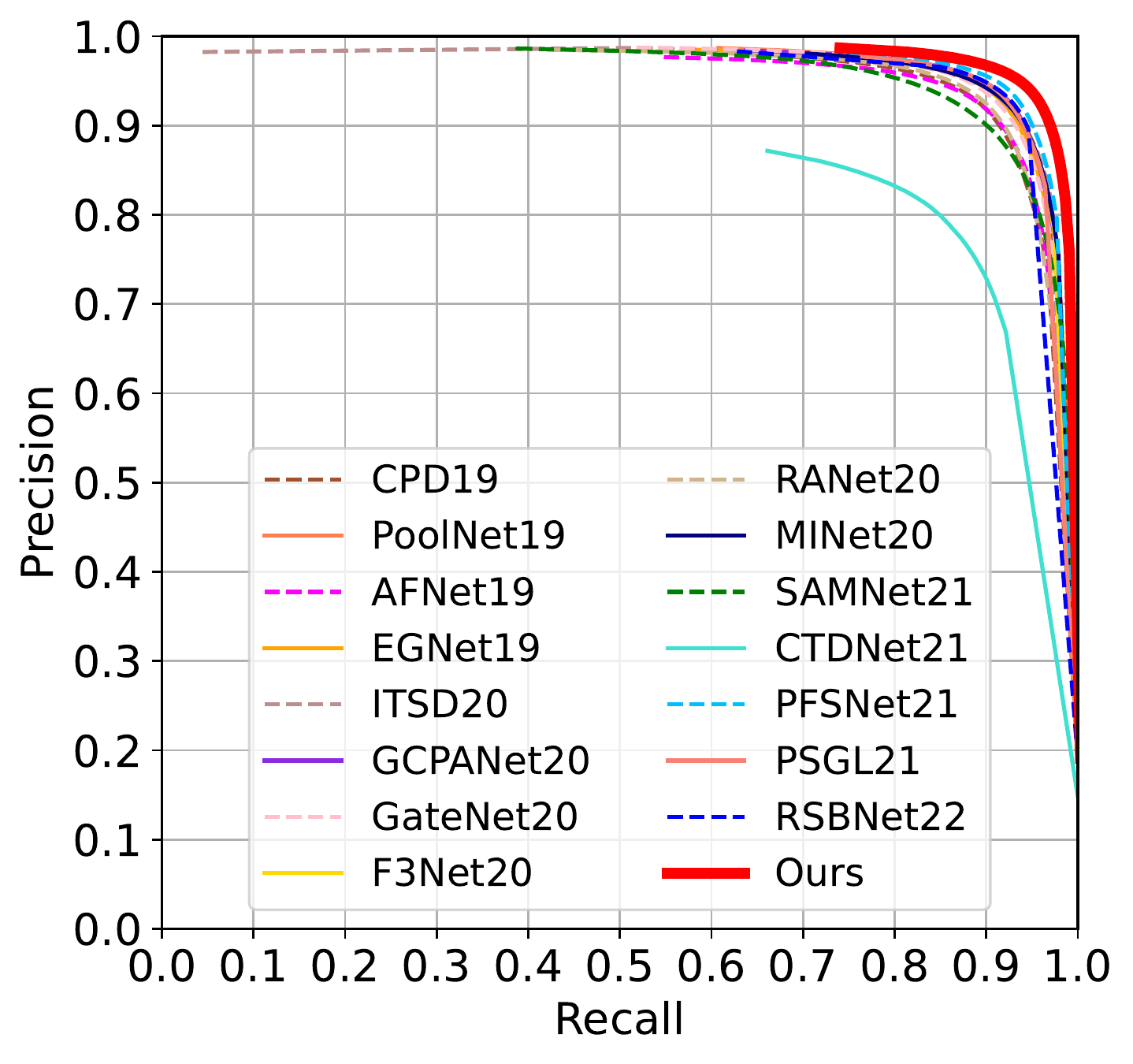}
\end{minipage}}\hspace{0.1pt}\hspace*{-0.8em}
\subfigure{
\begin{minipage}[b]{0.302\linewidth}
\includegraphics[width=1\linewidth]{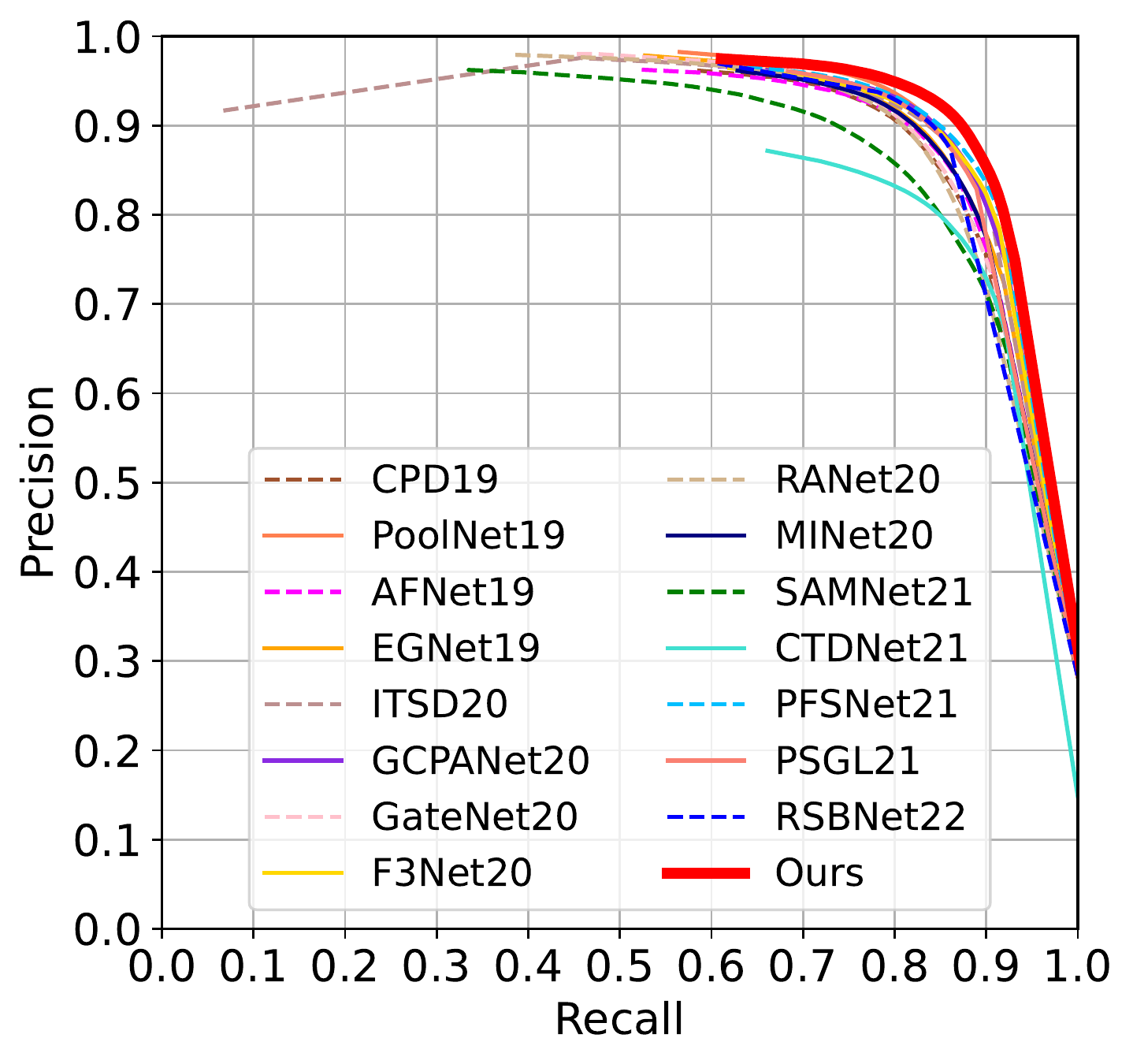}
\end{minipage}}\hspace{0.1pt}\hspace*{-0.8em}

}\vspace{-0.2cm}

\resizebox{1\linewidth}{!}{
\subfigure[\Large DUTS-OMRON]{
\begin{minipage}[b]{0.3\linewidth}
\includegraphics[width=1\linewidth]{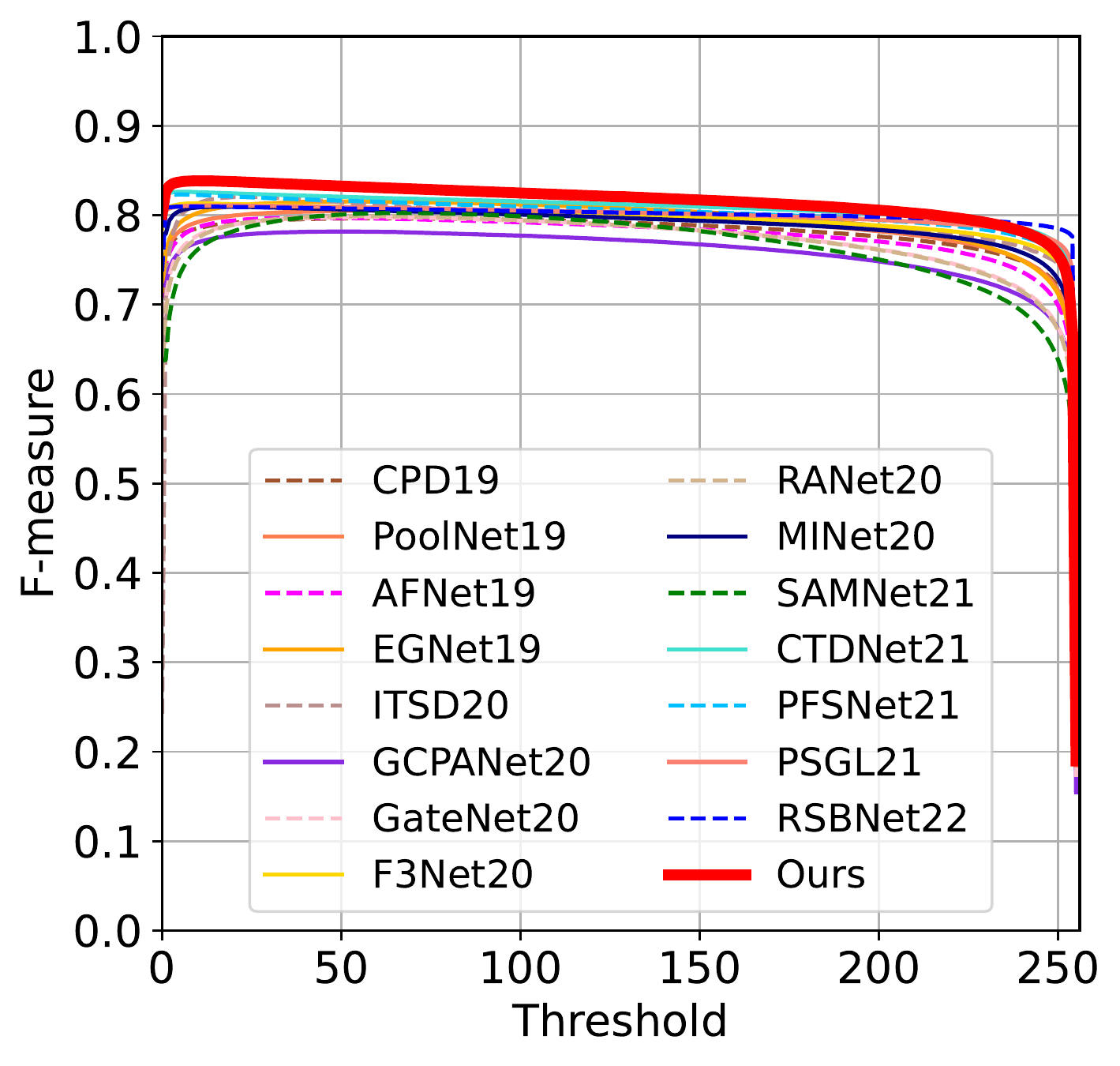}
\end{minipage}}\hspace{0.1pt}\hspace*{-0.5em}
\subfigure[\Large DUTS-TE]{
\begin{minipage}[b]{0.3\linewidth}
\includegraphics[width=1\linewidth]{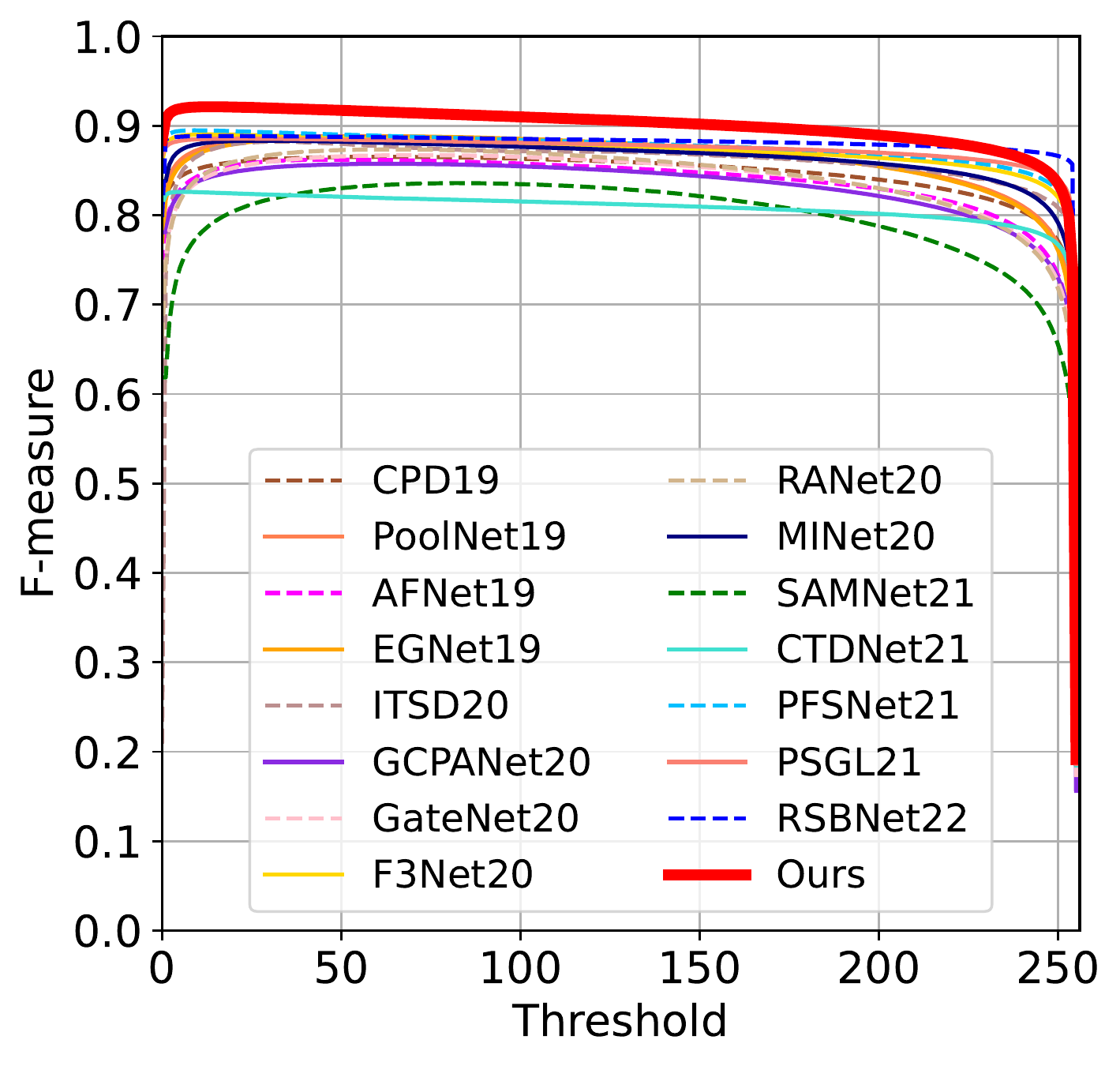}
\end{minipage}}\hspace{0.1pt}\hspace*{-0.5em}
\subfigure[\Large ECSSD]{
\begin{minipage}[b]{0.3\linewidth}
\includegraphics[width=1\linewidth]{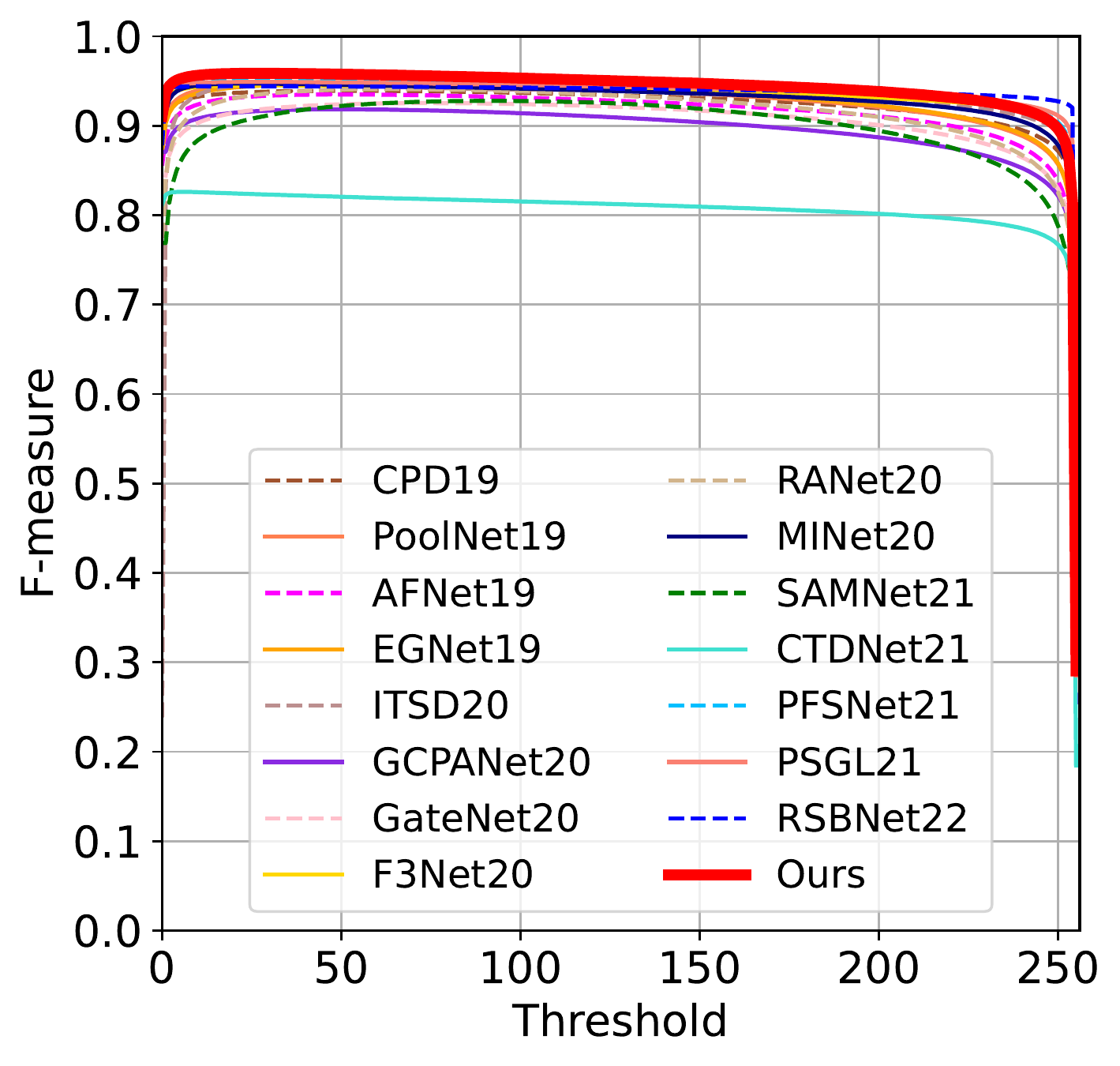}
\end{minipage}}\hspace{0.1pt}\hspace*{-0.5em}

\subfigure[\Large HKU-IS]{
\begin{minipage}[b]{0.3\linewidth}
\includegraphics[width=1\linewidth]{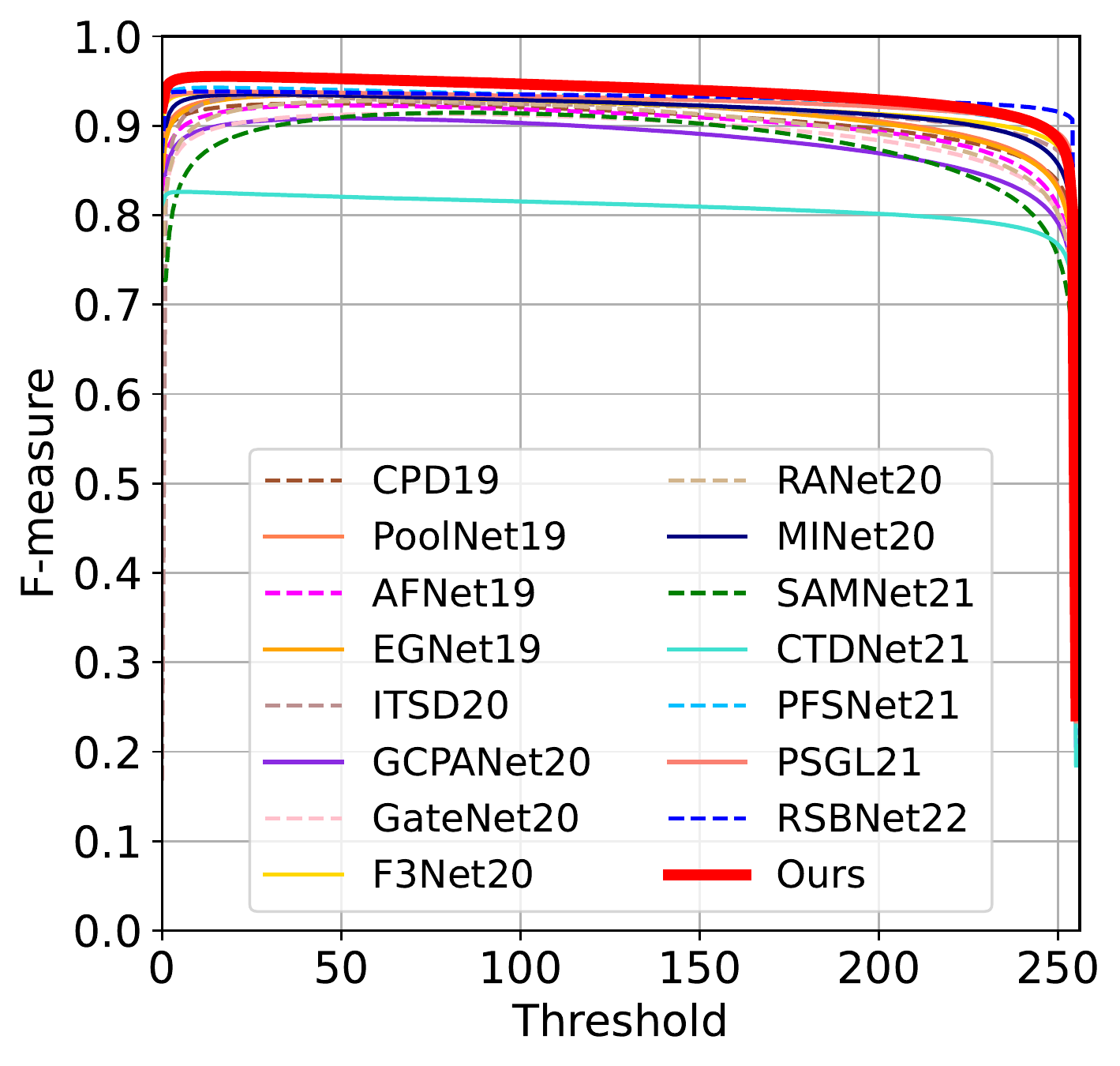}
\end{minipage}}\hspace{0.1pt}\hspace*{-0.5em}
\subfigure[\Large PASCAL-S]{
\begin{minipage}[b]{0.3\linewidth}
\includegraphics[width=1\linewidth]{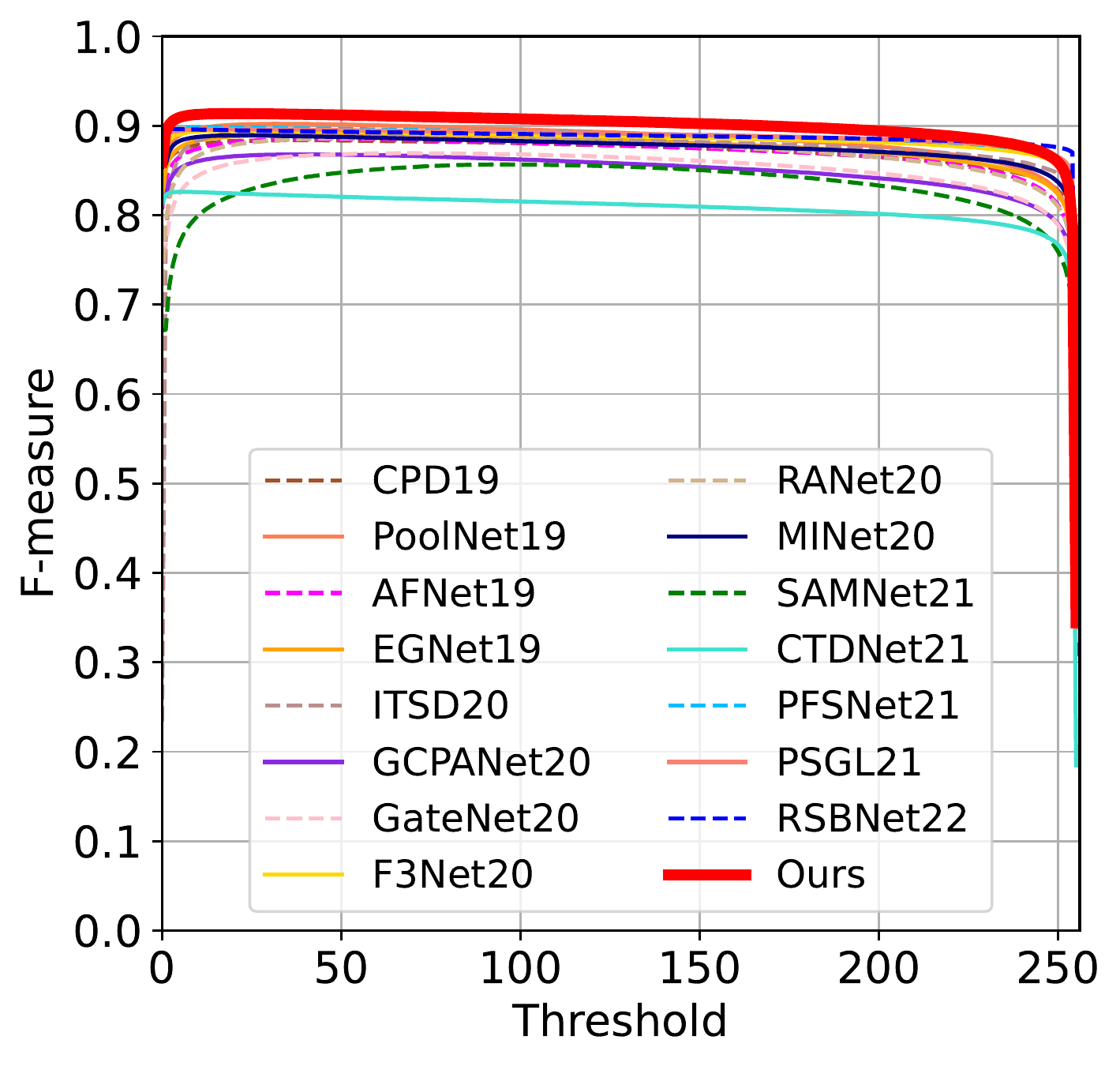}
\end{minipage}}\hspace{0.1pt}\hspace*{-0.5em}

}\vspace{-0.4cm}

\caption{The 1st and 2nd columns show the performance of the proposed DPNet with other SOTA models in terms of PR and F-measure curves respectively. The last column shows the average precision, recall, and F-measure scores.\vspace{-0.4cm}}
\label{PRcurves}
\end{figure*}

\subsection{Implementation Details}

Following the existing SOTA methods \cite{PiCANet,PAGRN,wei2020label,wei2020f3net}, we use the DUTS-TR \cite{wang2017learning} as the training set, and other sets are utilized to evaluate the proposed network. The proposed DPResNet first trained on the ImageNet, and then finetuned on the DUTS-TR dataset.
During the training stage, random horizontal flipping, random crop, and multi-scale input images were used as data augmentation to avoid over-fitting.
During the testing stage, images are resized to $352 \times 352$ and then fed into the proposed DPNet to obtain predictions ``\emph{without}'' any other post-processing (e.g., CRF). We use Pytorch to implement our model. Mini-batch stochastic gradient descent is used to optimize the whole network, and the batch size, momentum, and weight decay are assigned to \{32, 0.9, 5e-4\} respectively.
We use the warm-up and linear decay strategies with a maximum learning rate 0.005 for the DPResNet backbone and 0.05 for other parts to train our model and stop training after 32 epochs on an NVIDIA GTX 2080 GPU.

Following the previous work~\cite{wei2020f3net}, we use weighted cross entropy loss ($\mathcal{L}_{wbce}$) and weighted IoU loss ($\mathcal{L}_{wiou}$) to measure the residuals between predicted saliency maps and ground truths.
Besides, the widely-used multi-level supervision is applied as auxiliary losses. Our decoder consists of $N$ BiCFMs with $M$ levels in total, thus the total loss function ($\mathcal{L}_{total}$) is calculated as follows:
\begin{equation}
\begin{aligned}
\mathcal{L}_{total}  =  \frac{1}{N} \sum_{i=1}^{N}(\mathcal{L}_{wbce}^i&+\mathcal{L}_{wiou}^i)\\[-2ex]
&+ \sum_{j=2}^{M} \frac{1}{2^{j-1}} (\mathcal{L}_{wbce}^j + \mathcal{L}_{wiou}^j),
\end{aligned}
\end{equation}
where $M=5$ and the exact value of $N$ will be determined via ablation study.

\subsection{Datasets and Evaluation Metrics}

We have evaluated the performance of the proposed method on six commonly-used benchmark datasets, including DUTS-TE~\cite{wang2017learning}, DUT-OMRON ~\cite{yang2013saliency},  ECSSD~\cite{ecssd}, HKU-IS~\cite{zhao2015saliency}, and PASCAL-S~\cite{li2014secrets}.

Meanwhile, five widely-used metrics have been adopted to evaluate our method, including the \textbf{P}recision-\textbf{R}ecall (PR) curves, the F-measure curves, \textbf{M}ean \textbf{A}bsolute \textbf{E}rror (MAE), weighted F-measure \cite{w-fmeasure}, S-measure~\cite{Smeasure}  and E-measure~\cite{fan2018enhanced}.

\subsection{Comparison with the SOTA Models}

We have compared our proposed DPNet with SOTA models, including RSBNet22 \cite{ke2022recursive}, CTDNet21 \cite{zhao2021complementary}, PSGL21 \cite{yang2021progressive}, PFSNet21~\cite{ma2021PFSNet}, SAMNet21~\cite{liu2021samnet}, GCPANet20~\cite{chen2020global}, MINet20~\cite{pang2020multi},  F3Net20~\cite{wei2020f3net},  {RANet20}~\cite{RANet20}, {MRNet20}~\cite{MRNet20}, GateNet20~\cite{zhao2020suppress}, EGNet19~\cite{zhao2019egnet}, CPD-19~\cite{CPD}, PoolNet19~\cite{PoolNet}, AFNet19~\cite{AFNet}. 
Instead of copying and pasting the performance score from the original paper, we evaluate these SOTA models by using the same \href{https://github.com/ArcherFMY/sal_eval_toolbox}{Code} with the authors provided saliency maps (except for the MRNet20).

\begin{figure*}[t]
\centering
\includegraphics[width=\linewidth]{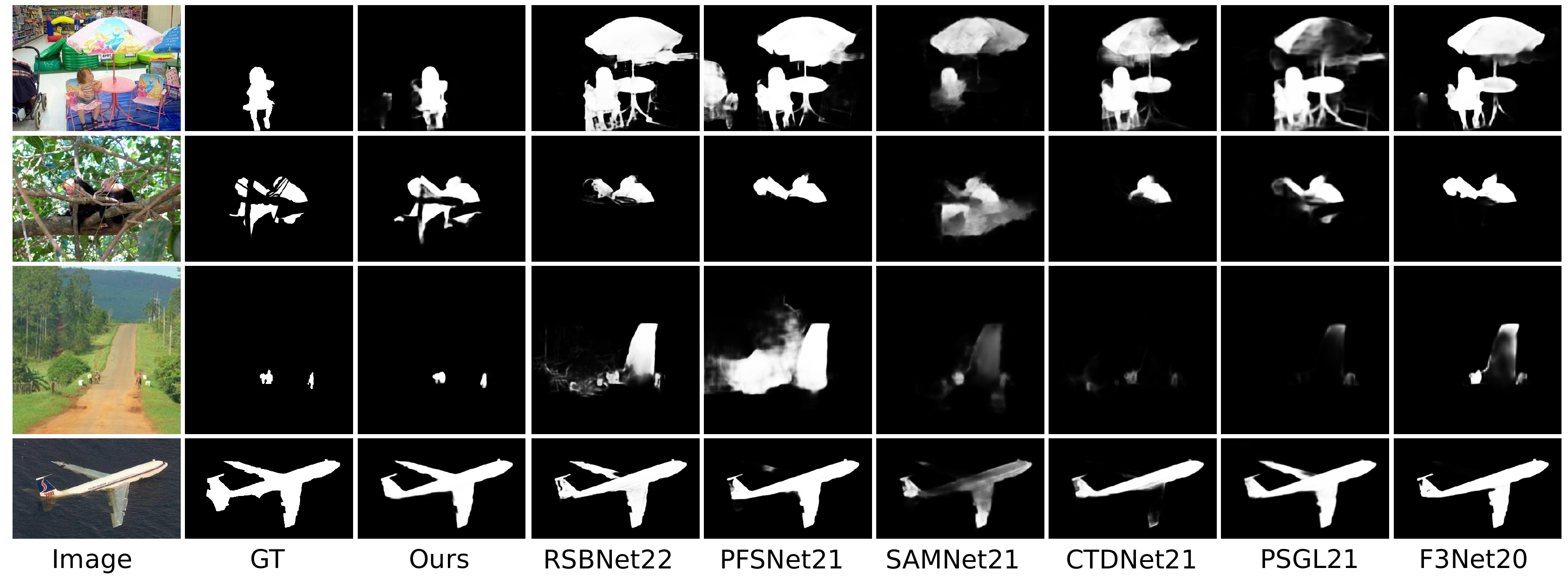}
\vspace{-0.6cm}
\caption{Visual comparison of the proposed DPNet with other SOTA SOD models, where saliency maps produced by our model are more accurate and complete in various challenging scenarios.\vspace{-0.2cm}}
\label{visual_comparsion}
\end{figure*}

\subsubsection{Quantitative Comparisons}

To demonstrate the effectiveness of the proposed DPNet, we compare the DPNet with SOTA methods in terms of five metrics in Table~\ref{results_tab2}.
As can be seen, the proposed DPNet consistently outperforms all other competitors in terms of different metrics on all datasets, even though some methods \cite{zhao2019egnet,zhao2021complementary} have used edge information as the extra supervision.
In particular, in terms of $F_\beta$ and MAE, the performances are improved by $1.9\%$ and $5\%$ respectively over the second best method PFSNet21~\cite{ma2021PFSNet} on DUTS-TE.

Further, we have also provided the standard PR and F-measure curves in Fig.~\ref{PRcurves}. From these curves, we can observe that DPNet consistently outperforms all other models under different thresholds, showing that the proposed method is more robust than other models even on challenging datasets such as DUT-OMRON~\cite{yang2013saliency}.

\subsubsection{Visual Comparison}

To further illustrate the advantages of the proposed DPNet, we provide some representative examples of our model along with other SOTA models for visual comparisons. As shown in Fig.~\ref{visual_comparsion}, our DPNet can handle various challenging scenarios, including cluttered background (1st row), occlusion (2nd row), small objects (3rd row) and underwater objects (4th row).
Compared with other competitors, the saliency maps generated by our DPNet are more accurate.

\subsubsection{Integrating DPConv into SOTA Models}

Since the proposed DPConv block doesn't need a specific network structure, we can easily integrate it into the existing SOTA SOD networks without special adjustment, where we take four most representative ones here, i.e., F3Net20~\cite{wei2020f3net}, MINet20~\cite{pang2020multi}, CPD19~\cite{CPD}, and PoolNet19~\cite{PoolNet}.
The corresponding models are referred to as F3Net20-DP, MINet20-DP, CPD19-DP and PoolNet19-DP, respectively.
For a fair comparison, we retrain those models by using the original codes with the same configurations instead of directly using the generated saliency maps provided by the authors.


\begin{table}[t]
\centering
{
\linespread{2}
\renewcommand\arraystretch{1.1}
\resizebox{0.9\textwidth}{!}{
\begin{tabular}{l||ccc|ccccc}
\Xhline{1pt}
\multirow{2}{*}{Method}
& \multicolumn{3}{c|}{DUT-OMRON}
& \multicolumn{3}{c}{DUTS-TE}
\\
\cline{2-7}
 &max$F_\beta$  & S-m$$ & MAE$$   
  &max$F_\beta$  & S-m$$ &MAE$$    
  \\
\hline\hline

F3Net20 \cite{wei2020f3net} & .813 & .839  &  \textbf{.053}&  {.891} &.888  & {.036} \\

\rowcolor{gray!30}\textbf{F3Net20-DP} & \textbf{.828}&   \textbf{.845} &  .055 &   \textbf{.899} &  \textbf{.896} &  \textbf{.035}  \\
\hline
MINet20 \cite{pang2020multi} & .803 &  .830 & .056 & .887 & .885 & .037 \\

\rowcolor{gray!30} \textbf{MINet20-DP} & \textbf{.816} &  \textbf{.839} &  \textbf{.051} &  \textbf{.896} &  \textbf{.890} &  \textbf{.034} \\
\hline
PoolNet19~\cite{PoolNet} & .796 & .825 & .055 & .874 & .876 & .041 \\

\rowcolor{gray!30} \textbf{PoolNet19-DP} & \textbf{.809} &  \textbf{.836} &  \textbf{.054} &  \textbf{.887}  & \textbf{.878} &  \textbf{.038}  \\
\hline

CPD19~\cite{CPD}  &  .787 & .799 & .054&  .854 & .839&  .045 \\
\rowcolor{gray!30} \textbf{CPD19-DP} & \textbf{.790} &  \textbf{.816} &  \textbf{.053}  & \textbf{.864} &  \textbf{.853} &  \textbf{.044} \\

\Xhline{1pt}
 \end{tabular}  }}
  \vspace*{-0.3em}
\caption{Performance of DPConv based SOTA models in terms of F-measure, S-measure and MAE. Note that the proposed DPConv is only equipped with the encoder network. We use postfix ``-DP'' to represent that the model has been equipped with our DPConv. The better performances highlighted via \textbf{bold} font.\vspace{-0.2cm}}
\label{ablation_tab1}
\end{table}

As shown in Table \ref{ablation_tab1}, these SOD models equipped with our DPConv could achieve persistent performance gains over the raw versions.
In particular, F3Net20-DP has an improvement of $1.5\%$ in terms of $F_\beta$ on the DUT-OMRON dataset.
Similarly, MINet20-DP outperforms MINet20~\cite{pang2020multi} by $1.3\%$ on DUT-OMRON in terms of $F_\beta$. We conclude that the proposed DPConv is more effective than the standard residual block when processing multi-scale information, which also demonstrates the effectiveness of the DPConv block.

\begin{figure}[t]
\centering
\includegraphics[width=1\linewidth]{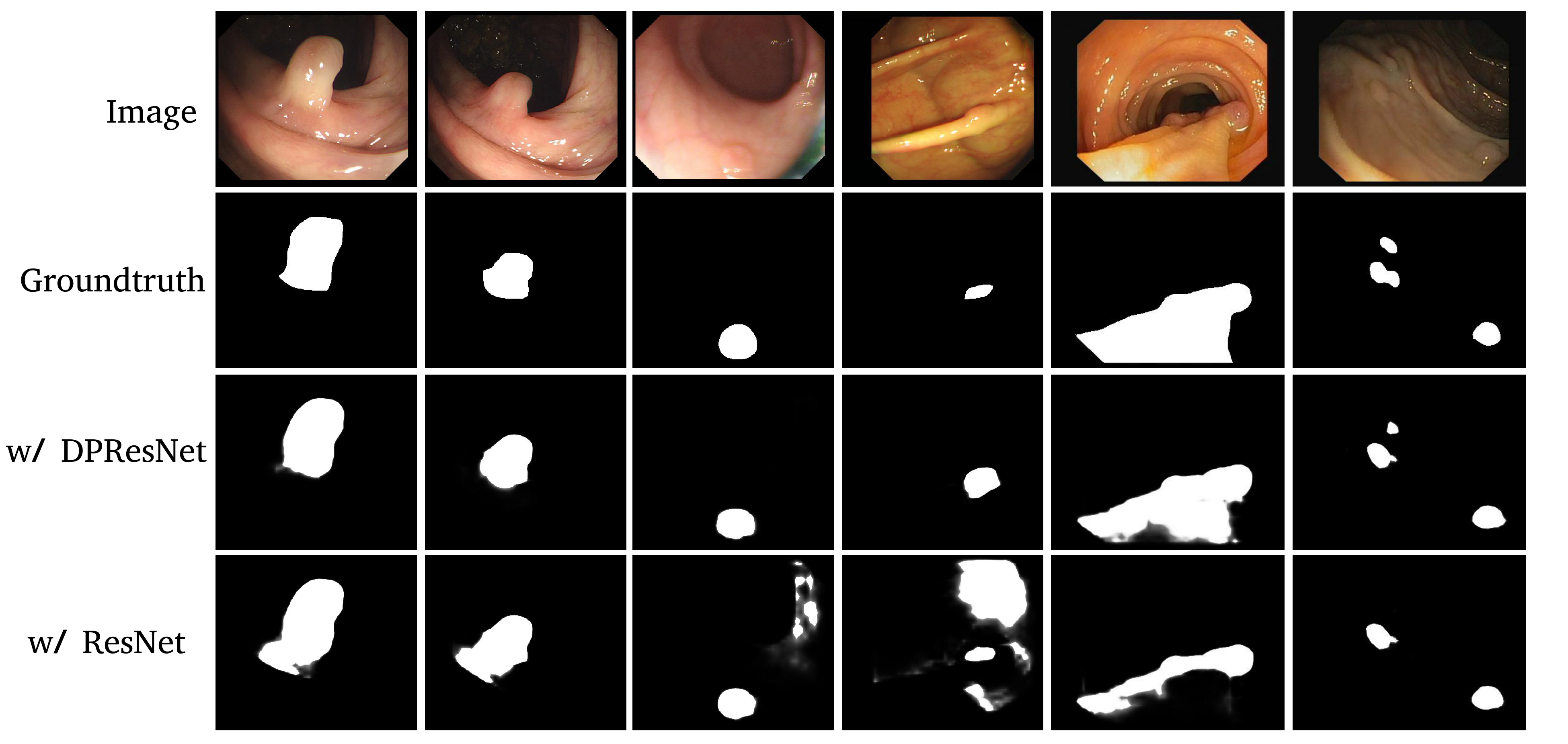}
\vspace{-0.7cm}
\caption{Visual comparisons of the proposed DPResNet with the ResNet, where segmentation maps produced by our DPResNet are more accurate and complete in various scenarios.}
\label{additional_segmentation}
\end{figure}

To further illustrate the advantages of the proposed DPResNet, we newly conducted experiments on polyp segmentation with the SOTA model, i.e., SANet \cite{SANet}. As shown in Fig. \ref{additional_segmentation}, our SANet with DPResNet can produce more accurate and complete segmentation maps in various scenarios. These experimental results further demonstrate the versatility of the DPResNet.

To understand the multi-scale representation capability of DPResNet, we introduce the commonly-used \textbf{c}lass \textbf{a}ctivation \textbf{m}apping (CAM) to explain why the predicted saliency maps of our DPResNet are better than that of the ResNet based one.
Concretely, we visualize the CAM by using Grad-CAM~\cite{selvaraju2017grad}, which is a simple yet effective method to visualize image regions with the strongest responses, i.e., the salient objects in our case.

\begin{figure}[!t]
\centering
\includegraphics[width=1\linewidth]{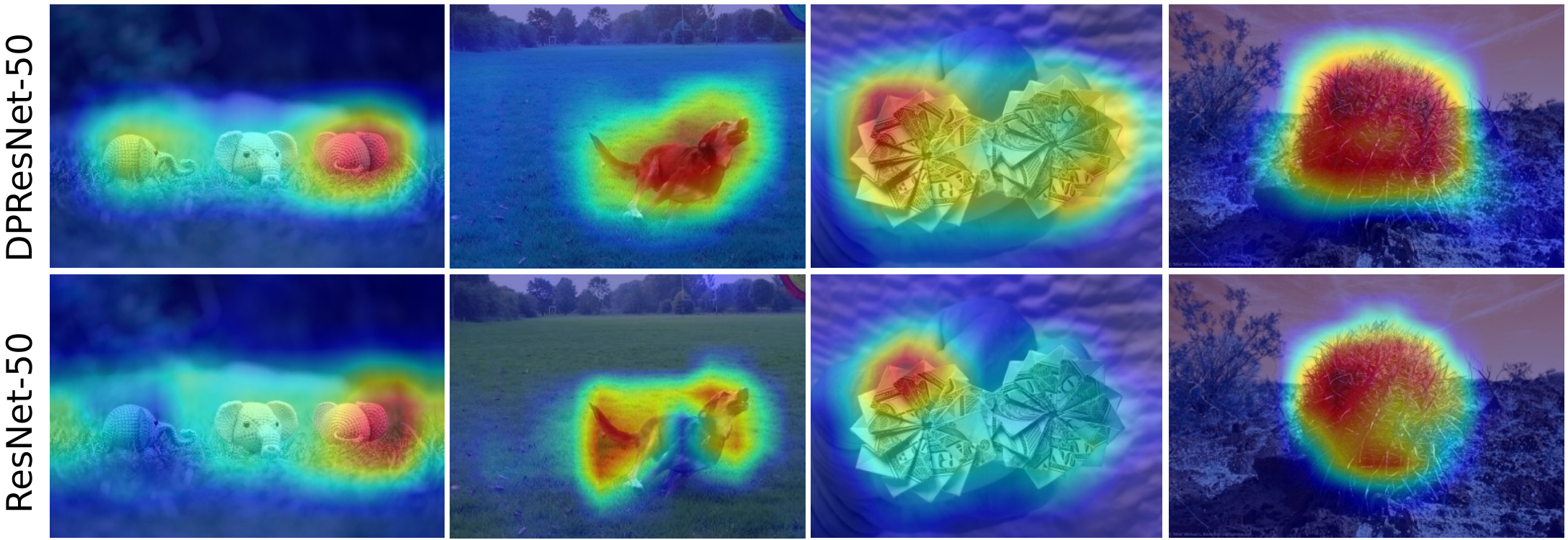}
\vspace{-0.6cm}
\caption{Visualization of class activation mapping \cite{selvaraju2017grad}, using DPResNet-50 and ResNet-50 as backbone networks.\vspace{-0.4cm}}
\label{heatmap}
\end{figure}

As shown in Fig.~\ref{heatmap}, compared with ResNet-50, the CAM generated by DPResNet-50 tends to cover the whole objects, while the CAM of ResNet-50 only covers the objects partially.
In addition, the proposed DPResNet-50 can simultaneously localize multiple objects with accurate activation maps, and such ability in localizing the most discriminative regions makes it potentially suitable for the pixel-wise SOD task.

\begin{table}[!t]
\Large{
\linespread{2}
\renewcommand\arraystretch{1.1}
\resizebox{1\textwidth}{!}{
\begin{tabular}{l||ccc|ccc|rrrr}
\Xhline{1.2pt}
\multirow{2}{*}{Backbone}

& \multicolumn{3}{c|}{DUT-OMRON}
& \multicolumn{3}{c|}{DUTS-TE}

 & \multirow{2}{*}{\#Params}
  & \multirow{2}{*}{\#FLOPs}
\\
\cline{2-7}
 &$F_\beta$  & S-m$\uparrow$ & MAE$$    &$F_\beta$  & S-m$$ &MAE$$    \\
\hline \hline

\rowcolor{gray!30} \textbf{DPResNet-50}(Ours)  & \textbf{.837} & \textbf{.853} &  \textbf{.049}& \textbf{.916} &  \textbf{.912} &  \textbf{.029} & 27.1M  & 9.2G \\

ResNet-50 \cite{he2016deep} & .819 &  .834 & .060 & .892 & .889 & .035 &27.8M &9.4G \\

DenseNet-121 \cite{huang2017densely} & .825 &  .844 & .057 & .893 & .889 & .039  &11.1M & 8.2G  \\

Res2Net50 \cite{2021res2net} & .823 &  .844 & .055 & .900 & .895 & .036 & 27.9M & 9.6G \\

\hline

\rowcolor{gray!30} \textbf{DPResNet-101}(Ours) & \textbf{.847} &   \textbf{.861} &  \textbf{.048} &  \textbf{.918} &  \textbf{.914} &  \textbf{.028} & 44.7M &12.6G  \\

ResNet-101 \cite{he2016deep} & .823 &  .841 &   .058 & .894 &   .890 &   .036    & 46.7M & 13.2G \\

Inception-V4 \cite{szegedy2017inception} & .826 & .850 & .049 & .908 & .905 & .030 &59.7M &16.5G\\

DenseNet-169 \cite{huang2017densely} & .825 &  .845&  .057 & .896 & .892&  .037 & 16.7M & 8.7G  \\

Res2Net101 \cite{2021res2net} & .836 & .852 & .049 & .907&  .903 & .031  & 47.4 M &  13.4G \\

\hline

\rowcolor{gray!30} \textbf{DPResNet-152}(Ours) & \textbf{.847} & \textbf{.864} &  \textbf{.047} &  \textbf{.920} &  \textbf{.915} &  \textbf{.027} & 59.1M & 16.0G \\

ResNet-152 \cite{he2016deep} &  .827   & .848 &  .053 & .901&  .898 & .033  &  62.4M & 16.9G \\

DenseNet-201 \cite{huang2017densely} & .837 &  .855 & .050 & .906 & .899 &   .033 & 22.4M & 9.7G\\

\hline
\Xhline{1.4pt}
 \end{tabular}  }}
  \vspace*{-0.5em}
\caption{Quantitative comparisons of deeper DPResNet with related multi-scale representative methods. Noticing that the official code of Res2Net doesn't provide the 152 layer implementation.\vspace{-0.2cm}}
\label{ablation_tab2}
\end{table}

\subsubsection{Going Deeper with DPResNet}

Previous works~\cite{vgg2015,he2016deep} indicated that deeper networks have more powerful feature representation capability. To validate such ability of our model, we compare the SOD performance of the DPResNet with SOTA models, including ResNet~\cite{he2016deep}, DenseNet~\cite{huang2017densely}, Inception-V4~\cite{szegedy2017inception}, and Res2Net~\cite{2021res2net}.
Three groups of results are shown in Table~\ref{ablation_tab2}.

In the 1st group, our DPResNet-50 shows $1.8\%$ better performance than ResNet-50 in terms of F$_\beta$ averagely without using additional \textit{parameters} and \textit{FLOPs}. Also, our DPResNet-50, which could obtain a better balance between accuracy and efficiency, achieves $1.4\%$ higher F-measure than Res2Net-50.
In the 2nd group, our DPResNet-101 achieves better performance than the Inception-V4 ($82.6$$\rightarrow$$84.7$\% on DUT-OMRON in terms of F$_\beta$) while reducing the computational cost by almost $25\%$.
In the 3rd group, the DPResNet-152 also outperforms the ResNet-152 while costing $5\%$ less computation.
When compared with the DenseNet, our DPResNet achieves better performance than DenseNet at the expense of more parameters. Noticing that, breaking the upper limit of performance is often more important than using fewer parameters. Thus, our method aims atbreaking such performance upper bound by using as few parameters as possible.

In a word, the proposed DPResNet achieves significant performance gains over other multi-scale representation methods, suggesting the effectiveness of our proposed DPConv for the SOD task.

\begin{table}[t]

\caption{ Quantitative comparisons among the proposed DPNet, CondConv and DCNet.} 
\vspace{-0.7em}
\centering
\LARGE{
\linespread{2}
\renewcommand\arraystretch{1.2}
\setlength\tabcolsep{2pt}
\resizebox{1\textwidth}{!}{
\begin{tabular}{l||ccccc|ccccccc}
\Xhline{1pt}
\multirow{2}{*}{Methods}
& \multicolumn{5}{c|}{DUT-OMRON}
& \multicolumn{5}{c}{DUTS-TE}
\\
\cline{2-11}
 &$F_\beta$  & MAE$$ &W-$F_\beta$  & S-m$$ & E-m$$   
&$F_\beta$  & MAE$$  &W-$F_\beta$  & S-m$$ &E-m$$   
\\
\hline\hline

\textbf{DPNet(Ours)} 
 & \textbf{.837} &    \textbf{.049} &   \textbf{.770}  & \textbf{.853}  & \textbf{.876} 
&  \textbf{.916}   &  \textbf{.029} & \textbf{.849} & \textbf{.912}  & \textbf{.921} \\

 DCNet \cite{chen2020dynamic}

 & {.828} &    {.050} &   {.762}  & {.847}  & {.864} 
&  {.906}   &  {.031} & {.837} & {.903}  & {.917} \\

{CondConv} \cite{yang2019condconv}
& .821    &   .0502  & {.758}  & .840  & .860
& .894    &   {.033}  &  {.843} & .895   & .913\\

\Xhline{1.4pt}
 \end{tabular}  }}

\label{results_condconv}
\end{table}

\begin{figure*}[t]
\centering

\resizebox{0.99\linewidth}{!}{
\subfigure{
\begin{minipage}[b]{0.33\linewidth}
\includegraphics[width=1\linewidth]{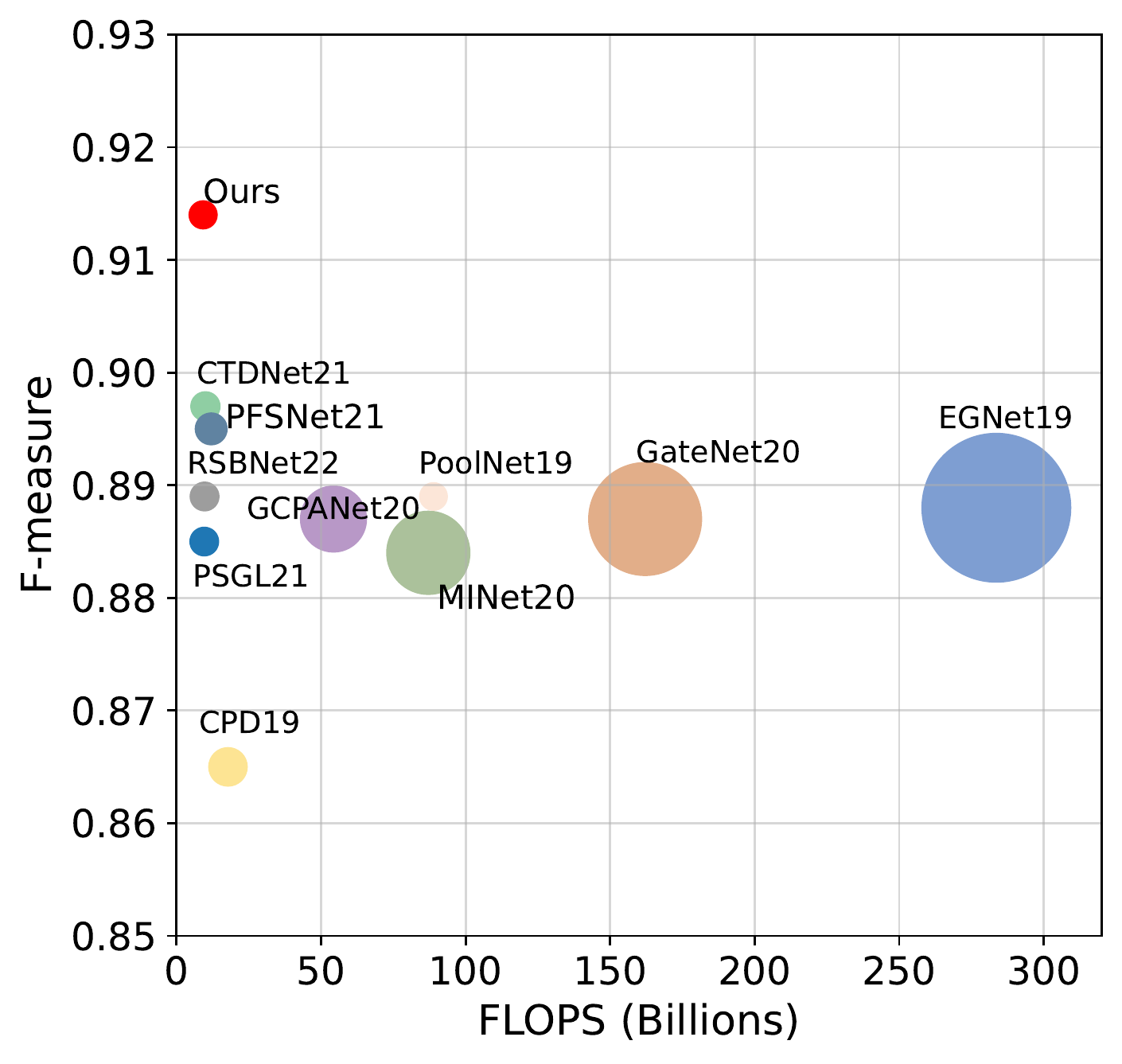}
\end{minipage}}\hspace{0.1pt}\hspace*{0.8em}

\subfigure{
\begin{minipage}[b]{0.33\linewidth}
\includegraphics[width=1\linewidth]{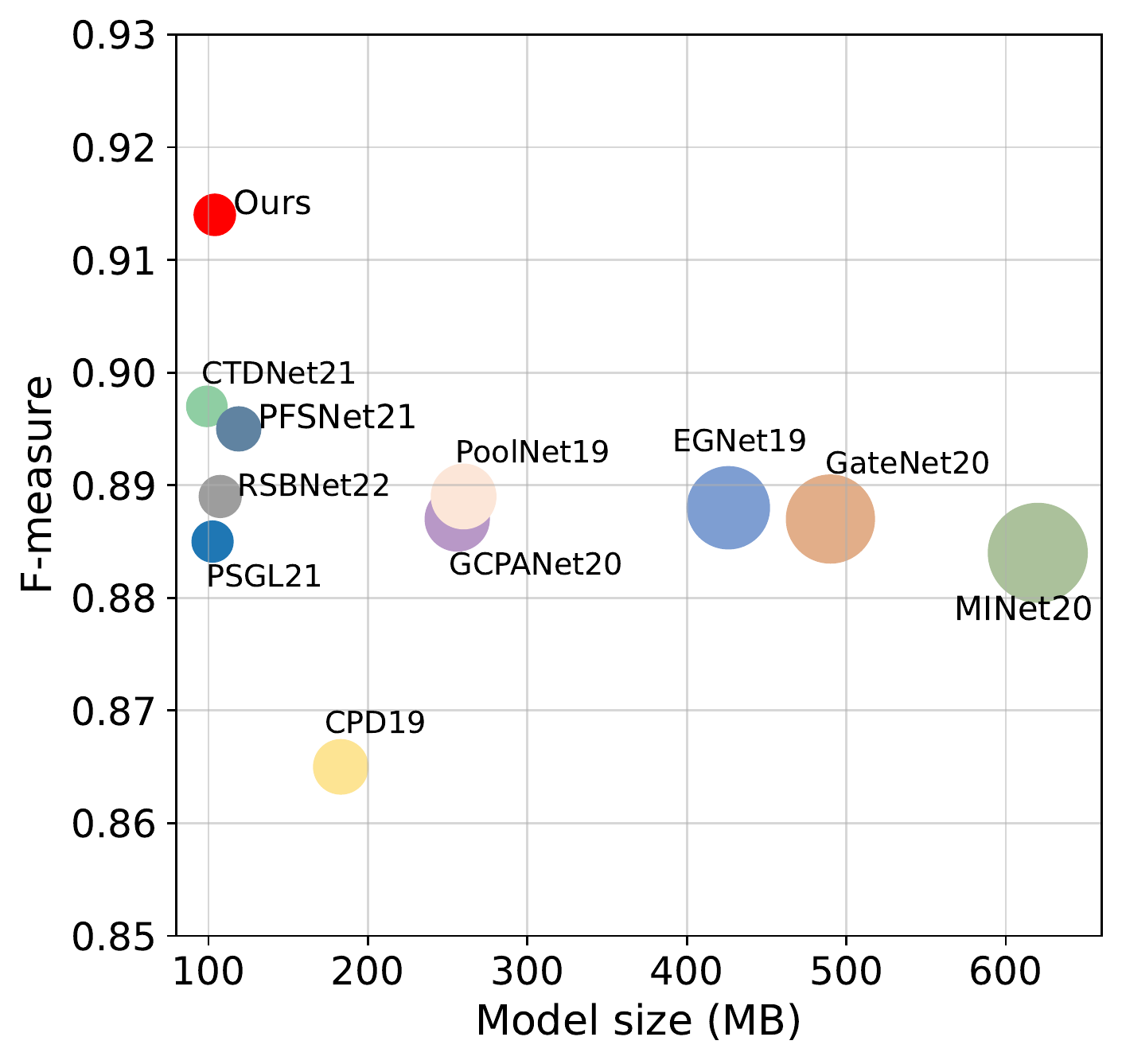}
\end{minipage}}\hspace{0.1pt}\hspace*{0.8em}

\subfigure{
\begin{minipage}[b]{0.34\linewidth}
\includegraphics[width=1\linewidth]{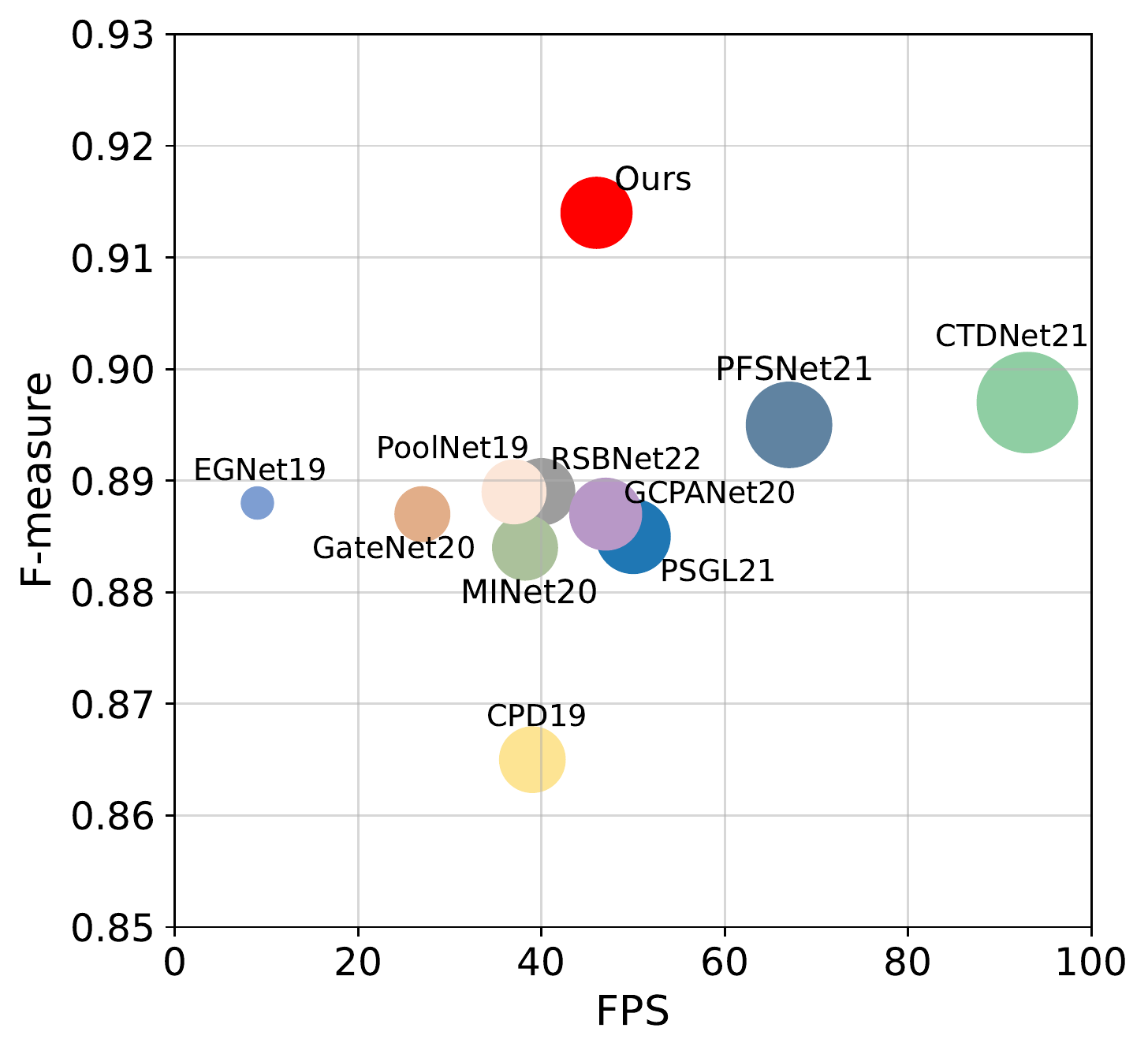}
\end{minipage}}\vspace*{-1.3em}
}

\vspace{-0.4cm}
\caption{Comparisons in terms of \textbf{maxF}, \textbf{FLOPs}, \textbf{Model-size}, and \textbf{FPS} towards our DPNet vs. RSBNet22 \cite{ke2022recursive}, CTDNet21 \cite{zhao2021complementary}, PSGL21 \cite{yang2021progressive}, PFSNet21~\cite{ma2021PFSNet}, GateNet20 \cite{zhao2020suppress}, GCPANet20 \cite{chen2020global}, MINet20 \cite{pang2020multi}, CPD19 \cite{CPD}, PoolNet19 \cite{PoolNet}, EGNet19 \cite{zhao2019egnet} on the challenge DUTS-TE set~\cite{yang2013saliency}. All methods are based on ResNet~\cite{he2016deep} network. Circles' sizes are proportional to their FLOPs/Model-sizes/FPS.\vspace{-0.2cm}}
\label{params_flops_model_size}
\end{figure*}

\subsubsection{Efficiency Analysis}

In the previous Sec.~\ref{DPConv}, we have shown the computational cost of DPConv over the vanilla convolution. Here we extensively compare the proposed method with others in terms of F-measure, parameters, FLOPs, model size and FPS with other popular methods which have publicly released their codes, and these methods are RSBNet22 \cite{ke2022recursive}, CTDNet21 \cite{zhao2021complementary}, PSGL21 \cite{yang2021progressive}, PFSNet21~\cite{ma2021PFSNet}, GateNet20 \cite{zhao2020suppress}, GCPANet20 \cite{chen2020global}, MINet20 \cite{pang2020multi}, CPD19 \cite{CPD}, PoolNet19 \cite{PoolNet}, and EGNet19 \cite{zhao2019egnet}.
The input sizes are set according to either their papers or the default implementations, and we employ the \textit{torchstat}\footnote{\url{https://github.com/Swall0w/torchstat}} as the analyzer tool to compute the total number of network parameters and FLOPs.
As shown in Fig.~\ref{params} and Fig.~\ref{params_flops_model_size}, our DPNet achieves better performance and efficiency than most previous methods on the DUTS-TE \cite{yang2013saliency}. In particular, our proposed DPNet achieves comparable computational cost with RSBNet22 \cite{ke2022recursive}, CTDNet21 \cite{zhao2021complementary}  and PFSNet21~\cite{ma2021PFSNet},  while it improves the F-measure score by $1.7\%$ averagely. Our DPNet model also significantly outperforms GateNet20 \cite{zhao2020suppress}, GCPANet20 \cite{chen2020global}, MINet20 \cite{pang2020multi}, CPD19 \cite{CPD},  EGNet19 \cite{zhao2019egnet}, and PoolNet19\cite{PoolNet}, in both F-measure and computational cost. For instance, MINet20 \cite{pang2020multi} achieves $88.4\%$ F-measure with $162.4 $ millions of parameters and $87.1$ GFLOPs. Compared to MINet20, our DPNet improves the F-measure by $3.2\%$, using about 6x fewer parameters and 9.5$\times$ fewer FLOPs.

Furthermore, we have also conducted a comparison regarding FPS in the 3rd column of Fig.~\ref{params_flops_model_size}. To ensure a fair comparison, all compared SOTA models were evaluated on the same machine. Among the compared real-time models, our approach also achieves competitive performance with 45 FPS on an NVIDIA GTX 2080 GPU.

\begin{figure}[t]
\centering
\includegraphics[width=0.98\linewidth]{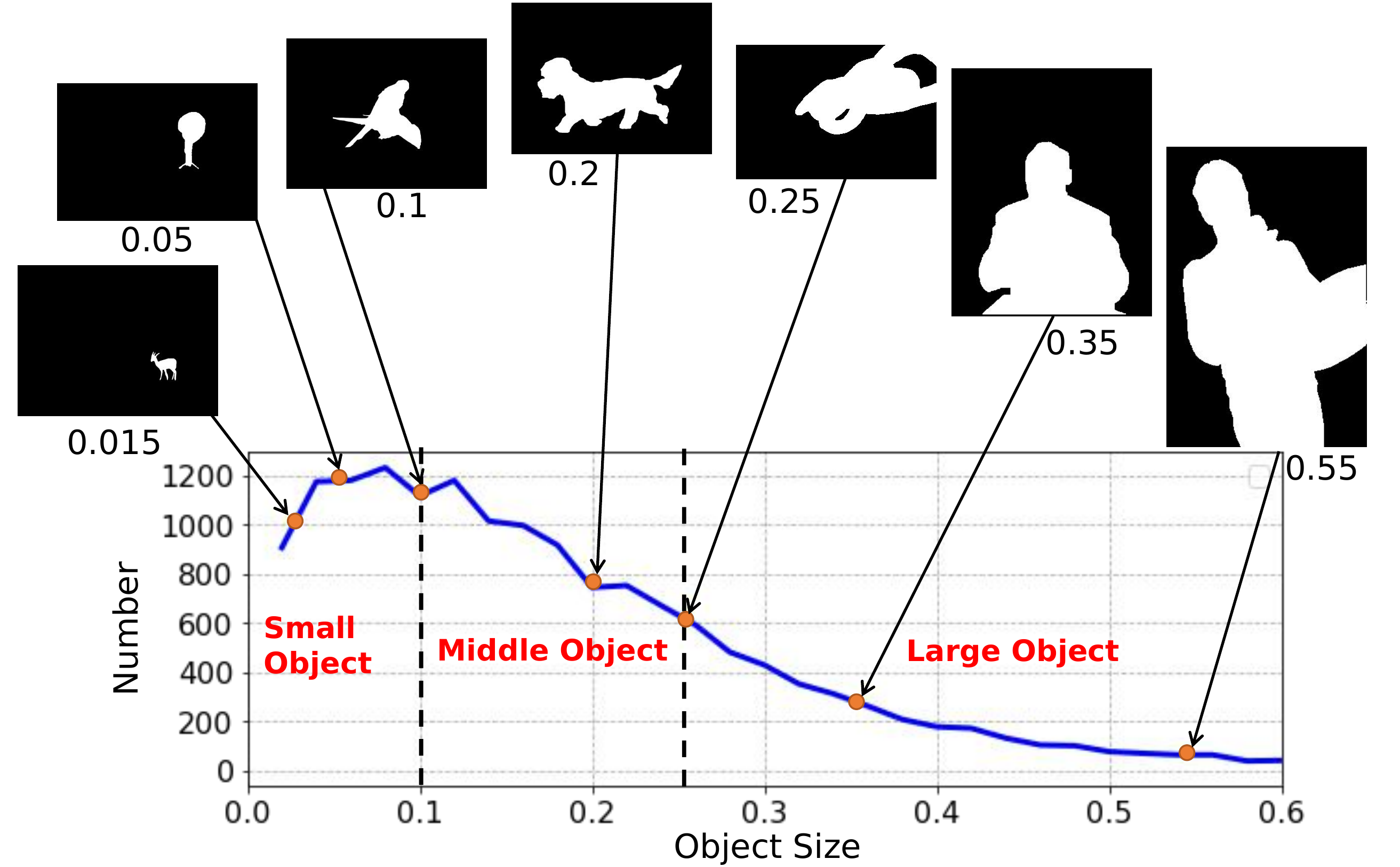}
\vspace{-0.2cm}
\caption{The statistics of object size of the DUTS-TE, DUT-OMRON, ECSSD, HKU-IS, and PASCAL-S datasets. \vspace{-0.2cm}}
\label{object_size}
\end{figure}

\begin{table}[!t]
\LARGE{
\linespread{2}
\renewcommand\arraystretch{1.1}
\resizebox{1\textwidth}{!}{
\begin{tabular}{c|c|c|c|c||ccc|ccccc}
\Xhline{1.2pt}
\multirow{2}{*}{ResNet50}
&\multirow{2}{*}{CFM}
&\multirow{2}{*}{BiCFM}
&\multirow{2}{*}{DPConv}
& \multirow{2}{*}{DWF}

& \multicolumn{3}{c|}{DUT-OMRON}
& \multicolumn{3}{c}{DUTS-TE}

\\
\cline{6-11}
 & & & &  &$F_\beta$  & S-m$$ & MAE$$    &$F_\beta$  & S-m$$ &MAE$$     \\
\hline\hline

\checkmark & \checkmark& &  & & .810 & .835 &  .057&  .887 &.885  & .041  \\

\checkmark & & \checkmark &  & & .818& .841  &  .053&  .895 &.892  & .034 \\

\checkmark& \checkmark &  &  \checkmark&   &  .823&  .842 &  .054 &   .896 &  .894 &  .035 \\

\checkmark &  & \checkmark & \checkmark  & & .829 & .850  &  .050&  .907 &.905  & .032   \\

\checkmark & \checkmark  & & \checkmark  & \checkmark  & .831 & .848  &  .052&  .909 &.908  & .031 \\

\checkmark & \checkmark  & \checkmark &   & \checkmark  & .825 & .844  &  .051&  .903 &.898  & .032 \\

\rowcolor{gray!30} \checkmark&  & \checkmark &  \checkmark& \checkmark   & \textbf{.837} & \textbf{.853} &  \textbf{.049}& \textbf{.916} &  \textbf{.912} &  \textbf{.029}   \\

\hline 

\multicolumn{5}{c||}{DPNet (w/o dynamic routing)} & .828 &  .847 &  .052&  .907 &.903  & .032  \\
 \rowcolor{gray!30} \multicolumn{5}{c||}{DPNet (w/ dynamic routing)} & \textbf{.837} & \textbf{.853} &  \textbf{.049}& \textbf{.916} &  \textbf{.912} &  \textbf{.029}  \\

\Xhline{1pt}
 \end{tabular}  }}
\vspace*{-0.4em}
\caption{Component evaluation in terms of max F-measure, S-measure and MAE on five commonly used dataset. The best results are shown in \textbf{bold} font.\vspace{-0.4cm}}
\label{ablation_study_4}
\end{table}

\subsubsection{Comparision with SOTA Dynamic Models}

As shown in Table \ref{results_condconv}, we also compared our DPNet with the recent proposed  CondConv \cite{yang2019condconv} and DCNet \cite{chen2020dynamic}. As we can see, the proposed DPNet consistently outperforms the CondConv and DCNet. Concretely, our DPNet has an improvement of $1.1\%$ in terms of $F_{\beta}$ on ECSSD dataset. Similarly, the proposed DPNet outperforms DCNet by $1.6\%$ on the ECSSD in terms of $F_{\beta}$. These experiments demonstrate that the proposed DPNet is more effective than the previous dynamic networks. 

\subsubsection{Statistics of Object Size}
 As shown in Fig. \ref{object_size}, we calculate the object size statistics of the DUTS-TE, DUT-OMRON, ECSSD, HKU-IS, and PASCAL-S datasets. Formally, we define the object size ($S$) by computing the proportion of foreground ($F$) against the whole image ($I$).
\begin{equation}
S = \frac{F}{ I}
\end{equation} 
We define the \textbf{Small Object} with $S<0.1$, \textbf{Middle Object} with $0.1 \le S \le 0.25$, and \textbf{Large Object} with $0.25 < S$. According to the statistics, small objects, middle objects and large objects account for $34.1\%$, $43.4\%$, and $22.5\%$ of the entire dataset, respectively. The results show that the Fig. \ref{multi_scale} of our original manuscript is not a special case. And small and large objects account for more than $56\%$ of the entire dataset. \\


\subsection{Component Evaluation}

Since our model consists of multiple components (i.e., DPConv, BiCFM, and DWF), we shall explore the exact contribution of each of these components. Table~\ref{ablation_study_4} shows the component evaluation results.
We start from the \{ResNet50+CFM\} as the baseline and then progressively extend it with different combinations.
The \{ResNet50+CFM\} is inherently limited by its weak multi-scale feature representation and one-directional feature propagation, and thus it performs the worst.
By replacing CFM with our BiCFM (in the 2nd row), we achieve $0.8\%$ improvement in F-measure on the DUT-OMRON.
We then replace the vanilla convolution with the proposed DPConv (in the 4th row), improving about $1.3\%$ in F-measure with fewer parameters and FLOPs.
Finally, we add the DWF to the model, achieving the best results.

As we can be seen, the model with all components has achieved the best performance, which demonstrates the necessity of each component for the proposed model to obtain the best SOD results. Besides, we have also evaluated the effect of the dynamic routing mechanism in the last two rows of Table~\ref{ablation_study_4}. With the dynamic routing, our DPNet achieves consistent performance enhancement.

\begin{figure*}[!t]
\centering
\resizebox{1\linewidth}{!}{
\subfigure{
\begin{minipage}[b]{0.2\linewidth}
\includegraphics[width=1\linewidth]{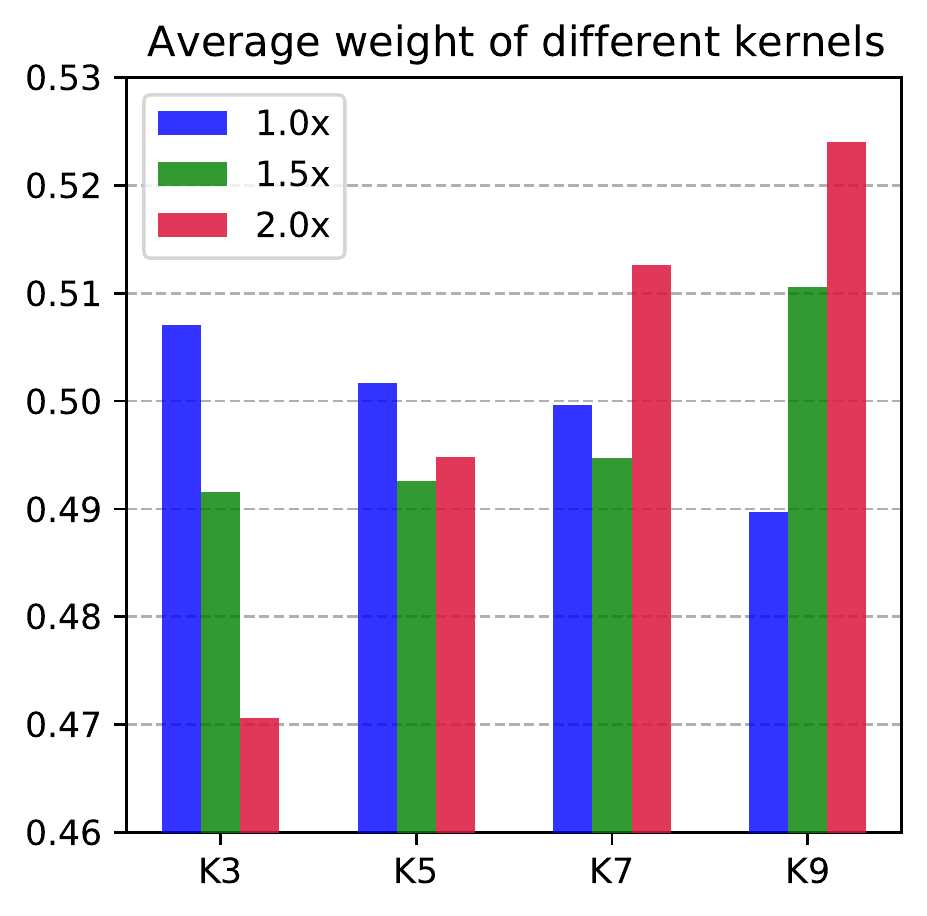}
\end{minipage}}\vspace*{-1em}
\subfigure{
\begin{minipage}[b]{0.2\linewidth}
\includegraphics[width=1\linewidth]{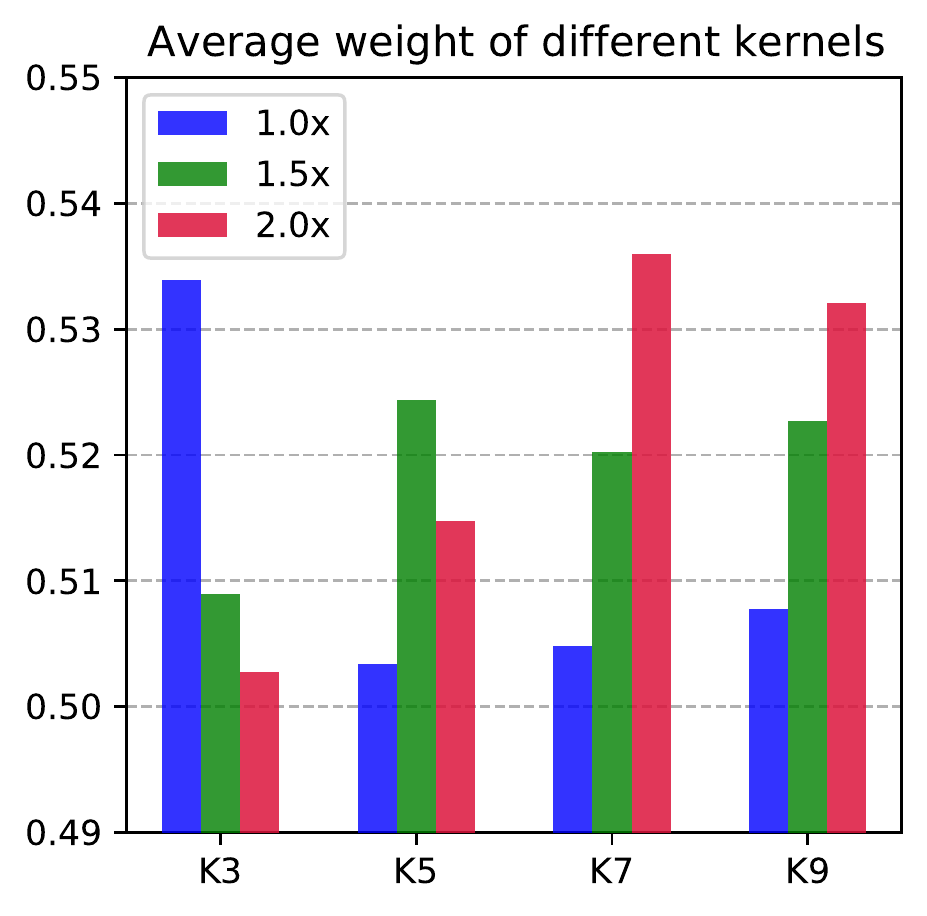}
\end{minipage}}\vspace*{-0.7em}
\subfigure{
\begin{minipage}[b]{0.2\linewidth}
\includegraphics[width=1\linewidth]{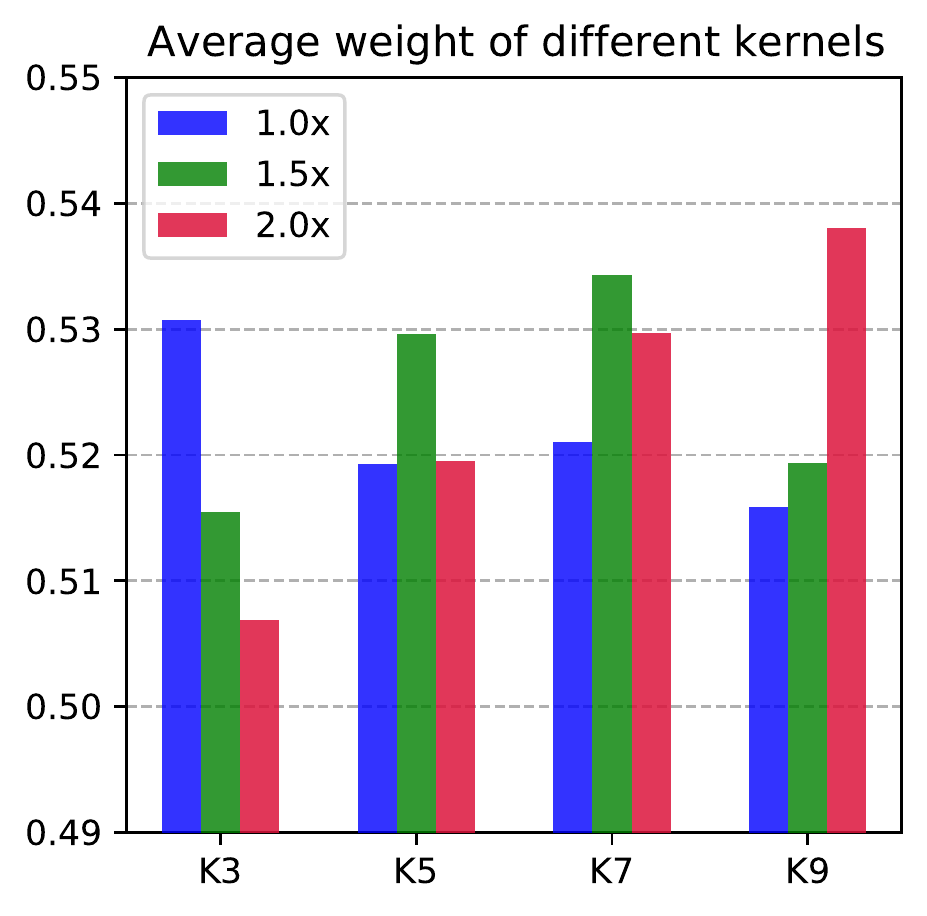}
\end{minipage}}\vspace*{-0.7em}
\subfigure{
\begin{minipage}[b]{0.2\linewidth}
\includegraphics[width=1\linewidth]{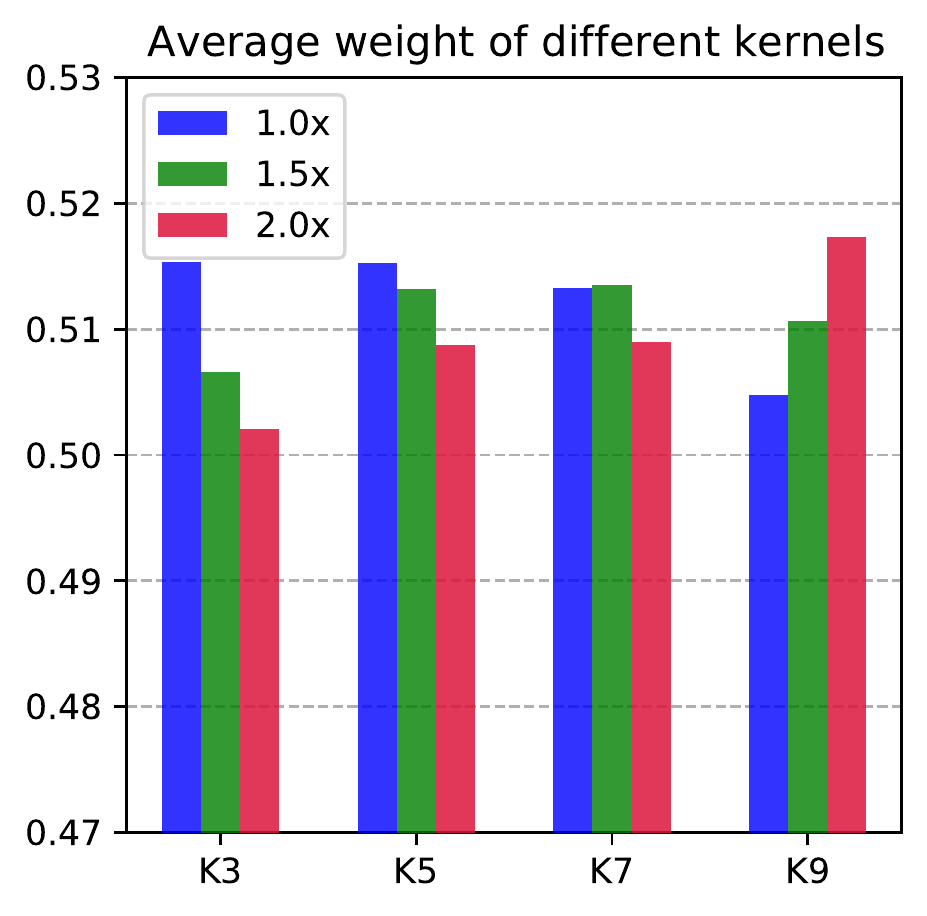}
\end{minipage}}
}

\vspace*{-1.3em}
\caption{Average kernels' scores for three different image sizes:$ 1.0\times$, $1.5\times$ and $2.0\times$.\vspace{-0.3cm}}
\label{analysis_weights}
\end{figure*}

\begin{table}[!t]
\Large{
\linespread{2}
\renewcommand\arraystretch{1.1}
\resizebox{0.9\textwidth}{!}{
\begin{tabular}{c|c|c|c|c||ccc|ccc}
\Xhline{1.2pt}
\multirow{2}{*}{K3}
&\multirow{2}{*}{K5 }
&\multirow{2}{*}{K7 }
&\multirow{2}{*}{K9 }
&\multirow{2}{*}{K11 }
& \multicolumn{3}{c|}{DUT-OMRON}
& \multicolumn{3}{c}{DUTS-TE}

\\
\cline{6-11}
  & & & &     &$F_\beta$  & S-m$$ &MAE$$  &$F_\beta$  & S-m$\uparrow$ &MAE$$    \\
\hline\hline

$\checkmark$ & &   & &     &  .824&  .848 &.053  & .891 & .892&  .034 \\

 & $\checkmark$ &   & &    &  .823 & .850 &.049 &  .907 & .903&  .037  \\

 &  & \checkmark   & &     &  .821&  .848&  .050   & .905&  .902&  .032\\

 &  &   & \checkmark &     &  .816&  .842&  .051  &  .900&  .896&  .034  \\

 &  &   &  &\checkmark     &  .806 &   .835 &   .053 &   .899 &   .895 &   .033 \\

\hline

 \checkmark& \checkmark  &   &  &    &  .827 & .850 & .048  &  .909 & .905 & .030  \\

 \checkmark& \checkmark  & \checkmark   &  &     &  .831 & .848 & .050   & .913 & .911 & .029  \\

 \rowcolor{gray!30} \checkmark& \checkmark  & \checkmark   & \checkmark  &    & \textbf{.837} & \textbf{.853} & \textbf{.049} & \textbf{.916} & \textbf{.912} & \textbf{.029}  \\

 \checkmark& \checkmark  & \checkmark   & \checkmark &  \checkmark&   .832 & .849 & .051  &  .913 & .907 & .031    \\
 
\Xhline{1.2pt}
 \end{tabular}  }}
 \vspace*{-0.4em}
\caption{Results of DPResNet-50 with different combinations of kernels. \vspace{-0.2cm}}
\label{ablation_kernel}
\end{table}

\subsection{Combination of Different Kernels in DPConv}

We investigate how the combination of different kernels influences saliency performance.
Here, we adopt five commonly-used kernels $3\times3, 5\times5, 7\times7, 9\times9$, and $11\times11$. For simplicity, we use ``Kn'' to represent the $n\times n$ kernel. The quantitative results are shown in Table~\ref{ablation_kernel}.
We make the following observations.

First, for the case with single-scale kernels, the trade-off between kernel size and performance is a concave curve, where performance rises up and then slowly goes down.
In particular, K5 achieves its best performance, i.e., $1.6\%$ higher than K3 in terms of F-measure on the DUTS-TE set.
We attribute this improvement to the enlarged receptive field.
When increasing the kernel size, the model detects non-salient regions, resulting in performance degradation.

Second, when the adopted kernel types are increased, the performance usually improves, except for the K11.
This can be attributed to the dynamic routing mechanism, and we shall give an intuitive explanation by visualizing the activation maps at Conv5\_3 level for different scale kernels.
As shown in Fig.~\ref{analyze_pic}, each activation map has its own detecting pattern, yet these maps generally complement each other, indicating the optimal features representation can be captured via the proposed scale routing.

Third, further integrating K11 into a model, there will be slight performance degradation. One possible reason is that the proposed DPConv can already capture the multi-scale feature representation well by using kernel size from $K1$ to $K9$. If we further integrate the $K11$ into PPConv, it will introduce unnecessary feature perturbation, resulting in slight performance degradation. To strike the performance trade-off, we eventually choose the \{K3, K5, K7, K9\} combination.

\begin{table}[!t]
\centering
{
\linespread{2}
\renewcommand\arraystretch{1}
\resizebox{0.8\linewidth}{!}{
\begin{tabular}{c||ccc|ccc}
\Xhline{1.2pt}
\multirow{2}{*}{Setting}
& \multicolumn{3}{c|}{DUT-OMRON}
& \multicolumn{3}{c}{DUTS-TE}
\\
\cline{2-7}
 &$F_\beta$  & S-m$$ & MAE$$    &$F_\beta$  & S-m$$ &MAE$$     \\
\hline\hline
N=1 &.825 &  .849 & .051 & .912 & .907 & .029 \\
\rowcolor{gray!30} \textbf{N=2} & \textbf{.837} & \textbf{.853} &{.049} & \textbf{.916} & {.912} & {.029}    \\
N=3 & .831 & .853 & .050 & .915 &  \textbf{.914} & .030 \\
N=4 &.829 &  .850 &  \textbf{.047} & .914 & .911 &  \textbf{.027} \\

\Xhline{1.2pt}
 \end{tabular}  }}
  \vspace*{-0.4em}
\caption{Ablation study for different BiCFM numbers ($N$, Sec.~\ref{sec:bicfm}). Clearly, when $N$=2, the model achieves the best performance.\vspace{-0.2cm}}
\label{paraN}
\end{table}

\subsection{BiCFM Number (N) }

The BiCFM number $N$ (Sec.~\ref{sec:bicfm})  is a important hyperparameters for our model. We conduct a series of experiments on DUT-OMRON and DUTS-TE to study their effects. We gradually increase $N$ from 1 to 4 and evaluate their performance scores of F-measure, S-measure and MAE.
As shown in Table~\ref{paraN}, we empirically find that the optimal setting for the number of BiCFM is $N=2$ (the 2nd column).

\subsection{Analysis and Interpretation}

To further reveal how DPConv works, we analyze the weight score by training our model in different image scales.
Concretely, we take all images from the DUTS-TR set and then progressively enlarge the image size from $1.0\times$ to $2.0\times$. Fig.~\ref{analysis_weights} illustrates the learned weights for different input scales in DPConv layers.

We first calculate the averaged weights for different kernels (K3, K5, K7, and K9) in each channel in the DPConv layer. Fig.~\ref{analysis_weights} shows the weight scores in all channels for Conv2\_3, Conv3\_4, Conv4\_6, and Conv5\_3. As we can see, when the selected images' sizes are enlarged, the weights for the small kernel (K3) tend to decrease, yet the weights for the large kernel (K9) get increased, which suggests that the dynamic routing can adaptively select
appropriate feature representation according to the input images.
We can easily find that the larger the input image is,
the more attention will be assigned to kernels with large sizes (e.g., K7 and K9). In stage 5, such differences start to shrink.

\begin{figure}[!t]
\centering
\includegraphics[width=1\linewidth]{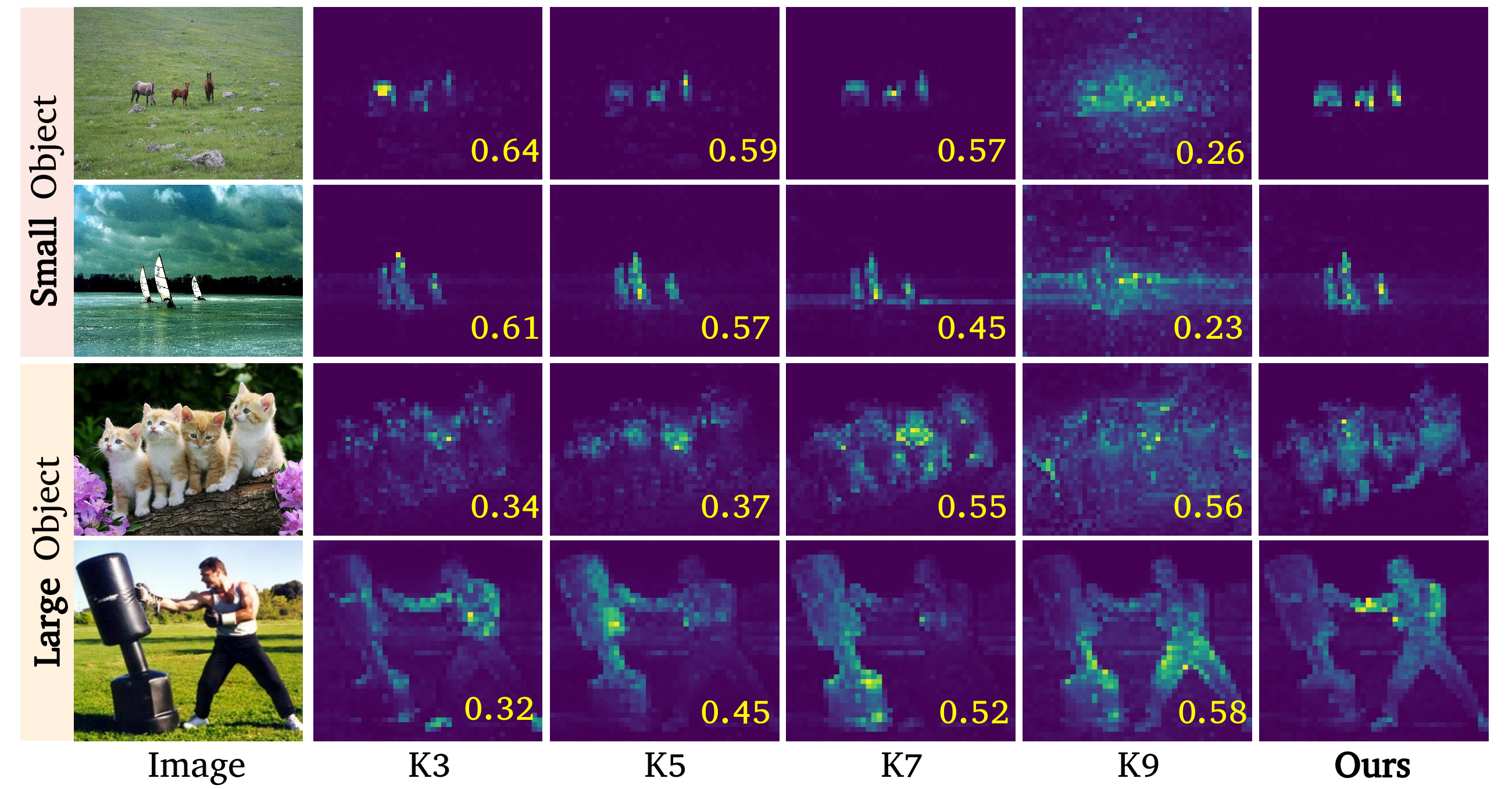}
\vspace{-0.6cm}
\caption{Visual interpretation of our proposed DPConv at the $Conv5\_3$ stage. K3-K9 denote the models equipped with different kernel sizes, e.g., K3 denotes the response maps using a $3\times 3$ kernel. ``Ours'' represent the response maps using the proposed DPConv.\vspace{-0.4cm}} 
\label{analyze_pic}
\end{figure}

In addition, a series of experiments have been conducted to verify our claim: the optimal feature representation is conditioned on the network’s input. To this end, we begin with using the popular ResNet-50 as the baseline and replacing the regular K3 convolution with K5, K7, and K9, respectively. We present some typical activation maps at Conv5\_3 level in Fig. \ref{analyze_pic}. We have the following observations.
First, feature representation with one type of kernel cannot achieve optimal performance. For instance, the horses (in the 1st row) are falsely activated in large kernel size (K9), but they can be predicted better in small kernels.
In the last row, the large salient regions, which prefer larger receptive fields, are better activated with a large kernel size (K9).
Second, different kernels tend to focus on different feature scales, which are complementary to each other.
Third, compared with the conventional settings, the proposed model with dynamic routing can simultaneously highlight salient regions and compress non-salient nearby surroundings.

This evidence suggested that generating feature representation conditioned on the input sample is promising.

\subsection{Limitations}
After taking a deeper look at the predicted saliency maps, we find that our DPNet tends to detect all possible saliency regions. The main reasons are two-fold: \textbf{1)} there are many controversial annotations scenes (people have different opinions about which is the salient object), which also confuses our DPNet as well as other SOTA models (e.g., PFSNet21 and GCPANet20). \textbf{2)} our DPNet contains multi-scale kernel size convolution layer, which can capture saliency objects with arbitrary size.
For instance, in the 1st row, our model can highlight the salient object completely, but it was confused by the tree trunk behind the bird. 
In the 2nd row,  all methods falsely detect the boat behind the camel as the salient object.

\begin{figure}[!t]
\centering
\includegraphics[width=0.98\linewidth]{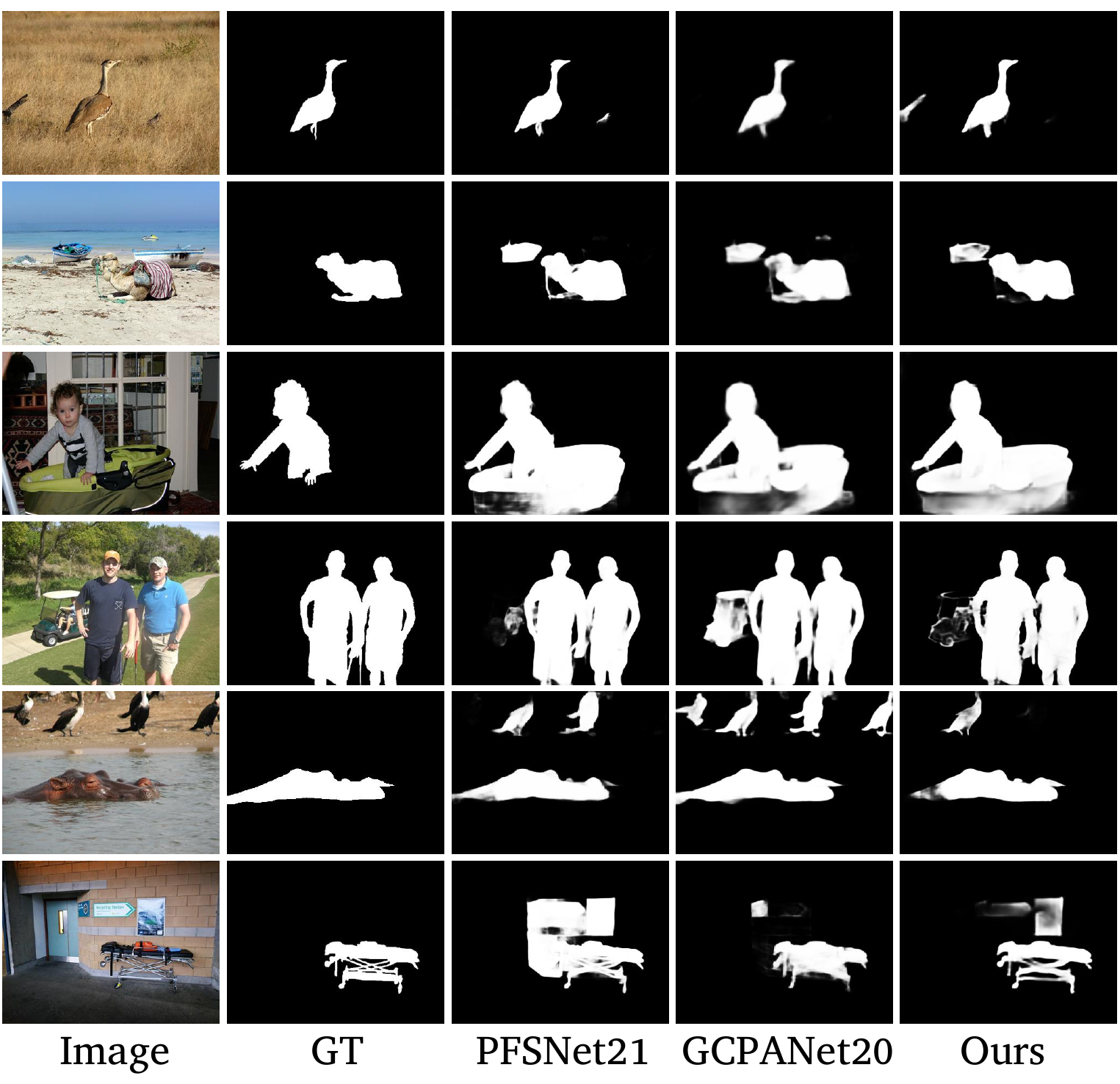}
\vspace{-0.2cm}
\caption{Failure cases. For a better demonstration, the 3d and 4th columns show the saliency maps produced by the current SOTA models, i.e., PFSNet21 \cite{ma2021PFSNet} and GCPANet20 \cite{chen2020global}.}
\label{failure_case}
\end{figure}

\section{Conclusion}

In this paper, we attempt to tackle the ``scale
confusion'' problem that is ubiquitous in all SOD-relevant tasks. We
sought a novel design of a DPConv layer, which tries to revolutionize
existing static convolutions with a novel dynamic routing mechanism. The newly-proposed DPConv is a generic plug-in for any
existing standard convolutional layer without much modification. The dynamic nature of the proposed DPConv ensures its
features are scale-aware, thus formulating implicit decision rules
to automatically bias towards those most useful scales. 
Moreover, to better match the DPConv-based encoder, we focused on a novel decoder's design inspired by the same dynamic rationale. The new decoder could dynamically collect scale-aware features in a
bidirectional cross-scale fashion, significantly improving the SOTA SOD performance. Extensive experiments on six benchmarks confirm the superiority of our proposed model with a much better trade-off between
performance and model efficiency.

\setcounter{equation}{0}
\renewcommand{\theequation}{A\arabic{equation}}
\setcounter{figure}{0}
\renewcommand{\thefigure}{A\arabic{figure}}

\vspace{0.4cm}
\textbf{Acknowledgments}. This research is supported in part by the National Natural Science Foundation of China (No. 62172437 and 62172246), the Open Project
Program of State Key Laboratory of Virtual Reality Technology
and Systems (VRLAB2021A05), the Youth Innovation and Technology
Support Plan of Colleges and Universities in Shandong Province
(2021KJ062).

\ifCLASSOPTIONcaptionsoff
  \newpage
\fi
\bibliographystyle{IEEEtran}
\bibliography{reference}

\begin{thebibliography}{10}
\providecommand{\url}[1]{#1}
\csname url@samestyle\endcsname
\providecommand{\newblock}{\relax}
\providecommand{\bibinfo}[2]{#2}
\providecommand{\BIBentrySTDinterwordspacing}{\spaceskip=0pt\relax}
\providecommand{\BIBentryALTinterwordstretchfactor}{4}
\providecommand{\BIBentryALTinterwordspacing}{\spaceskip=\fontdimen2\font plus
\BIBentryALTinterwordstretchfactor\fontdimen3\font minus
  \fontdimen4\font\relax}
\providecommand{\BIBforeignlanguage}[2]{{%
\expandafter\ifx\csname l@#1\endcsname\relax
\typeout{** WARNING: IEEEtran.bst: No hyphenation pattern has been}%
\typeout{** loaded for the language `#1'. Using the pattern for}%
\typeout{** the default language instead.}%
\else
\language=\csname l@#1\endcsname
\fi
#2}}
\providecommand{\BIBdecl}{\relax}
\BIBdecl

\bibitem{wang2019inferring}
W.~Wang, J.~Shen, X.~Dong, A.~Borji, and R.~Yang, ``Inferring salient objects
  from human fixations,'' \emph{IEEE Transactions on Pattern Analysis and
  Machine Intelligence}, vol.~42, no.~8, pp. 1913--1927, 2019.

\bibitem{wang2020paying}
W.~Wang, J.~Shen, X.~Lu, S.~C. Hoi, and H.~Ling, ``Paying attention to video
  object pattern understanding,'' \emph{IEEE Transactions on Pattern Analysis
  and Machine Intelligence}, vol.~43, no.~7, pp. 2413--2428, 2020.

\bibitem{wang2022looking}
W.~Wang, G.~Sun, and L.~Van~Gool, ``Looking beyond single images for weakly
  supervised semantic segmentation learning,'' \emph{IEEE Transactions on
  Pattern Analysis and Machine Intelligence}, 2022.

\bibitem{OurTIP15}
C.~Chen, S.~Li, H.~Qin, and A.~Hao, ``Structure-sensitive saliency detection
  via multilevel rank analysis in intrinsic feature space,'' \emph{IEEE
  Transactions on Image Processing}, vol.~24, no.~8, pp. 2303--2316, 2015.

\bibitem{cheng2015global}
M.-M. Cheng, N.~J. Mitra, X.~Huang, P.~H. Torr, and S.-M. Hu, ``Global contrast
  based salient region detection,'' \emph{IEEE Transactions on Pattern Analysis
  and Machine Intelligence}, vol.~37, no.~3, pp. 569--582, 2015.

\bibitem{ma2003contrast}
Y.-F. Ma and H.-J. Zhang, ``Contrast-based image attention analysis by using
  fuzzy growing,'' in \emph{Proceedings of the eleventh ACM international
  conference on Multimedia}, 2003, pp. 374--381.

\bibitem{achanta2009frequency}
R.~Achanta, S.~Hemami, F.~Estrada, and S.~Susstrunk, ``Frequency-tuned salient
  region detection,'' in \emph{Proceedings of the IEEE/CVF Conference on
  Computer Vision and Pattern Recognition}, 2009, pp. 1597--1604.

\bibitem{long2015fully}
J.~Long, E.~Shelhamer, and T.~Darrell, ``Fully convolutional networks for
  semantic segmentation,'' in \emph{Proceedings of the IEEE/CVF Conference on
  Computer Vision and Pattern Recognition}, 2015, pp. 3431--3440.

\bibitem{wei2020f3net}
J.~Wei, S.~Wang, and Q.~Huang, ``F$^3$net: Fusion, feedback and focus for
  salient object detection,'' in \emph{Proceedings of the AAAI Conference on
  Artificial Intelligence}, 2020, pp. 12\,321--12\,328.

\bibitem{pang2020multi}
Y.~Pang, X.~Zhao, L.~Zhang, and H.~Lu, ``Multi-scale interactive network for
  salient object detection,'' in \emph{Proceedings of the IEEE/CVF Conference
  on Computer Vision and Pattern Recognition}, 2020, pp. 9413--9422.

\bibitem{zhou2020interactive}
H.~Zhou, X.~Xie, J.-H. Lai, Z.~Chen, and L.~Yang, ``Interactive two-stream
  decoder for accurate and fast saliency detection,'' in \emph{Proceedings of
  the IEEE/CVF Conference on Computer Vision and Pattern Recognition}, 2020,
  pp. 1449--1457.

\bibitem{AFNet}
M.~Feng, H.~Lu, and E.~Ding, ``Attentive feedback network for boundary-aware
  salient object detection,'' in \emph{Proceedings of the IEEE/CVF Conference
  on Computer Vision and Pattern Recognition}, 2019, pp. 1623--1632.

\bibitem{su2019selectivity}
J.~Su, J.~Li, Y.~Zhang, C.~Xia, and Y.~Tian, ``Selectivity or invariance:
  Boundary-aware salient object detection,'' in \emph{Proceedings of the
  IEEE/CVF International Conference on Computer Vision}, 2019, pp. 3799--3808.

\bibitem{zhao2019egnet}
J.~Zhao, J.~Liu, D.~Fan, Y.~Cao, J.~Yang, and M.~Cheng, ``Egnet: Edge guidance
  network for salient object detection,'' in \emph{Proceedings of the IEEE/CVF
  International Conference on Computer Vision}, 2019, pp. 8779--8788.

\bibitem{wei2020label}
J.~Wei, S.~Wang, Z.~Wu, C.~Su, Q.~Huang, and Q.~Tian, ``Label decoupling
  framework for salient object detection,'' in \emph{Proceedings of the
  IEEE/CVF Conference on Computer Vision and Pattern Recognition}, 2020, pp.
  13\,025--13\,034.

\bibitem{li2020rgb}
C.~Li, R.~Cong, Y.~Piao, Q.~Xu, and C.~C. Loy, ``Rgb-d salient object detection
  with cross-modality modulation and selection,'' in \emph{European Conference
  on Computer Vision}, 2020, pp. 225--241.

\bibitem{zhang2021uncertainty}
J.~Zhang, D.-P. Fan, Y.~Dai, S.~Anwar, F.~Saleh, S.~Aliakbarian, and N.~Barnes,
  ``Uncertainty inspired rgb-d saliency detection,'' \emph{IEEE Transactions on
  Pattern Analysis and Machine Intelligence}, 2021.

\bibitem{fu2021siamese}
K.~Fu, D.-P. Fan, G.-P. Ji, Q.~Zhao, J.~Shen, and C.~Zhu, ``Siamese network for
  rgb-d salient object detection and beyond,'' \emph{IEEE Transactions on
  Pattern Analysis and Machine Intelligence}, 2021.

\bibitem{chen2018reverse}
S.~Chen, X.~Tan, B.~Wang, and X.~Hu, ``Reverse attention for salient object
  detection,'' in \emph{European Conference on Computer Vision}, 2018, pp.
  234--250.

\bibitem{PiCANet}
N.~Liu, J.~Han, and M.-H. Yang, ``Picanet: Learning pixel-wise contextual
  attention for saliency detection,'' in \emph{Proceedings of the IEEE/CVF
  Conference on Computer Vision and Pattern Recognition}, 2018, pp. 3089--3098.

\bibitem{wang2019salient}
W.~Wang, S.~Zhao, J.~Shen, S.~C. Hoi, and A.~Borji, ``Salient object detection
  with pyramid attention and salient edges,'' in \emph{Proceedings of the
  IEEE/CVF Conference on Computer Vision and Pattern Recognition}, 2019, pp.
  1448--1457.

\bibitem{PAGRN}
X.~Zhang, T.~Wang, J.~Qi, H.~Lu, and G.~Wang, ``Progressive attention guided
  recurrent network for salient object detection,'' in \emph{Proceedings of the
  IEEE/CVF Conference on Computer Vision and Pattern Recognition}, 2018, pp.
  714--722.

\bibitem{CPD}
Z.~Wu, L.~Su, and Q.~Huang, ``Cascaded partial decoder for fast and accurate
  salient object detection,'' in \emph{Proceedings of the IEEE/CVF Conference
  on Computer Vision and Pattern Recognition}, 2019, pp. 3907--3916.

\bibitem{wang2016saliency}
L.~Wang, L.~Wang, H.~Lu, P.~Zhang, and X.~Ruan, ``Saliency detection with
  recurrent fully convolutional networks,'' in \emph{European Conference on
  Computer Vision}, 2016, pp. 825--841.

\bibitem{wang2020progressive}
B.~Wang, Q.~Chen, M.~Zhou, Z.~Zhang, X.~Jin, and K.~Gai, ``Progressive feature
  polishing network for salient object detection,'' in \emph{Proceedings of the
  AAAI Conference on Artificial Intelligence}, 2020, pp. 12\,128--12\,135.

\bibitem{Amulet}
P.~Zhang, D.~Wang, H.~Lu, H.~Wang, and X.~Ruan, ``Amulet: Aggregating
  multi-level convolutional features for salient object detection,'' in
  \emph{Proceedings of the IEEE/CVF International Conference on Computer
  Vision}, 2017, pp. 202--211.

\bibitem{RADF}
X.~Hu, L.~Zhu, J.~Qin, C.-W. Fu, and P.-A. Heng, ``Recurrently aggregating deep
  features for salient object detection.'' in \emph{Proceedings of the AAAI
  Conference on Artificial Intelligence}, 2018, pp. 6943--6950.

\bibitem{BASNet19}
X.~Qin, Z.~Zhang, C.~Huang, C.~Gao, M.~Dehghan, and M.~Jagersand, ``Basnet:
  Boundary-aware salient object detection,'' in \emph{Proceedings of the
  IEEE/CVF Conference on Computer Vision and Pattern Recognition}, 2019, pp.
  7479--7489.

\bibitem{he2016deep}
K.~He, X.~Zhang, S.~Ren, and J.~Sun, ``Deep residual learning for image
  recognition,'' in \emph{Proceedings of the IEEE/CVF Conference on Computer
  Vision and Pattern Recognition}, 2016, pp. 770--778.

\bibitem{szegedy2017inception}
C.~Szegedy, S.~Ioffe, V.~Vanhoucke, and A.~Alemi, ``Inception-v4,
  inception-resnet and the impact of residual connections on learning,'' in
  \emph{Proceedings of the AAAI Conference on Artificial Intelligence}, 2017.

\bibitem{2021res2net}
S.-H. Gao, M.-M. Cheng, K.~Zhao, X.-Y. Zhang, M.-H. Yang, and P.~Torr,
  ``Res2net: A new multi-scale backbone architecture,'' \emph{IEEE Transactions
  on Pattern Analysis and Machine Intelligence}, 2021.

\bibitem{wang2017learning}
L.~Wang, H.~Lu, Y.~Wang, M.~Feng, D.~Wang, B.~Yin, and X.~Ruan, ``Learning to
  detect salient objects with image-level supervision,'' in \emph{Proceedings
  of the IEEE/CVF Conference on Computer Vision and Pattern Recognition}, 2017,
  pp. 136--145.

\bibitem{zhao2021complementary}
Z.~Zhao, C.~Xia, C.~Xie, and J.~Li, ``Complementary trilateral decoder for fast
  and accurate salient object detection,'' in \emph{Proceedings of the eleventh
  ACM international conference on Multimedia}, 2021, pp. 4967--4975.

\bibitem{zhao2020suppress}
X.~Zhao, Y.~Pang, L.~Zhang, H.~Lu, and L.~Zhang, ``Suppress and balance: A
  simple gated network for salient object detection,'' in \emph{European
  Conference on Computer Vision}, 2020, pp. 35--51.

\bibitem{wei2012geodesic}
Y.~Wei, F.~Wen, W.~Zhu, and J.~Sun, ``Geodesic saliency using background
  priors,'' in \emph{European Conference on Computer Vision}, 2012, pp. 29--42.

\bibitem{borji2015salient}
A.~Borji, M.-M. Cheng, H.~Jiang, and J.~Li, ``Salient object detection: A
  benchmark,'' \emph{IEEE Transactions on Image Processing}, vol.~24, no.~12,
  pp. 5706--5722, 2015.

\bibitem{xu2021locate}
B.~Xu, H.~Liang, R.~Liang, and P.~Chen, ``Locate globally, segment locally: A
  progressive architecture with knowledge review network for salient object
  detection,'' in \emph{Proceedings of the AAAI Conference on Artificial
  Intelligence}, vol.~35, no.~4, 2021, pp. 3004--3012.

\bibitem{liu2021rethinking}
J.-J. Liu, Z.-A. Liu, P.~Peng, and M.-M. Cheng, ``Rethinking the u-shape
  structure for salient object detection,'' \emph{IEEE Transactions on Image
  Processing}, vol.~30, pp. 9030--9042, 2021.

\bibitem{li2021salient}
J.~Li, J.~Su, C.~Xia, M.~Ma, and Y.~Tian, ``Salient object detection with
  purificatory mechanism and structural similarity loss,'' \emph{IEEE
  Transactions on Image Processing}, vol.~30, pp. 6855--6868, 2021.

\bibitem{wu2021decomposition}
Z.~Wu, L.~Su, and Q.~Huang, ``Decomposition and completion network for salient
  object detection,'' \emph{IEEE Transactions on Image Processing}, vol.~30,
  pp. 6226--6239, 2021.

\bibitem{tang2021disentangled}
L.~Tang, B.~Li, Y.~Zhong, S.~Ding, and M.~Song, ``Disentangled high quality
  salient object detection,'' in \emph{Proceedings of the IEEE/CVF
  International Conference on Computer Vision}, 2021, pp. 3580--3590.

\bibitem{zhuge2022salient}
M.~Zhuge, D.-P. Fan, N.~Liu, D.~Zhang, D.~Xu, and L.~Shao, ``Salient object
  detection via integrity learning,'' \emph{IEEE Transactions on Pattern
  Analysis and Machine Intelligence}, 2022.

\bibitem{wu2022synthetic}
Z.~Wu, L.~Wang, W.~Wang, T.~Shi, C.~Chen, A.~Hao, and S.~Li, ``Synthetic data
  supervised salient object detection,'' in \emph{Proceedings of the eleventh
  ACM international conference on Multimedia}, 2022.

\bibitem{wu2020deeper}
Z.~Wu, S.~Li, C.~Chen, A.~Hao, and H.~Qin, ``A deeper look at image salient
  object detection: Bi-stream network with a small training dataset,''
  \emph{IEEE Transactions on Multimedia}, 2020.

\bibitem{wu2022recursive}
------, ``Recursive multi-model complementary deep fusion for robust salient
  object detection via parallel sub-networks,'' \emph{Pattern Recognition},
  vol. 121, p. 108212, 2022.

\bibitem{CCTIP17}
C.~Chen, S.~Li, Y.~Wang, H.~Qin, and A.~Hao, ``Video saliency detection via
  spatial-temporal fusion and low-rank coherency diffusion,'' \emph{IEEE
  Transactions on Image Processing}, vol.~26, no.~7, pp. 3156--3170, 2017.

\bibitem{chen2020xuehao}
X.~Wang, S.~Li, C.~Chen, Y.~Fang, A.~Hao, and H.~Qin, ``Data-level
  recombination and lightweight fusionscheme for rgb-d salient object
  detection,'' \emph{IEEE Transactions on Image Processing}, 2020.

\bibitem{ChenPR16}
C.~Chen, S.~Li, and H.~Qin, ``Robust salient motion detection in non-stationary
  videos via novel integrated strategies of spatio-temporal coherency clues and
  low-rank analysis,'' \emph{Pattern Recognition}, vol.~52, pp. 410--432, 2016.

\bibitem{SRM}
T.~Wang, A.~Borji, L.~Zhang, P.~Zhang, and H.~Lu, ``A stagewise refinement
  model for detecting salient objects in images,'' in \emph{Proceedings of the
  IEEE/CVF International Conference on Computer Vision}, 2017, pp. 4019--4028.

\bibitem{DGRL}
T.~Wang, L.~Zhang, S.~Wang, H.~Lu, G.~Yang, X.~Ruan, and A.~Borji, ``Detect
  globally, refine locally: A novel approach to saliency detection,'' in
  \emph{Proceedings of the IEEE/CVF Conference on Computer Vision and Pattern
  Recognition}, 2018, pp. 3127--3135.

\bibitem{BMP}
L.~Zhang, J.~Dai, H.~Lu, Y.~He, and G.~Wang, ``A bi-directional message passing
  model for salient object detection,'' in \emph{Proceedings of the IEEE/CVF
  Conference on Computer Vision and Pattern Recognition}, 2018, pp. 1741--1750.

\bibitem{wang2019an}
W.~Wang, J.~Shen, M.~Cheng, and L.~Shao, ``An iterative and cooperative
  top-down and bottom-up inference network for salient object detection,'' in
  \emph{Proceedings of the IEEE/CVF Conference on Computer Vision and Pattern
  Recognition}, 2020, pp. 5968--5977.

\bibitem{DSS}
Q.~Hou, M.-M. Cheng, X.~Hu, A.~Borji, Z.~Tu, and P.~Torr, ``Deeply supervised
  salient object detection with short connections,'' in \emph{Proceedings of
  the IEEE/CVF Conference on Computer Vision and Pattern Recognition}, 2017,
  pp. 5300--5309.

\bibitem{chen2020global}
Z.~Chen, Q.~Xu, R.~Cong, and Q.~Huang, ``Global context-aware progressive
  aggregation network for salient object detection,'' in \emph{Proceedings of
  the AAAI Conference on Artificial Intelligence}, 2020, pp. 10\,599--10\,606.

\bibitem{wang2021salient}
W.~Wang, Q.~Lai, H.~Fu, J.~Shen, H.~Ling, and R.~Yang, ``Salient object
  detection in the deep learning era: An in-depth survey,'' \emph{IEEE
  Transactions on Pattern Analysis and Machine Intelligence}, vol.~44, no.~6,
  pp. 3239--3259, 2021.

\bibitem{tan2020efficientdet}
M.~Tan, R.~Pang, and Q.~V. Le, ``Efficientdet: Scalable and efficient object
  detection,'' in \emph{Proceedings of the IEEE/CVF Conference on Computer
  Vision and Pattern Recognition}, 2020, pp. 10\,781--10\,790.

\bibitem{bolukbasi2017adaptive}
T.~Bolukbasi, J.~Wang, O.~Dekel, and V.~Saligrama, ``Adaptive neural networks
  for efficient inference,'' in \emph{International Conference on Machine
  Learning}.\hskip 1em plus 0.5em minus 0.4em\relax PMLR, 2017, pp. 527--536.

\bibitem{wang2018skipnet}
X.~Wang, F.~Yu, Z.-Y. Dou, T.~Darrell, and J.~E. Gonzalez, ``Skipnet: Learning
  dynamic routing in convolutional networks,'' in \emph{European Conference on
  Computer Vision}, 2018, pp. 409--424.

\bibitem{wu2018blockdrop}
Z.~Wu, T.~Nagarajan, A.~Kumar, S.~Rennie, L.~S. Davis, K.~Grauman, and
  R.~Feris, ``Blockdrop: Dynamic inference paths in residual networks,'' in
  \emph{Proceedings of the IEEE/CVF Conference on Computer Vision and Pattern
  Recognition}, 2018, pp. 8817--8826.

\bibitem{klein2015dynamic}
B.~Klein, L.~Wolf, and Y.~Afek, ``A dynamic convolutional layer for short range
  weather prediction,'' in \emph{Proceedings of the IEEE/CVF Conference on
  Computer Vision and Pattern Recognition}, 2015, pp. 4840--4848.

\bibitem{jia2016dynamic}
X.~Jia, B.~De~Brabandere, T.~Tuytelaars, and L.~Van~Gool, ``Dynamic filter
  networks,'' in \emph{Advances in Neural Information Processing Systems},
  2016.

\bibitem{yang2019condconv}
B.~Yang, G.~Bender, Q.~V. Le, and J.~Ngiam, ``Condconv: Conditionally
  parameterized convolutions for efficient inference,'' \emph{Advances in
  Neural Information Processing Systems}, vol.~32, 2019.

\bibitem{chen2020dynamic}
Y.~Chen, X.~Dai, M.~Liu, D.~Chen, L.~Yuan, and Z.~Liu, ``Dynamic convolution:
  Attention over convolution kernels,'' in \emph{Proceedings of the IEEE/CVF
  Conference on Computer Vision and Pattern Recognition}, 2020, pp.
  11\,030--11\,039.

\bibitem{li2019selective}
X.~Li, W.~Wang, X.~Hu, and J.~Yang, ``Selective kernel networks,'' in
  \emph{Proceedings of the IEEE/CVF Conference on Computer Vision and Pattern
  Recognition}, 2019, pp. 510--519.

\bibitem{resnext2017}
S.~Xie, R.~Girshick, P.~Doll{\'a}r, Z.~Tu, and K.~He, ``Aggregated residual
  transformations for deep neural networks,'' in \emph{Proceedings of the
  IEEE/CVF Conference on Computer Vision and Pattern Recognition}, 2017, pp.
  1492--1500.

\bibitem{zeiler2014visualizing}
M.~D. Zeiler and R.~Fergus, ``Visualizing and understanding convolutional
  networks,'' in \emph{European Conference on Computer Vision}, 2014, pp.
  818--833.

\bibitem{liu2018path}
S.~Liu, L.~Qi, H.~Qin, J.~Shi, and J.~Jia, ``Path aggregation network for
  instance segmentation,'' in \emph{Proceedings of the IEEE/CVF Conference on
  Computer Vision and Pattern Recognition}, 2018, pp. 8759--8768.

\bibitem{zhang2017amulet}
P.~Zhang, D.~Wang, H.~Lu, H.~Wang, and X.~Ruan, ``Amulet: Aggregating
  multi-level convolutional features for salient object detection,'' in
  \emph{Proceedings of the IEEE/CVF Conference on Computer Vision and Pattern
  Recognition}, 2017, pp. 202--211.

\bibitem{ke2022recursive}
Y.~Y. Ke and T.~Tsubono, ``Recursive contour-saliency blending network for
  accurate salient object detection,'' in \emph{WACV}, 2022, pp. 2940--2950.

\bibitem{ma2021PFSNet}
M.~Ma, C.~Xia, and J.~Li, ``Pyramidal feature shrinking for salient object
  detection,'' in \emph{Proceedings of the AAAI Conference on Artificial
  Intelligence}, 2021.

\bibitem{yang2021progressive}
S.~Yang, W.~Lin, G.~Lin, Q.~Jiang, and Z.~Liu, ``Progressive self-guided loss
  for salient object detection,'' \emph{IEEE Transactions on Image Processing},
  vol.~30, pp. 8426--8438, 2021.

\bibitem{liu2021samnet}
Y.~Liu, X.-Y. Zhang, J.-W. Bian, L.~Zhang, and M.-M. Cheng, ``Samnet:
  Stereoscopically attentive multi-scale network for lightweight salient object
  detection,'' \emph{IEEE Transactions on Image Processing}, vol.~30, pp.
  3804--3814, 2021.

\bibitem{li2021uncertainty}
A.~Li, J.~Zhang, Y.~Lv, B.~Liu, T.~Zhang, and Y.~Dai, ``Uncertainty-aware joint
  salient object and camouflaged object detection,'' in \emph{Proceedings of
  the IEEE/CVF Conference on Computer Vision and Pattern Recognition}, 2021,
  pp. 10\,071--10\,081.

\bibitem{RANet20}
S.~{Chen}, X.~{Tan}, B.~{Wang}, H.~{Lu}, X.~{Hu}, and Y.~{Fu}, ``Reverse
  attention-based residual network for salient object detection,'' \emph{IEEE
  Transactions on Image Processing}, vol.~29, pp. 3763--3776, 2020.

\bibitem{PoolNet}
J.-J. Liu, Q.~Hou, M.-M. Cheng, J.~Feng, and J.~Jiang, ``A simple pooling-based
  design for real-time salient object detection,'' in \emph{Proceedings of the
  IEEE/CVF Conference on Computer Vision and Pattern Recognition}, 2019, pp.
  3917--3926.

\bibitem{qin2019basnet}
X.~Qin, Z.~Zhang, C.~Huang, C.~Gao, M.~Dehghan, and M.~Jagersand, ``Basnet:
  Boundary-aware salient object detection,'' in \emph{Proceedings of the
  IEEE/CVF Conference on Computer Vision and Pattern Recognition}, 2019, pp.
  7479--7489.

\bibitem{zhao2019pyramid}
T.~Zhao and X.~Wu, ``Pyramid feature attention network for saliency
  detection,'' in \emph{Proceedings of the IEEE/CVF Conference on Computer
  Vision and Pattern Recognition}, 2019, pp. 3085--3094.

\bibitem{yang2013saliency}
C.~Yang, L.~Zhang, H.~Lu, X.~Ruan, and M.-H. Yang, ``Saliency detection via
  graph-based manifold ranking,'' in \emph{Proceedings of the IEEE/CVF
  Conference on Computer Vision and Pattern Recognition}, 2013, pp. 3166--3173.

\bibitem{ecssd}
Q.~Yan, L.~Xu, J.~Shi, and J.~Jia, ``Hierarchical saliency detection,'' in
  \emph{Proceedings of the IEEE/CVF Conference on Computer Vision and Pattern
  Recognition}, 2013, pp. 1155--1162.

\bibitem{zhao2015saliency}
R.~Zhao, W.~Ouyang, H.~Li, and X.~Wang, ``Saliency detection by multi-context
  deep learning,'' in \emph{Proceedings of the IEEE/CVF Conference on Computer
  Vision and Pattern Recognition}, 2015, pp. 1265--1274.

\bibitem{li2014secrets}
Y.~Li, X.~Hou, C.~Koch, J.~M. Rehg, and A.~L. Yuille, ``The secrets of salient
  object segmentation,'' in \emph{Proceedings of the IEEE/CVF Conference on
  Computer Vision and Pattern Recognition}, 2014, pp. 280--287.

\bibitem{w-fmeasure}
M.~Ran, Z.-M. Lihi, and T.~Ayellet, ``How to evaluate foreground maps?'' in
  \emph{Proceedings of the IEEE/CVF Conference on Computer Vision and Pattern
  Recognition}, 2014, pp. 248--255.

\bibitem{Smeasure}
D.-P. Fan, M.-M. Cheng, Y.~Liu, T.~Li, and A.~Borji, ``Structure-measure: A new
  way to evaluate foreground maps.'' in \emph{Proceedings of the IEEE/CVF
  Conference on Computer Vision and Pattern Recognition}, 2017, pp. 4548--4557.

\bibitem{fan2018enhanced}
D.-P. Fan, C.~Gong, Y.~Cao, B.~Ren, M.-M. Cheng, and A.~Borji,
  ``Enhanced-alignment measure for binary foreground map evaluation,'' in
  \emph{International Joint Conferences on Artificial Intelligence}, 2018, pp.
  698--704.

\bibitem{MRNet20}
L.~{Zhang}, J.~{Wu}, T.~{Wang}, A.~{Borji}, G.~{Wei}, and H.~{Lu}, ``A
  multistage refinement network for salient object detection,'' \emph{IEEE
  Transactions on Image Processing}, vol.~29, pp. 3534--3545, 2020.

\bibitem{SANet}
J.~Wei, Y.~Hu, R.~Zhang, Z.~Li, S.~Zhou, and S.~Cui, ``Shallow attention
  network for polyp segmentation,'' in \emph{International Conference on
  Medical Image Computing and Computer-Assisted Intervention}.\hskip 1em plus
  0.5em minus 0.4em\relax Springer, 2021, pp. 699--708.

\bibitem{selvaraju2017grad}
R.~R. Selvaraju, M.~Cogswell, A.~Das, R.~Vedantam, D.~Parikh, and D.~Batra,
  ``Grad-cam: Visual explanations from deep networks via gradient-based
  localization,'' in \emph{Proceedings of the IEEE/CVF International Conference
  on Computer Vision}, 2017, pp. 618--626.

\bibitem{huang2017densely}
G.~Huang, Z.~Liu, L.~Van Der~Maaten, and K.~Q. Weinberger, ``Densely connected
  convolutional networks,'' in \emph{Proceedings of the IEEE/CVF Conference on
  Computer Vision and Pattern Recognition}, 2017, pp. 4700--4708.

\bibitem{vgg2015}
K.~Simonyan and A.~Zisserman, ``Very deep convolutional networks for
  large-scale image recognition,'' in \emph{International Conference on
  Learning Representations}, May 2015.

\end{thebibliography}

\end{document}